\documentclass[11pt,3p,review,authoryear]{elsarticle}

\usepackage[utf8]{inputenc} 
\usepackage[T1]{fontenc}
\usepackage{fix-cm}        
\usepackage{url}                 
\usepackage{amsfonts}       
\usepackage{nicefrac}       
\usepackage{changepage}   
\usepackage{graphicx}
\usepackage{enumitem}

\usepackage{silence}
\usepackage{caption}
\usepackage{epsfig} 
\usepackage{rotating}
\usepackage{amsmath} 
\usepackage{amssymb} 
\usepackage{dsfont} 
\usepackage{mathtools}
\usepackage{empheq}
\usepackage{listings}
\usepackage{bm} 
\usepackage{pgfgantt}
\usepackage{booktabs}  
\usepackage[flushleft]{threeparttable}
\usepackage[colorlinks=true, citecolor = blue, urlcolor = blue, linkcolor = blue]{hyperref}

\usepackage{scalerel}
\usepackage{algorithm}
\usepackage{algpseudocode}
\usepackage{multicol}
\usepackage{lmodern}

\usepackage{multirow}
\usepackage{booktabs}
\usepackage{longtable}

\usepackage{threeparttable}
\usepackage{threeparttablex}
\usepackage[normalem]{ulem}

\renewcommand\labelenumi{(\roman{enumi})}
\renewcommand\theenumi\labelenumi 
\usepackage{subcaption}

\usepackage{setspace}
\begin{document}

\author[ss]{\small{Alexander März}\corref{cor}}
\cortext[cor]{Corresponding author}
\ead{alex.maerz@gmx.net}
\address[ss]{\small{Independent Researcher}}
\renewcommand{\thefootnote}{}  
\footnotetext{Preprint submitted to  \textit{arXiv.}} 
\renewcommand{\thefootnote}{\arabic{footnote}}  

\author[bb]{\small{Kashif Rasul}}
\address[bb]{\small{Morgan Stanley Research}}


\begin{frontmatter}	
	
\vspace*{-2cm}
\enlargethispage{3cm}
	
\title{Forecasting with Hyper-Trees}

\begin{abstract}
We introduce Hyper-Trees as a novel framework for modeling time series data using gradient boosted trees. Unlike conventional tree-based approaches that forecast time series directly, Hyper-Trees learn the parameters of a target time series model, such as ARIMA or Exponential Smoothing, as functions of features. These parameters are then used by the target model to generate the final forecasts. Our framework combines the effectiveness of decision trees on tabular data with classical forecasting models, thereby inducing a time series inductive bias into tree-based models. To resolve the scaling limitations of boosted trees when estimating a high-dimensional set of target model parameters, we combine decision trees and neural networks within a unified framework. In this hybrid approach, the trees generate informative representations from the input features, which a shallow network then uses as input to learn the parameters of a time series model. With our research, we explore the effectiveness of Hyper-Trees across a range of forecasting tasks and extend tree-based modeling beyond its conventional use in time series analysis.
\end{abstract}


\end{frontmatter}

\begin{abstract}
We introduce Hyper-Trees as a novel framework for modeling time series data using gradient boosted trees. Unlike conventional tree-based approaches that forecast time series directly, Hyper-Trees learn the parameters of a target time series model, such as ARIMA or Exponential Smoothing, as functions of features. These parameters are then used by the target model to generate the final forecasts. Our framework combines the effectiveness of decision trees on tabular data with classical forecasting models, thereby inducing a time series inductive bias into tree-based models. To resolve the scaling limitations of boosted trees when estimating a high-dimensional set of target model parameters, we combine decision trees and neural networks within a unified framework. In this hybrid approach, the trees generate informative representations from the input features, which a shallow network then uses as input to learn the parameters of a time series model. With our research, we explore the effectiveness of Hyper-Trees across a range of forecasting tasks and extend tree-based modeling beyond its conventional use in time series analysis.
\end{abstract}

\section{Introduction} \label{sec:intro}

Gradient boosted decision trees (GBDTs) \citep{Friedman.2001} are widely recognized for their accuracy in both regression and classification tasks, while also scaling efficiently to large datasets. Transitioning from their original design as general-purpose models for tabular data, GBDTs have gained prominence in the forecasting domain, demonstrating strong performance across the VN1 \citep{Vandeput.2024}, M5 \citep{Makridakis.2022, Makridakis.2022b}, Corporaci\'{o}n Favorita \citep{Favorita.2017}, and Rossmann \citep{Knauer.2015} competitions. In an extensive overview, \cite{Januschowski.2022} attribute the success of GBDTs within a time series context to several factors. Key among these are their robustness, relative insensitivity to hyper-parameters, and versatility in handling numerical and categorical features. Further, GBDTs are effective at capturing complex non-linear relationships and interactions, as well as their adaptability to diverse time series characteristics. These attributes position tree-based models as competitive, out-of-the-box forecasting models for effective use across a wide range of time series applications.

Despite their advantages, general-purpose tree-based models are not designed to model sequential data naturally, which can limit their effectiveness in forecasting applications \citep{Godahewa.2023}. Unlike time series models such as ARIMA \citep{Box.2015} or Exponential Smoothing \citep{Holt.2004, Winters.1960}, GBDTs do not inherently capture the sequential structure of time series data. This usually necessitates careful feature engineering to embed time series information effectively into tree-models, e.g., via adding lagged or rolling features derived from the target to be forecasted \citep{Sprangers.2023, Nasios.2022}. A key reason for this constraint is that GBDTs operate under the assumption that observations are independent and identically distributed (i.i.d.). Although temporal dependencies can be approximated through engineered features, GBDTs process them in the same manner as any other input, reducing them to just another set of variables used for tree splitting without explicitly recognizing temporal dependencies that characterize time series. As a result, conventional GBDTs cannot naturally incorporate temporal inductive biases, such as autoregressive or trend and seasonality components in Exponential Smoothing models \citep{Holt.2004, Winters.1960}, without substantial algorithmic modifications. Furthermore, GBDTs face a structural limitation in their forecasting capabilities. Since most trees output piece-wise constants in the leaf nodes, standard implementations of GBDTs are not adept at forecasting beyond the range of the data they have been trained on \citep{Godahewa.2023}. A commonly adopted strategy for addressing some of these limitations involves first removing seasonality and trend from the series and then fitting a GBDT to the residuals. The final forecast is then composed of two parts: a local model that forecasts trend and seasonality, e.g., ARIMA \citep{Box.2015}, and a tree-based forecast. However, this approach has its own set of challenges. It is less suited for series that lack clear trends or seasonality, becomes complex to maintain for large-scale forecasting applications involving numerous time series, and prevents leveraging cross-series information since separate local models are trained to account for trend and seasonality. To better adapt tree-based models to a time series domain, several approaches have in response been proposed in the literature. Notable contributions include those of \cite{Shi.2019} and \cite{deVito.2017}, who enhance tree-models by replacing piece-wise constant trees with piece-wise linear trees. This adaptation enables tree-based forecasts to extend beyond the training data range, overcoming the limitations of standard GBDTs. Building on the relationship between threshold autoregressive models and decision trees, \cite{Godahewa.2023} introduce a specialized tree algorithm for forecasting. Their approach employs time series-specific splitting and stopping procedures while training global pooled regression models in the leaves to leverage cross-series information. 

These advancements have markedly improved the ability of GBDTs to better adapt to a time series context. Nevertheless, effective modeling of temporal dynamics using GBDTs presents ongoing challenges and remains an active area of research. To further progress in this domain, we propose a novel approach that extends tree-based models more naturally to a time series context via the use of Hyper-Trees. Unlike conventional tree-based approaches that forecast time series directly, Hyper-Trees generate forecasts through the following jointly trained process: the tree-component learns the parameters of a classical time series model as functions of input features, which the target model then uses to generate the final forecast. By reformulating GBDTs into a Hyper-Tree architecture, we combine the effectiveness of GBDTs on tabular data with classical forecasting models, such as ARIMA \citep{Box.2015} or Exponential Smoothing \citep{Holt.2004, Winters.1960}. While our framework is built upon the well-established LightGBM model \citep{Ke.2017}, it can in principle be used with any modern GBDT framework. To illustrate how our Hyper-Tree framework differs from and extends conventional time series forecasting, we contrast it with a traditional AR($p$) model in the following using the \texttt{Rossmann Store Sales} dataset, which we analyze in more detail in Section \ref{sec:experiments}. Rossmann operates over 3,000 drug stores across 7 European countries, and the task of the competition was to forecast 6 weeks of daily sales for 1,115 stores located in Germany \citep{Knauer.2015}. The data is characteristic of the complexity typical in operational time series forecasting \citep{Januschowski.2019}, with a rich set of features beyond basic time series information, including store types, sales campaigns, holiday information, as well as various categorical features. Conventional applications of an AR($p$) model would require training a separate, local model for each store $i=1, \ldots, N$
\begin{align}
	\text{Store 1}: y_{_{1t}} &= \theta_{_{1,1}}y_{_{1t-1}} + \theta_{_{2,1}}y_{_{1t-2}} + \cdots + \theta_{_{p,1}}y_{_{1t-p}} \nonumber \\
	&\raisebox{-1.0ex}{$\vdots$}  \\
	\text{Store N}: y_{_{Nt}} &= \theta_{_{1,N}}y_{_{Nt-1}} + \theta_{_{2,N}}y_{_{Nt-2}} + \cdots + \theta_{_{p,N}}y_{_{Nt-p}} \nonumber
\end{align}

with each store being forecasted in isolation and no information sharing across stores or series. In contrast, our proposed Hyper-Tree framework combines the strengths of GBDTs for modeling tabular data (categorical features, missing values, feature interactions and non-linear relationships, $\ldots$) with the effectiveness and simplicity of traditional time series models. Specifically, we assume an autoregressive structure for this example, with the Hyper-Tree serving as a global model that learns the AR($p$) parameters as functions of temporal and store-specific features via gradient boosted trees
\begin{align}     
	y_{_{it}} = \theta_{_{1,it}}(\mathbf{x}_{_{it}})y_{_{it-1}} + \theta_{_{2,it}}(\mathbf{x}_{_{it}})y_{_{it-2}} + \cdots + \theta_{_{p,it}}(\mathbf{x}_{_{it}})y_{_{it-p}}, \quad i = 1,\ldots,N
\end{align}

where $\mathbf{x}_{_{it}} = \{\text{store-type}_{i}, \text{region}_{i},\text{sales-campaign}_{it}, \text{holiday}_{it}, \text{week}_{t}, \text{month}_{t}, \ldots\}$. Among recent efforts to adapt gradient boosted trees for time series tasks, Hyper-Trees offer a new perspective by combining classical statistical models with machine learning. Rather than using GBDTs to forecast directly, we repurpose them to learn the parameters of a target time series model. This hybrid formulation offers several advantages for time series analysis and forecasting:

\begin{itemize}
	\item \textbf{Improved Extrapolation in Tree-Based Models} Hyper-Trees can forecast beyond the value range observed during training, overcoming a key limitation of standard GBDTs. Conventional GBDTs output piece-wise constant values, which limits their extrapolation capability compared to parametric models that can naturally extend beyond the training data range \citep{Godahewa.2023}. This limitation is particularly problematic in scenarios where sales need to be projected beyond historical ranges. For instance, a store launching a new sales campaign might expect sales significantly higher than any historical values, or a newly renovated store might show stronger growth trends than observed historically. While LightGBM \citep{Ke.2017} allows fitting linear \citep{Shi.2019} rather than piece-wise constant trees to partially address this issue, other GBDT implementations lack this functionality. To overcome these limitations, Hyper-Trees generate forecasts through the following jointly trained process: first, they model the AR($p$) parameters $\theta_{_{j, it}}(\mathbf{x}_{_{it}})$ for $j=1,\ldots,p$ and each store $i=1, \ldots, N$ based on its specific characteristics $\mathbf{x}_{_{it}}$ (e.g., store type, location, sales campaign, $\ldots$). Then, these parameters are applied to the target model to generate the final forecast. This approach allows Hyper-Trees to leverage the inductive bias of the target model, enabling forecasts to naturally extend beyond historical ranges. 		
	
	\item \textbf{Cross-Series Learning with Local Adaptivity} Hyper-Trees offer a framework that incorporates tree-based cross-series learning while preserving local adaptivity of classical models. Traditional approaches like ARIMA \citep{Box.2015} or Exponential Smoothing \citep{Holt.2004, Winters.1960} typically require individual model training for each time series. Hyper-Trees, in contrast, combine global tree-learning with local model adaptivity. Rather than estimating separate parameters for each store in isolation, Hyper-Trees learn a global mapping from features to target model parameters $f: \mathcal{X}^{^{C}} \longrightarrow \Theta^{^{P}}$, where $\mathcal{X}^{^{C}}$ represents the $C$-dimensional feature space of temporal and store-specific characteristics and $\Theta^{^{P}}$ represents the $P$-dimensional parameter space of the target model. This enables stores with similar characteristics to share information, which is particularly valuable when forecasts for a new store with limited history can leverage temporal dynamics learned from existing locations. Despite this global approach, Hyper-Trees preserve local adaptivity through store-specific features. While all stores are modeled jointly, each receives unique parameters based on its individual characteristics. For instance, a store in a high-competition area might have parameters that emphasize sales campaign effects, while a store near schools might have parameters that reflect holiday-related patterns.
	
	\item \textbf{Time-Varying Parameters} Hyper-Trees enable dynamic parameter estimation that adapts to changing conditions. Unlike classical time series models that estimate static parameters, Hyper-Trees generate time-varying coefficients $\theta_{_{j,it}}(\mathbf{x}_{_{it}})$ at each time point. These parameters vary both cross-sectionally through store-specific features like store type and region, as well as temporally through features such as week, month, and year. As an example, Hyper-Trees can learn distinct AR($p$) coefficients for weekday versus weekend patterns, automatically adjusting to capture different sales dynamics throughout the week. By updating its coefficients at each time step as new information becomes available, Hyper-Trees provide a flexible framework that responds to evolving environments \citep{Lee.2023}.
	
	\item \textbf{Parameter Estimation} Hyper-Trees provide a more comprehensive approach to parameter estimation. While conventional estimation of ARIMA \citep{Box.2015} or Exponential Smoothing \citep{Holt.2004, Winters.1960} models typically relies on first-order optimization methods using gradients only, Hyper-Trees leverage both gradient and Hessian information, potentially improving on parameter estimates. Furthermore, unlike estimation of classical linear time series models, Hyper-Trees learn parameters as non-linear functions of features, allowing for more flexible parameter estimation in complex, high-dimensional feature settings.
	
	\item \textbf{Model Transparency and Interpretability} Hyper-Trees provide a transparent modeling framework by combining the flexibility of gradient boosted trees with the interpretability of classical time series models. While the GBDT component learns parameters by modeling complex feature interactions, final forecasts are generated through well-established time series models whose parameters retain clear statistical and domain-specific meaning. Furthermore, practitioners can trace how these parameters evolve in response to changing conditions or external events. This level of transparency not only facilitates understanding of how the model arrives at its outputs but also aligns with interpretability requirements in high-stakes domains such as insurance, healthcare, and public policy \citep{Rudin.2019}.
					
	\item \textbf{Full Functionality of GBDTs} Hyper-Trees maintain compatibility with established GBDT frameworks like LightGBM \citep{Ke.2017}, preserving their full functionality within our framework. This includes handling missing values, feature importance analysis\footnote{Though SHAP values \citep{Lundberg.2020, Lundberg.2017} are supported to investigate feature effects on parameter estimates, this functionality is currently restricted by the LightGBM implementation and is only available when the linear tree option \citep{Shi.2019} is not used (as of LightGBM version \texttt{4.6.0}).}, as well as efficient processing of categorical and numeric features. Additionally, our framework supports imposing structural assumptions on estimated parameters, enabling monotonic time series patterns to be modeled either through GBDTs' built-in monotonicity constraints or by applying restrictions directly on parameter estimation. This compatibility enables practitioners to combine existing GBDT capabilities with classical time series modeling.
\end{itemize}

To maintain the scalability of our approach when estimating a large set of target model parameters, we propose a hybrid architecture that combines GBDTs and neural networks within a single framework. This approach preserves the strong feature learning capabilities of GBDTs for tabular data while leveraging the efficiency of neural networks in multi-output tasks. Traditionally, GBDTs use a one-vs-all strategy, estimating one tree per parameter, which can be computationally expensive with an increasing number of target model parameters \citep{Iosipoi.2022}. In contrast, our hybrid approach employs GBDTs to first generate informative representations from the input features, which are then decoded by a shallow neural network into the parameters of a target time series model. Both the GBDTs and the neural network are trained jointly, enabling end-to-end optimization and the integration of tree-based feature learning with neural-network-based parameter mapping. This contrasts with most existing approaches where GBDTs are trained first and then a neural network is trained separately. 

The remainder of this paper is organized as follows: Section \ref{sec:hyper_tree} introduces the Hyper-Tree framework and describes its integration with different target model classes. Section \ref{sec:literature} gives an overview of related works, while Section \ref{sec:experiments} presents the application of Hyper-Trees across various datasets. Section \ref{sec:framework_considerations} provides a conceptual analysis of our Hyper-Tree framework. Finally, Section \ref{sec:conclusion} concludes. Our exploration of Hyper-Trees aims not only to lay a foundation for future research but also to broaden the perspective on the potential of tree-based models beyond their conventional use in time series analysis and forecasting.

\section{Hyper-Trees} \label{sec:hyper_tree}

Hyper-networks, introduced by \cite{Ha.2017}, have proven to be a valuable concept in machine learning. These networks are designed to generate parameters for a target network, with their key feature being that the generated parameters are modeled as functions of features. The integration of hyper-networks has represented a considerable advancement in the field, offering new possibilities for adaptive and context-dependent modeling. Such integration enables the target network to adapt more seamlessly to changes in data patterns, without the need to retrain the target model \citep{Lee.2023}. While hyper-models have shown notable potential, they have primarily been explored within the realm of deep learning. In this paper, we recast this concept for GBDTs in time series forecasting. We term our framework Hyper-Trees, drawing a parallel to hyper-networks. Just as hyper-networks generate parameters for a target network, Hyper-Trees learn to generate parameters for a target time series model. The gradient-based nature of GBDTs allows the target model to be a member of a wide class of time series models, e.g., ARIMA \citep{Box.2015}, Exponential Smoothing \citep{Holt.2004, Winters.1960}, as well as other target model architectures. By training the Hyper-Tree and applying the estimated parameters to the target model, our approach flexibly and efficiently incorporates time series inductive biases into tree-based models. Compared to hyper-networks, which often require large amounts of data and are prone to overfitting, modeling the parameters of a target model via GBDTs offers a more efficient and robust approach, particularly in settings with limited data, as commonly encountered in real-world settings \citep{Bansal.2022, Vogelsgesang.2018}. While applicable to a wide range of forecasting tasks, our Hyper-Tree framework is also well-aligned with operational time series data and forecasting \citep{Januschowski.2019}, which typically involves high-dimensional feature sets that extend beyond basic time series information, including a mix of continuous and categorical features.

\subsection{Gradient Boosted Decision Trees} \label{sec:gbdt}

\begin{figure}[h!]
	\centering
	\caption{Conventional GBDT architecture showing feature input, decision tree processing, output generation, and loss calculation, with backward pass for gradient-based optimization.}
	\includegraphics[width=0.55\linewidth]{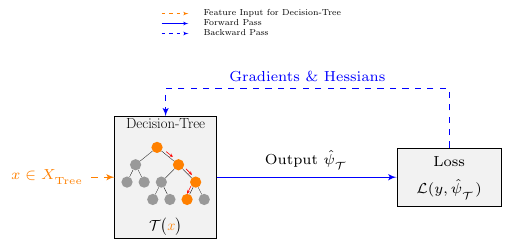}
	\label{fig:gbdt}
\end{figure}

While gradient-based parameter estimation is well-established for neural networks, using GBDTs to learn the parameters of a target time series model is less common and perhaps non-obvious. To better understand how GBDTs can be repurposed for learning parameters of target time series models, we briefly review their gradient-based nature. Modern GBDT implementations such as XGBoost \citep{Chen.2016}, LightGBM \citep{Ke.2017}, and CatBoost \citep{Prokhorenkova.2018} rely on gradients $g_{i}$ and Hessians $h_{i}$, where

\begin{equation}
	g_{i} = \frac{\partial \mathcal{L}\bigl(y_{i}, \hat{\psi}_{i_{\scriptscriptstyle \mathcal{T}}}\bigr)}{\partial \bigl(\hat{\psi}_{i_{\scriptscriptstyle \mathcal{T}}}\bigr)}, \quad h_{i} = \frac{\partial^{2} \mathcal{L}\bigl(y_{i}, \hat{\psi}_{i_{\scriptscriptstyle \mathcal{T}}}\bigr)}{\partial \bigl(\hat{\psi}_{i_{\scriptscriptstyle \mathcal{T}}}\bigr)^{\!2}}
\end{equation}

are the first and second order derivatives of a loss function $\mathcal{L}$ with respect to the output $\hat{\psi}_{i_{\scriptscriptstyle \mathcal{T}}}$ of tree $\mathcal{T}$, for observations $i = 1, \ldots, N$. Among others, gradients and Hessians serve two crucial functions in each boosting iteration: they guide the tree construction via split decisions and determine the optimal values assigned to leaf nodes via

\begin{equation}
	w^{*}_{j} = - \frac{G_{j}}{H_{j} + \lambda}, \quad \text{with} \quad 
	G_{j} = \sum_{i \in I_{j}} g_{i}, \quad 
	H_{j} = \sum_{i \in I_{j}} h_{i} 
\end{equation}

where $I_{j} = \{i|q(x_{i})=j\}$ is the set of indices of observations assigned to the $j$-th leaf, $q(\cdot)$ being the learned tree structure that maps an input $x$ to its corresponding leaf $j$, and $\lambda$ is a regularization term \citep{Chen.2016}. For split decisions, a greedy approach is used that maximizes the loss reduction

\begin{equation}
	\mathcal{L}_{split} \propto \frac{(\sum_{i\in I_L} g_i)^2}{\sum_{i\in I_L} h_i + \lambda} + \frac{(\sum_{i\in I_R} g_i)^2}{\sum_{i\in I_R} h_i + \lambda} - \frac{(\sum_{i\in I} g_i)^2}{\sum_{i\in I} h_i + \lambda} 
\end{equation}

where $I_L$ and $I_R$ denote the instance sets of left and right nodes after a candidate split respectively and $I = I_L \cup I_R$ represents their union \citep{Chen.2016}. The use of gradient and Hessian information, as illustrated in Figure~\ref{fig:gbdt}, provides GBDTs with considerable flexibility and enables their reformulation for parameter learning, which we outline in more detail below.

\newpage

\subsection{Hyper-Tree Architecture} \label{sec:hta}

Conventional GBDTs operate in function space, mapping input features $x$ to outputs $\hat{\psi}_{\scriptscriptstyle \mathcal{T}}$ by minimizing a specified loss function. For instance, when using Mean Squared Error (MSE) in a regression setting, $\hat{\psi}_{\scriptscriptstyle \mathcal{T}}$ represents the conditional mean of the target variable. While the use of an $\mathcal{L}_{\scriptscriptstyle 2}$-type loss may suggest that GBDTs create outputs by directly averaging target values, they instead operate through gradient and Hessian-based updates of the following form\footnote{For the specific case of MSE loss, the gradient is proportional to the residuals, and the Hessian is constant. This can lead to the impression that GBDTs simply fit residuals, when in fact they always follow gradient-based updates derived from the specified loss function.}

\begin{equation}
	\hat{\psi}^{(m)}_{i_{\scriptscriptstyle \mathcal{T}}}(x_i) = \hat{\psi}^{(m-1)}_{i_{\scriptscriptstyle \mathcal{T}}}(x_i) + \hat{\delta}^{(m)}_{i_{\scriptscriptstyle \mathcal{T}}}(x_i), \quad \text{where} \quad \hat{\delta}^{(m)}_{i_{\scriptscriptstyle \mathcal{T}}}(x_i) = \nu \cdot w^{*}_{j(i)}   
\end{equation}

where $\hat{\psi}^{(m)}_{i_{\scriptscriptstyle \mathcal{T}}}$ denotes the output after $m=1, \ldots, M$ iterations, $\hat{\delta}^{(m)}_{i_{\scriptscriptstyle \mathcal{T}}}$ is the incremental update at iteration $m$, $\nu$ is the learning rate, and $w^{*}_{j(i)}$ is the weight assigned to the leaf $j(i)$ corresponding to observation $i$, where the leaf assignment is determined by the feature vector $x_i$ through the structure of the learned tree, i.e., $j(i) = q(x_i)$. The leaf weights approximate a Newton-Raphson update \citep{Sigrist.2021}, where $w^{*}_{j} = -\tfrac{G_j}{H_j + \lambda}$ aggregates individual gradients and Hessians across all observations in leaf $j$. This update step makes GBDTs highly adaptive for a wide range of tasks beyond conventional forecasting, allowing $\hat{\psi}_{\scriptscriptstyle \mathcal{T}}$ to represent any quantity as long as the associated loss function is twice continuously differentiable. 

\begin{figure}[h!]
	\centering
	\caption{Hyper-Tree architecture illustrating a unified framework where a Hyper-Tree generates parameters for a target model. The output of the target model is passed to a loss function, with gradients and Hessians flowing back, enabling learning of temporal dependencies and integration of diverse feature types.}
	\includegraphics[width=0.85\linewidth]{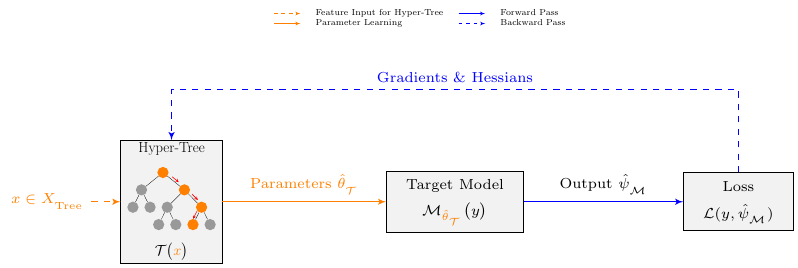}
	\label{fig:hyper_tree}
\end{figure}

By conceptualizing GBDTs as Hyper-Trees, we shift from performing gradient descent in function space, where the tree output $\hat{\psi}_{\scriptscriptstyle \mathcal{T}}$ is a forecasted value, to gradient descent in parameter space, where the tree output represents parameters of a target time series model, $\hat{\psi}_{\scriptscriptstyle \mathcal{T}} \equiv \hat{\theta}_{\scriptscriptstyle \mathcal{T}}$. In our framework, Hyper-Trees are trained by solving
\begin{equation}
	\min_{\hat{\theta}_{\scriptscriptstyle \mathcal{T}}} \; \mathcal{L}\Bigl(y,\, \mathcal{M}_{\hat{\theta}_{\scriptscriptstyle \mathcal{T}}}(y)\Bigr)
	\label{eq:hypertree_objective}
\end{equation}
where the loss $\mathcal{L}$ is evaluated on the forecast $\mathcal{M}_{\hat{\theta}_{\scriptscriptstyle \mathcal{T}}}(y)=\hat{\psi}_{\scriptscriptstyle \mathcal{M}}$ of the target time series model $\mathcal{M}_{\hat{\theta}_{\scriptscriptstyle \mathcal{T}}}$. Since $\hat{\psi}_{\scriptscriptstyle \mathcal{M}}$ is a composition of the target model and the tree-generated parameters, gradients and Hessians are obtained by applying the chain rule through $\mathcal{M}_{\hat{\theta}_{\scriptscriptstyle \mathcal{T}}}$. This requires the target model $\mathcal{M}_{\hat{\theta}_{\scriptscriptstyle \mathcal{T}}}$ to be twice continuously differentiable with respect to its parameters, as the GBDT update relies on both first- and second-order derivatives. For the $j$-th parameter estimated by the Hyper-Tree and observations $i = 1, \ldots, N$, the gradient and Hessian are given by
\begin{equation}
	g_i^{(j)} = \frac{\partial\mathcal{L}\bigl(y_i,\, \mathcal{M}_{\hat{\theta}_{i_{\scriptscriptstyle \mathcal{T}}}}(y_i)\bigr)}{\partial\bigl(\hat{\theta}^{(j)}_{i_{\scriptscriptstyle \mathcal{T}}}\bigr)}, \quad h_i^{(j)} = \frac{\partial^2\mathcal{L}\bigl(y_i,\, \mathcal{M}_{\hat{\theta}_{i_{\scriptscriptstyle \mathcal{T}}}}(y_i)\bigr)}{\partial\bigl(\hat{\theta}^{(j)}_{i_{\scriptscriptstyle \mathcal{T}}}\bigr)^{\!2}} \label{eq:hypertree_gradient_hessian}
\end{equation}
Consistent with the standard multi-output GBDT strategy, a separate tree is grown for each parameter, and only the diagonal entries of the Hessian matrix are required. Due to its modularity, where both the Hyper-Tree and the target model can be modified as required by the specific characteristics of the data, gradients and Hessians typically do not have an analytical solution. We therefore leverage the automatic differentiation capabilities as available in PyTorch \citep{Paszke.2019} for deriving gradients and Hessians. The architectural design of our approach is illustrated in Figure~\ref{fig:hyper_tree} with the Hyper-Tree as the central component.

The Hyper-Tree $\mathcal{T}$ takes a set of input features $x$ and generates parameters $\hat{\theta}_{\scriptscriptstyle \mathcal{T}}(x)$ that are used by a target model $\mathcal{M}$ to generate a forecast $\hat{\psi}_{\scriptscriptstyle \mathcal{M}}$ which is passed to a loss function $\mathcal{L}$. Gradients and Hessians, derived from the loss function with respect to the target model parameters, guide the Hyper-Tree to learn parameters optimized to improve the accuracy of the target time series model. This architecture enables integrated training and ensures that the loss incorporates information from both components: the parameters generated by the Hyper-Tree and the forecasts created by the target model. For the architecture shown in Figure \ref{fig:hyper_tree}, the temporal component can, for example, be modeled via an AR($p$) process, though the target model $\mathcal{M}$ may belong to a broad class of time series models. The input $x$ may include a diverse range of categorical and numerical features, extending beyond basic time series information. Unlike conventional time series models that learn directly from data, the target model in this framework does not. Instead, it acts as a fixed structure that imposes an inductive bias, guiding the Hyper-Tree $\mathcal{T}$ to encode temporal dependencies into the parameters it generates. This happens through gradient-based feedback, allowing the Hyper-Tree to generate time- and feature-specific parameters $\hat{\theta}_{\scriptscriptstyle \mathcal{T}}(x)$. In this sense, the target model acts more as a function applying the parameters generated by the Hyper-Tree, rather than learning the temporal dependencies on its own. By reformulating well-established GBDTs in this way, the Hyper-Tree architecture combines the strengths of tree-based learning of target-model parameters with classical time series models, thereby extending GBDTs to a time series context more naturally. Notably, because Hyper-Trees operate in parameter space by learning the parameters of a target time series model, which then creates the final forecasts, they can extrapolate beyond the value range of the training data, overcoming a key limitation of standard tree-based forecasting.

The modularity of our framework enables researchers and practitioners to easily customize the Hyper-Tree framework to their specific contexts, including probabilistic extensions where, instead of $\hat{\psi}_{\scriptscriptstyle \mathcal{M}}$ being a point-forecast, the Hyper-Tree can be used to model and forecast parameters of a distribution, from which forecast intervals can be derived.\footnote{For extensions of tree-based models to a probabilistic setting, we refer the reader to \cite{Marz.2022b, Marz.2022, Marz.2019, Sprangers.2021, Hasson.2021, Duan.2020}.} In general, any suitable distribution that characterizes the data-generating process can be parameterized. Assuming, e.g., a Normal distribution $y_t \mid \mathbf{x}_t \sim N\bigl(\mu_t(\mathbf{x}_t), \sigma_t(\mathbf{x}_t)\bigr)$, the mean and standard deviation can be specified as
\begin{align}
	\mu_t(\mathbf{x}_t)    &= \sum_{j=1}^{p} \theta^{\mu}_{j,t}(\mathbf{x}_t)\, y_{t-j} \\
	\sigma_t(\mathbf{x}_t) &= \exp\bigl(\theta^{\sigma}_{t}(\mathbf{x}_t)\bigr)
\end{align} 

\begin{figure}[h!]
	\centering
	\caption{Distributional Hyper-Tree Architecture for a probabilistic framework, where the Hyper-Tree generates parameters for both a target model and an output distribution. The target model outputs the estimated mean $\hat{\mu}(\mathbf{x})$ while the Hyper-Tree directly estimates the standard deviation $\hat{\sigma}(\mathbf{x})$, enabling probabilistic forecasting.}
	\includegraphics[width=0.75\linewidth]{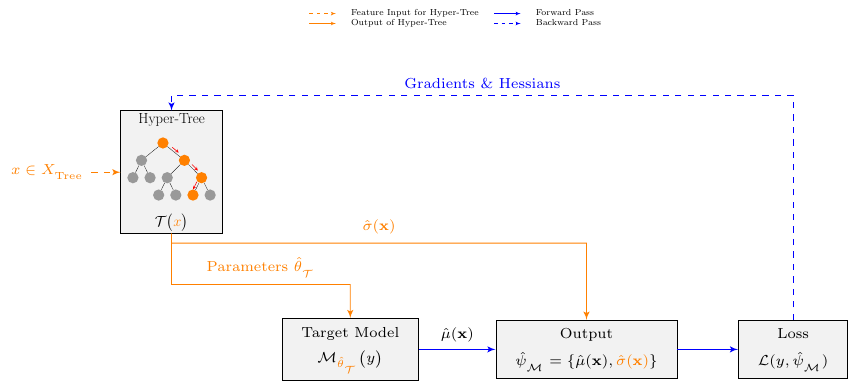}
	\label{fig:Hyper-Tree_dist}
\end{figure}

In this example, we parameterize $\mu_t(\mathbf{x}_t)$ via an AR($p$) model. While we can impose temporal structure on $\sigma_t(\mathbf{x}_t)$ as well, e.g., ARCH/GARCH models \citep{Engle.1982, Bollerslev.1986}, one can alternatively estimate $\sigma_t(\mathbf{x}_t)$ directly via the Hyper-Tree without using an intermediate target model as illustrated in Figure \ref{fig:Hyper-Tree_dist}. Instead of modeling parameters of a distribution, creating probabilistic forecasts via conformal predictive distributions \citep{Vovk.2022, Johansson.2023} or conformalized quantile regression \citep{Romano.2019} present interesting alternatives. For the scope of this paper, however, we focus on point forecasts, leaving probabilistic extensions for future research. 

\newpage

\begin{figure}[h!]
	\centering
	\caption{Scaling performance comparison between \texttt{Hyper-Tree-AR($p$)} and \texttt{Hyper-TreeNet-AR($p$)} models. The figure shows runtimes as the number of AR-parameters increases. All runtimes are normalized with respect to the runtime of estimating one target model parameter.}
	\includegraphics[width=0.65\linewidth]{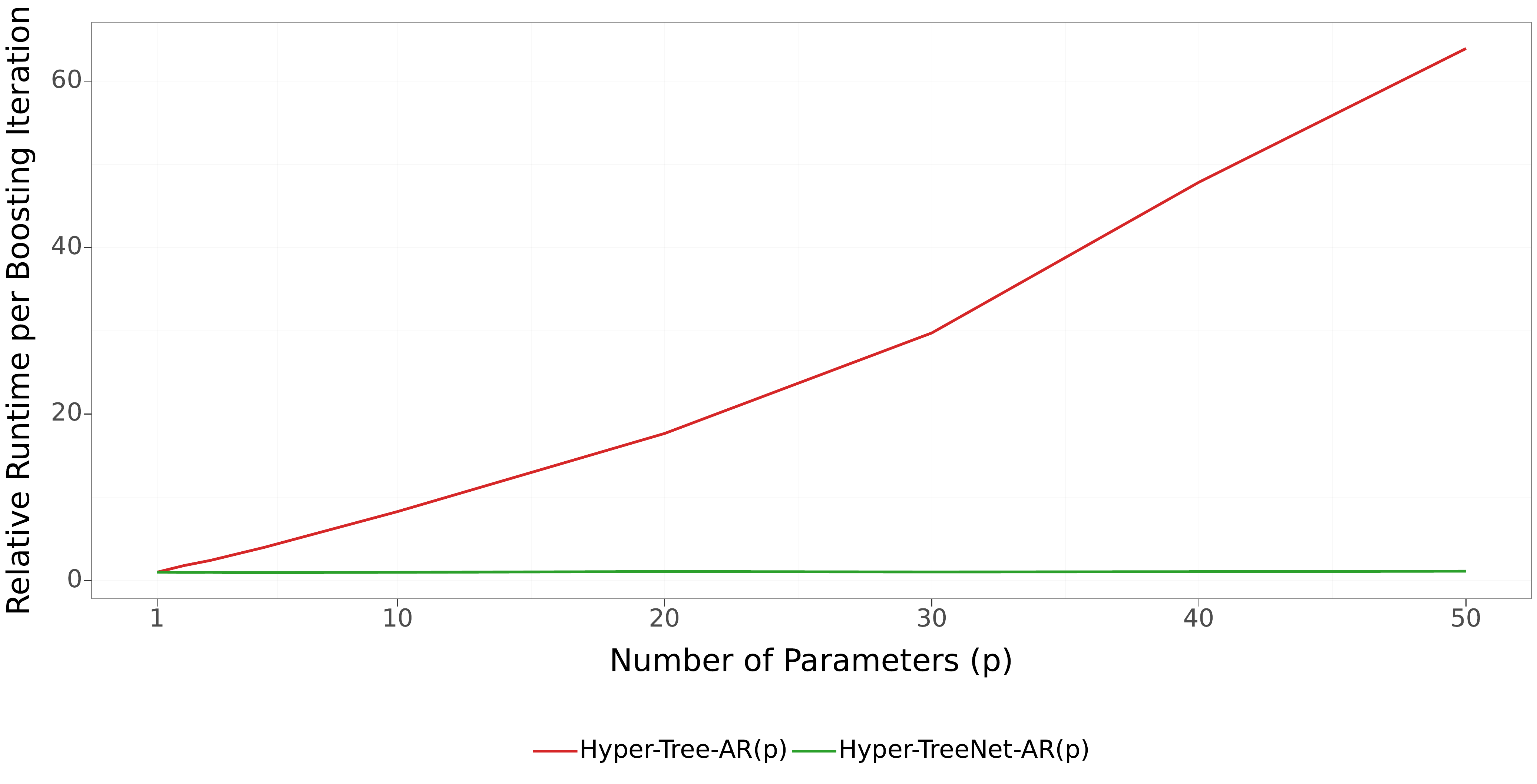}
	\label{fig:gbdt_scaling}
\end{figure}

To maintain computational efficiency, we propose a hybrid architecture that integrates GBDTs with neural networks, thereby addressing the scalability limitations of GBDTs when estimating a large number of target model parameters. We refer to this model as Hyper-TreeNet. Traditionally, GBDTs employ a one-vs-all strategy, growing a separate tree for each parameter. While efficient for a small number of parameters, this approach becomes computationally prohibitive as the number of target model parameters increases \citep{Iosipoi.2022}. The scaling issue is illustrated in Figure \ref{fig:gbdt_scaling}, where we plot the relative runtime per boosting iteration for a \texttt{Hyper-Tree-AR($p$)} model trained with the one-vs-all strategy, along with the proposed \texttt{Hyper-TreeNet-AR($p$)} architecture.

\begin{figure}[h!]
	\centering
	\caption{Hyper-TreeNet architecture illustrating a hybrid approach combining GBDTs and neural networks. A Hyper-Tree generates low-dimensional parameter-space embeddings, which are transformed by a Multi-Layer Perceptron (MLP) to generate parameters for a target model. The architecture allows for joint optimization, enabling integrated tree-based feature learning and network-based parameter mapping.}
	\includegraphics[width=0.9\linewidth]{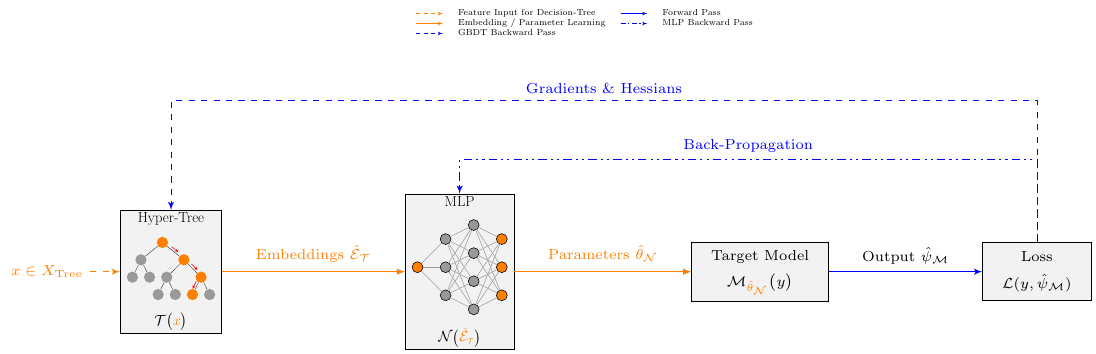}
	\label{fig:Hyper-TreeNet}
\end{figure}

Clearly, the \texttt{Hyper-TreeNet-AR($p$)} model maintains a nearly constant runtime regardless of the number of parameters. Our hybrid approach leverages GBDTs to generate low-dimensional embeddings, which are then transformed by a random projection matrix before serving as input to a shallow neural network, typically an MLP \citep{Rumelhart.1986}. The \texttt{Hyper-TreeNet-AR($p$)} architecture illustrated in Figure \ref{fig:Hyper-TreeNet} resembles an encoder-decoder framework, where the GBDT part acts as an encoder for the input features, and the MLP functions as a decoder, mapping the tree-embeddings to the target model parameters. Notably, the GBDT component generates embeddings in parameter rather than forecasting space. This ensures that both the tree embeddings and network outputs operate within the same space, facilitating more effective learning of time-varying parameters.

For all experiments, we use a one-dimensional tree-embedding to keep the runtime low.\footnote{While $d=1$ may appear restrictive, our analyses in \ref{app:embedding_dim} indicate that low-dimensional tree-embeddings yield competitive accuracy on most datasets.} To compensate for this low dimensionality and to enhance the learning capacity of the network, we incorporate a random projection matrix $\mathbf{W}_{\text{proj}}$ following the implementation of \cite{Yeh.2024}. This expansion diversifies the tree-representations and increases the input dimensions for the subsequent neural network without increasing computational complexity \citep{Yeh.2024, Iosipoi.2022, Casale.2011}. Specifically, we generate a matrix $\mathbf{W}_{\text{proj}} \in \mathbb{R}^{k \times d}$ where $d$ is the dimension of tree embeddings and $k$ is the target dimension. Each element of $\mathbf{W}_{\text{proj}}$ is sampled from a standard Normal distribution. The projection operation can be expressed as $\mathbf{z} = \mathbf{W}_{\text{proj}}\mathcal{E}_{_{\mathcal{T}}}$, where $\mathbf{\mathcal{E}_{_{\mathcal{T}}}} \in \mathbb{R}^{d \times N}$ represents the tree embeddings for all $i=1, \ldots, N$ observations and $\mathbf{z} \in \mathbb{R}^{k \times N}$ represents the expanded representation passed to subsequent layers of the MLP. From an implementation perspective, $\mathbf{W}_{\text{proj}}$ is a linear network layer with weights that are randomly initialized and then fixed during training, ensuring they remain constant during backpropagation. While higher-dimensional tree embeddings can be used as well, this comes at the cost of increased runtime. Our experiments indicate that a one-dimensional embedding combined with the random projection matrix offers a good trade-off between accuracy and computational efficiency. For all of our experiments, we use the tree-embeddings as the sole input to the MLP, rather than also including the original features, leveraging the GBDT's ability to effectively encode both categorical and numeric features. Our choice of an MLP is primarily motivated by its lightweight architecture and by evidence in the time series literature showing that MLP-type models are competitive with more complex architectures \citep{Zeng.2023, Chen.2023, Ekambaram.2024}. 

Both the GBDTs and the MLP are trained jointly, rather than in a two-step procedure. This joint learning enables GBDTs to create informative low-dimensional representations that neural networks can efficiently transform into high-dimensional parameter vectors. Throughout our experiments, we employ a shallow MLP structure following the sequence \texttt{$\{\text{Tree Embeddings} \rightarrow \text{Random Projection} \rightarrow \text{Hidden-Layer} \rightarrow \text{ReLU} \rightarrow \text{Output-Layer} \rightarrow \text{Dropout}\}$}.\footnote{Placing dropout after the output layer randomly zeros individual AR coefficients during training, acting as parameter regularization.} We maintain a fixed hidden dimension, dropout-rate, and learning rate for the MLPs across all datasets and experiments. We set the projection matrix dimensions $\mathbf{W}_{\text{proj}} \in \mathbb{R}^{k \times d}$ with input dimension $d=1$ to match our one-dimensional tree embeddings, and output dimension $k$ equal to the number of target model parameters. We experimented with various batching strategies for the Hyper-TreeNet model and found that training both components on the full dataset yields the best results. This is likely because GBDTs capture global patterns more effectively when exposed to all observations, producing more informative tree embeddings.

\newpage

Integrating GBDTs and neural networks in our Hyper-TreeNet architecture requires careful consideration of how gradients are propagated during training. GBDTs operate in a discrete optimization space with threshold-based splits, whereas neural networks rely on continuous gradient-based updates. We consider two strategies: a fully shared gradient flow (\textit{Option 1}) and a separated approach with per-iteration updates but distinct gradient flows (\textit{Option 2}). While \textit{Option 1} allows for joint optimization through a single backpropagation pass, it may limit the ability of each component to maintain its characteristic learning behavior. \textit{Option 2}, which is used as the default in our experiments, preserves the concept of integrated training between both components but avoids neural network gradients interfering with GBDT updates. The backward flows in Figure~\ref{fig:Hyper-TreeNet} use different dash patterns to indicate this separation. We empirically compare both options in our ablation studies (Section \ref{sec:ablation}), as the impact of gradient flow choice may vary across datasets and become more pronounced under specific conditions. More details on the two gradient flow options, including pseudocode and further analysis, are provided in \ref{sec:gradients}.

\section{Related Work} \label{sec:literature}

Recent advancements in the application of hyper-networks have yielded considerable contributions across various fields, reflecting an increasing interest. In the area of time series forecasting, \cite{Lee.2023} propose HyperGPA, a hyper-network designed to address the challenges of temporal drifts in time series data by generating time-varying parameters, yielding marked improvements in forecast accuracy. \cite{Duan.2022} introduce Hyper Time Series Forecasting (HTSF), a hyper-network-based framework that tackles distribution shift problems. HTSF jointly learns time-varying distributions and corresponding forecasting models, demonstrating state-of-the-art performance on several benchmarks. \cite{Fons.2022} explore the use of hyper-networks in Implicit Neural Representations (INRs) for time series data representation and analysis. Their work demonstrates the potential of INRs in time series imputation and generative tasks, showcasing competitive performance against existing approaches. To bridge classical and deep learning approaches, \cite{Rangapuram.2018} propose using RNNs to learn the parameters of linear state space models for each time series, combining the interpretability of state space formulations with the expressiveness of neural networks. Early work in this direction includes \cite{Medeiros.2000}, who introduce a neural coefficient smooth transition autoregressive model. Their approach employs a linear model with time-varying parameters controlled by a neural network, offering advantages over traditional approaches through the incorporation of linear multivariate thresholds and smooth transitions between regimes. In a related approach, \cite{Smyl.2020} introduces a hybrid method that integrates exponential smoothing with recurrent neural networks and demonstrates its effectiveness by winning the M4 forecasting competition \citep{Makridakis.2018}. In the econometrics literature, \cite{Primiceri.2005} introduces Time-Varying Structural Vector Autoregressions, which allow both coefficients and the variance-covariance matrix of innovations in vector autoregressive models to evolve over time. The study shows how time-varying parameters can capture changing economic relationships, particularly in monetary policy analysis where the transmission mechanisms between interest rates, inflation, and output may shift across different time periods.\footnote{Early seminal work on time-varying parameter models includes \cite{Belsley.1973, Belsley.1973b, Cooley.1973, Cooley.1976}.} 

Representing model parameters as functions of features has a counterpart in the statistics literature in the form of varying-coefficient models (VCMs) \citep{Hastie.1993}, where the coefficients are modeled as functions of covariates. In the time series literature, functional-coefficient autoregressive models \citep{Chen.1993, Cai.2000} estimate autoregressive coefficients as smooth functions of an index variable using kernel and local-linear methods. Along these lines, \citet{Yousuf.2021} estimate the coefficients of a high-dimensional predictive regression as smooth functions of time through $L_2$ boosting with componentwise local-constant or local-linear base learners, and establish consistency under local stationarity, noting that regression trees can serve as alternative base learners. On the tree-based side, \citet{Wang.2014} use vector-valued trees in a boosted VCM for product-demand prediction, \citet{Zhou.2022} grow per-coefficient gradient-boosted ensembles with an accompanying consistency guarantee, and \citet{Berger.2019} and \citet{Buergin.2017} employ single recursive-partitioning trees, while \citet{Puth.2020} extend the former to time-varying coefficients in discrete time-to-event models. More recently, \citet{Zakrisson.2025} model the varying coefficients with a cyclic gradient boosting machine \citep{Delong.2023}, and \citet{Spuck.2025} develop selective-inference confidence intervals for tree-structured varying coefficients.

To harness the complementary strengths of tree-based methods and neural networks, there has been growing interest in hybrid models that integrate the two frameworks. In their approaches, \cite{Karingula.2022} and \cite{Badirli.2020} develop variants of gradient boosting that use deep neural networks as weak learners. In a parallel line of research, \cite{Emami.2022} explore training the final layers of a neural network using gradient boosting, whereas \cite{Kontschieder.2016} propose integrating differentiable decision trees into neural networks for end-to-end training. Building on and generalizing the work of \cite{Prokhorenkova.2018}, \cite{Popov.2020} introduce NODE, a deep learning architecture for tabular data that reformulates oblivious decision tree ensembles as differentiable layers. NODE supports end-to-end gradient-based optimization and hierarchical representation learning, enabling integration into standard deep learning frameworks and allowing for multi-layer architectures akin to deep, end-to-end trained GBDTs. Related to this is the approach of \cite{Horrell.2025} that implements GBDTs as layers within the PyTorch framework of \cite{Paszke.2019}. For a more exhaustive overview of combining GBDTs with deep learning, we refer to \cite{Li.2023}.

\section{Experiments} \label{sec:experiments}

For our experiments, we use the following commonly used publicly available datasets:

\begin{itemize}[itemsep=-2pt]
	\item \texttt{Air Passengers}: This dataset contains the monthly totals of international airline passengers and is taken from \cite{Alexandrov.2020, Athanasopoulos.2011}.
	\item \texttt{Australian Electricity Demand}: This dataset comprises five time series, each representing the electricity demand for five Australian states, and is borrowed from \cite{Godahewa.2021}. Originally recorded at 30-minute intervals, we aggregate the data to a monthly time granularity for our analysis. This aggregation results in some months with incomplete data (when the original 30-minute observations do not span the entire month), which we exclude from the test set to ensure evaluation on complete months only.
	\item \texttt{Australian Retail Turnover}: This dataset consists of a total of 133 monthly series and is borrowed from \cite{OHaraWild.2022}. 		
	\item \texttt{M3 Monthly}: This dataset comprises 1,428 monthly time series, belonging to six distinct domains: demographic, microeconomic, macroeconomic, industry, finance, and other categories, and is borrowed from \cite{Makridakis.2000, Godahewa.2021}. 	
	\item \texttt{M3 Yearly}: This dataset comprises 645 yearly time series, belonging to six distinct domains: demographic, microeconomic, macroeconomic, industry, finance, and other categories, and is borrowed from \cite{Makridakis.2000, Godahewa.2021}. 	
	\item \texttt{M5}: This dataset comprises daily sales records from Walmart, the world's largest retailer by revenue. It encompasses 3,049 products across multiple stores and states, resulting in 42,840 distinct time series due to its multi-level hierarchical structure \citep{Makridakis.2022, Makridakis.2022b}. We aggregate the data to the store-department level, yielding 70 daily time series. This aggregation is motivated primarily to reduce the intermittency present at the most granular level of the data. 
	\item \texttt{Rossmann Store Sales}: This dataset consists of a total of 1,115 daily series and is borrowed from \cite{Knauer.2015}.	We subset the data to include only stores that are open. 	
	\item \texttt{Tourism}: This dataset consists of a total of 366 monthly series and is borrowed from \cite{Alexandrov.2020, Athanasopoulos.2011}. 
\end{itemize}

\noindent As references, we compare our Hyper-Trees to the following models:

\begin{itemize}[itemsep=-2pt]
	\item \texttt{AR($p$)}: An autoregressive model with $p$ lags.
	\item \texttt{AR($p$)-X}: same as \texttt{AR($p$)} with additional features.
	\item \texttt{AutoArima}: An ARIMA \citep{Box.2015} model where optimal $(p,d,q)(P,D,Q)$ are selected.
	\item \texttt{AutoArima-X}: same as \texttt{AutoArima} with additional features.	
	\item \texttt{AutoETS}: An Exponential Smoothing \citep{Holt.2004, Winters.1960} model, where the model specification is selected automatically.
	\item \texttt{Chronos}: Chronos is a family of pre-trained forecasting models based on language model architectures \citep{Ansari.2024}. For all experiments, we use the \texttt{chronos-t5-base} version.	
	\item \texttt{Deep-AR}: A probabilistic forecasting model that uses autoregressive recurrent neural networks \citep{Salinas.2020}.	
	\item \texttt{LightGBM}: LightGBM model of \cite{Ke.2017}.
	\item \texttt{LightGBM-AR($p$)}: Same as \texttt{LightGBM} with additional autoregressive lagged-target values of order $p$ as features.
	\item \texttt{LightGBM-STL}: Each series is first de-trended and de-seasonalized via a cubic polynomial. The remainder is then modeled via a local LightGBM \citep{Ke.2017} model. 	
	\item \texttt{Temporal Fusion Transformer (TFT)}: An attention-based architecture for multi-horizon time series forecasting \citep{Lim.2021}.	
\end{itemize}

Classical time series models are trained locally on each series and evaluated on datasets where only time-derived features (e.g., month, quarter) are available. Due to their cross-learning capabilities, as well as their ability to incorporate a richer feature set including categorical and numeric variables alongside temporal information, deep-learning-based models are used for global model training and evaluation only. All local forecasting models are trained using the Nixtla-implementation of \cite{Garza.2022}, whereas the deep-learning models are trained via the implementations available in GluonTS of \cite{Alexandrov.2020, Athanasopoulos.2011}. The \texttt{LightGBM-STL} model is trained using the sktime implementation of \cite{Kiraly.2024, Loning.2019}. To ensure a practical comparison across models, we standardized hyper-parameters as much as possible across all models for each specific dataset. Following real-world practices where exhaustive tuning is often impractical due to time or computational constraints, we opted not to perform extensive hyper-parameter optimization and used mostly default configurations. We refer to Tables \ref{tab:local_hyperparams} and \ref{tab:global_hyperparams} for details on the hyper-parameters. It is also important to note that, unlike conventional tree-based time series models, Hyper-Trees exclude all target-derived features, including lags, moving averages, and other transformations of the target variable, from their feature set, relying instead on the parameterized target time series model to capture temporal dynamics. For our experiments, we start with a simple illustrative example in Section \ref{sec:local_single} and continue with an evaluation across multiple datasets in Sections \ref{sec:local_multiple} and \ref{sec:global}. Owing to its proven accuracy and popularity in the time series community, we use LightGBM \citep{Ke.2017} as the GBDT for our Hyper-Tree models across all experiments.

\subsection{Local Time Series Model: Introductory Example} \label{sec:local_single}

\begin{figure}[h!]
	\centering
	\caption{\texttt{Hyper-Tree-STL} based Decomposition.}
	\begin{subfigure}{0.5\textwidth}
		\centering
		\caption{Estimated Trend}
		\includegraphics[width=0.60\linewidth]{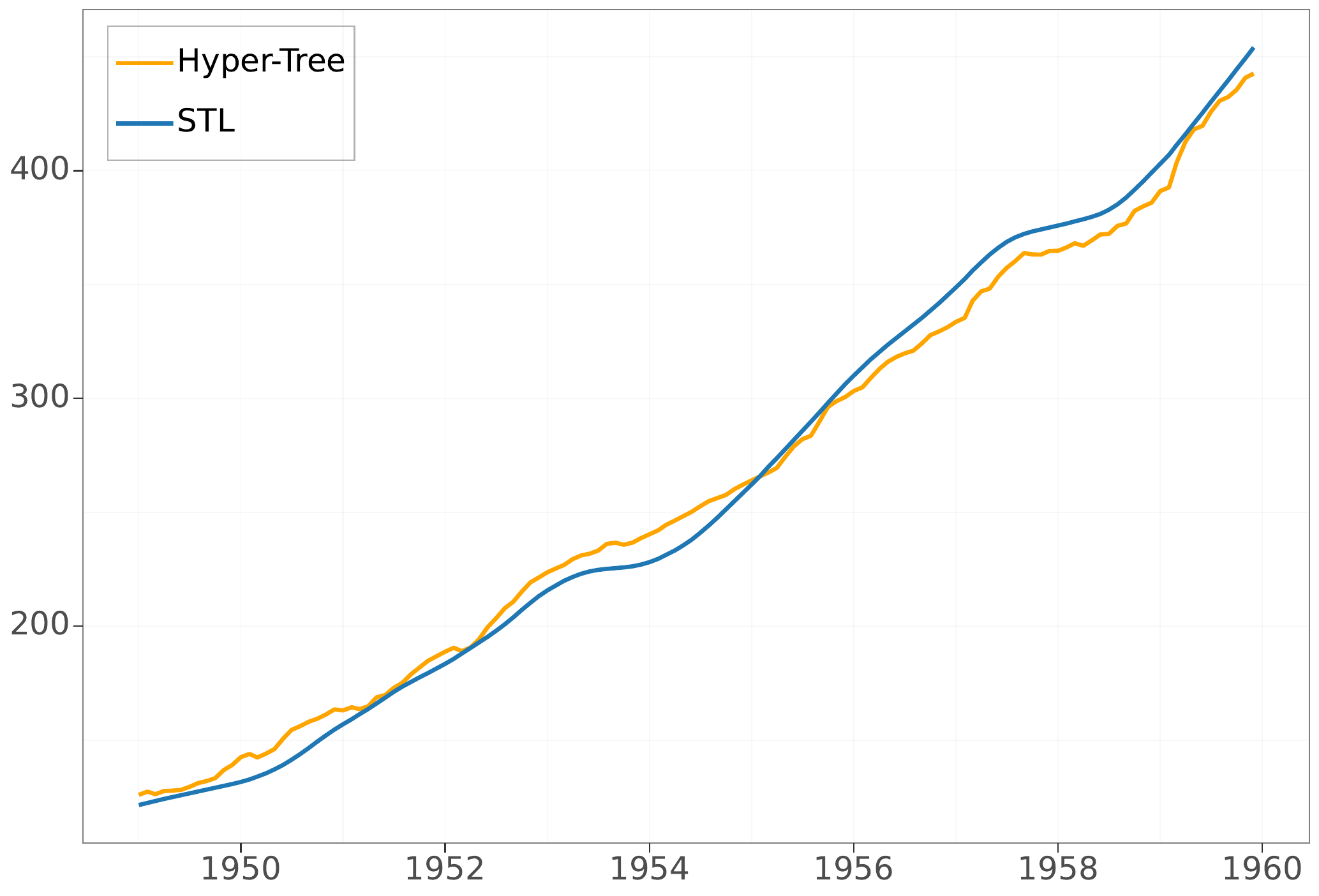}
		\label{fig:stl_trend}
	\end{subfigure} \hspace{-4em}  %
	\begin{subfigure}{0.5\textwidth}
		\centering
		\caption{Estimated Seasonality}
		\includegraphics[width=0.60\linewidth]{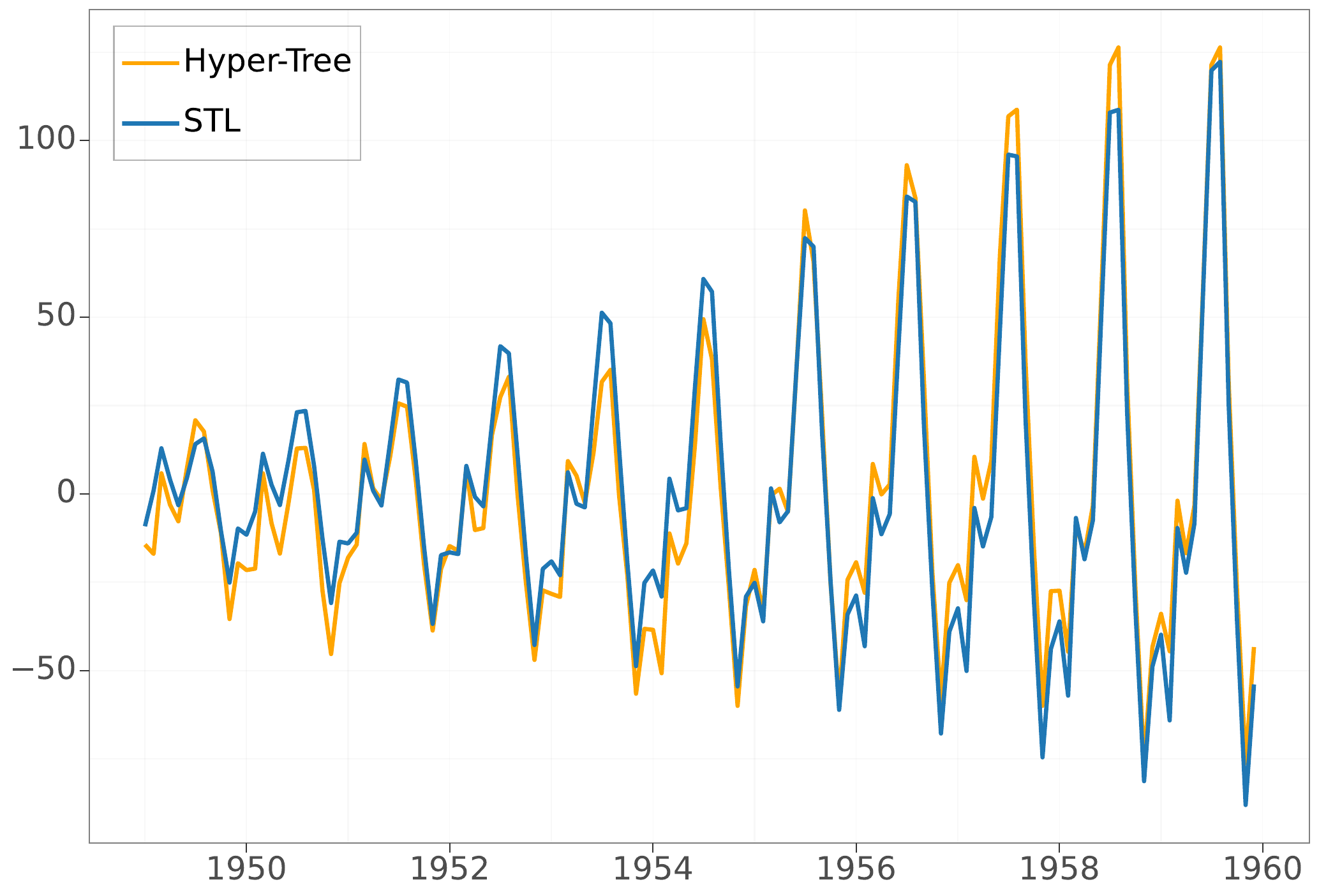}
		\label{fig:stl_seasonality}
	\end{subfigure}
	\label{fig:stl}
\end{figure}

We start our analysis with the \texttt{Air Passengers} dataset, which consists of monthly totals of airline passengers from 1949 to 1960. It is a popular dataset in time series forecasting and its trend and pronounced seasonality characteristics make it particularly suitable for the ARIMA \citep{Box.2015} to be a strong baseline. We split the data into train and test, where the last year is used as a hold-out set. The first step for analyzing such a time series dataset is typically a decomposition into seasonality, trend, and remainder. This STL decomposition is usually done via a LOESS-based estimation of each component \citep{Cleveland.1990}. However, in the following, we use our Hyper-Tree architecture for decomposing the train series. The \texttt{Hyper-Tree-STL} parameterizes the following trend and seasonality components
\begin{align}
	\text{Trend}_{t} &= a_{\scriptscriptstyle 0,t}(\mathbf{x}_{\scriptscriptstyle t}) + a_{\scriptscriptstyle 1,t}(\mathbf{x}_{\scriptscriptstyle t}) \cdot t \\
	\text{Seasonality}_{t} &= \sum_{r=1}^{N_{\text{season}}} \Biggl( c_{\scriptscriptstyle r,t}(\mathbf{x}_{\scriptscriptstyle t}) \cdot \sin\left(t \cdot r \cdot \tfrac{2\pi}{m_{\text{season}}}\right) + d_{\scriptscriptstyle r,t}(\mathbf{x}_{\scriptscriptstyle t}) \cdot \cos\left(t \cdot r \cdot \tfrac{2\pi}{m_{\text{season}}}\right) \Biggr)
\end{align}

where the parameters $\theta_{\scriptscriptstyle \mathcal{T}}(\mathbf{x}_{\scriptscriptstyle t}) = \bigl[a_{\scriptscriptstyle 0,t}(\mathbf{x}_{\scriptscriptstyle t}), a_{\scriptscriptstyle 1,t}(\mathbf{x}_{\scriptscriptstyle t}), \{c_{\scriptscriptstyle r,t}(\mathbf{x}_{\scriptscriptstyle t}), d_{\scriptscriptstyle r,t}(\mathbf{x}_{\scriptscriptstyle t})\}_{r=1}^{N_{\text{season}}}\bigr]$ are modeled as functions of time-derived features $\mathbf{x}_{_{t}} = \{\text{month}, \text{quarter}, \text{year}, \text{time}\}$, with $\text{time}$ being a linearly increasing integer. Even though the trend part is specified as a linear function of $t$, the interactions and non-linearities induced by the Hyper-Tree allow for modeling complex patterns. To ensure the trend is a smooth function of time, we add a term that penalizes squared first and second order differences, thereby encouraging neighboring estimates to be close to each other. We represent the seasonality via periodic Fourier-terms that are composed of sine and cosine components, each with different frequencies and amplitudes, where $\{c_{\scriptscriptstyle r,t}(\mathbf{x}_{\scriptscriptstyle t}), d_{\scriptscriptstyle r,t}(\mathbf{x}_{\scriptscriptstyle t})\}_{r=1}^{N_{\text{season}}}$ represent the sine and cosine weights for the $r$-th seasonal component. For this experiment, we set $N_{\text{season}}=1$ and $m_{\text{season}}=12$ since the data is recorded on a monthly frequency. Instead of tuning the hyper-parameters, we mostly use default values and the MSE as the loss function. Figure \ref{fig:stl} shows the trend and seasonality estimated on the training dataset. As a reference, we add the components of a conventional STL decomposition as available in \cite{Seabold.2010}.

\begin{figure}[h!]
	\centering
	\caption{Estimated Parameters of \texttt{Hyper-Tree-STL}.}
	\begin{subfigure}{0.5\textwidth}
		\centering
		\caption{$\hat{a}_{\scriptscriptstyle 0,t}$}
		\includegraphics[width=0.6\linewidth]{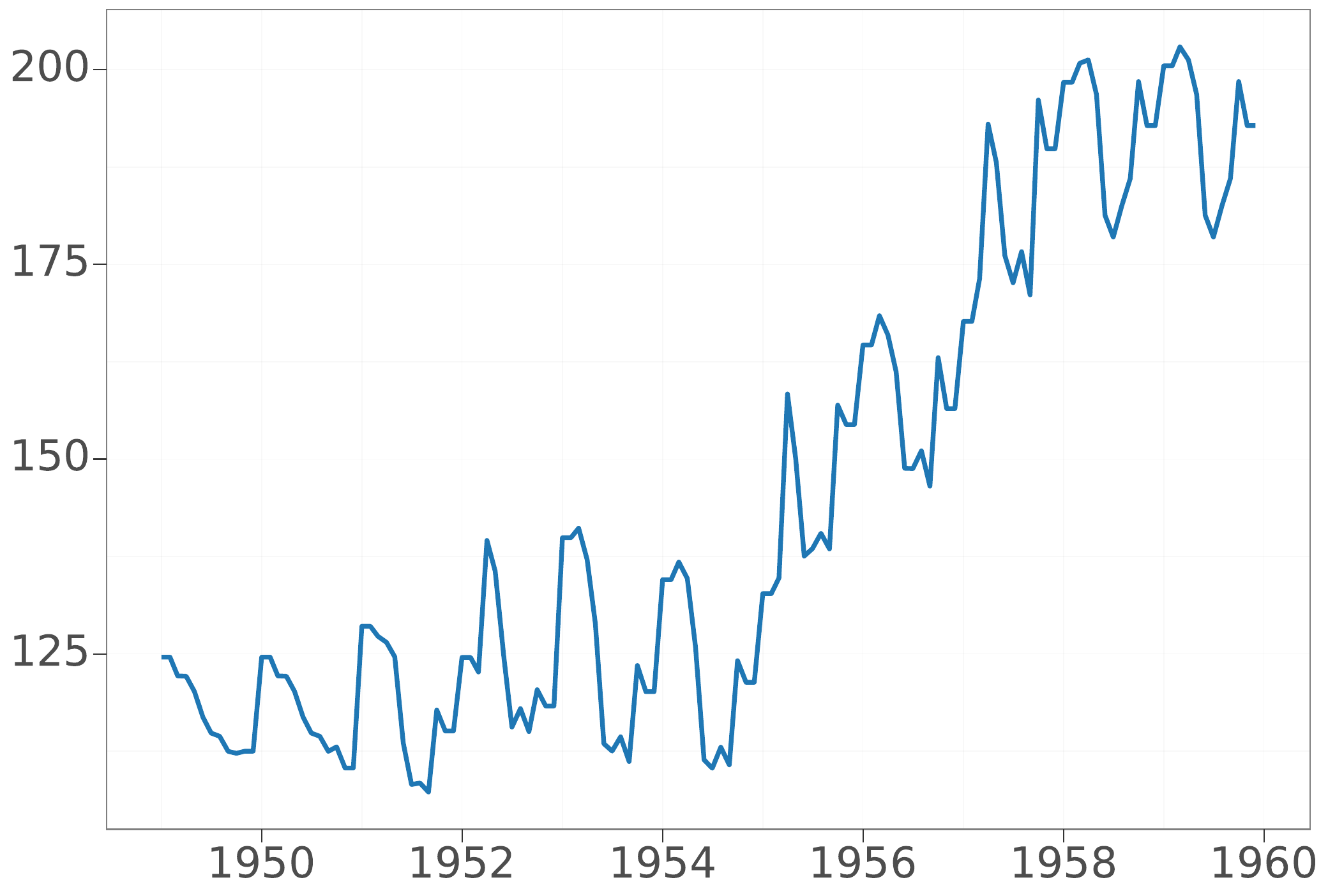}		
		\label{fig:beta0}
	\end{subfigure} \hspace{-8em}  %
	\begin{subfigure}{0.5\textwidth}
		\centering
		\caption{$\hat{a}_{\scriptscriptstyle 1,t}$}
		\includegraphics[width=0.6\linewidth]{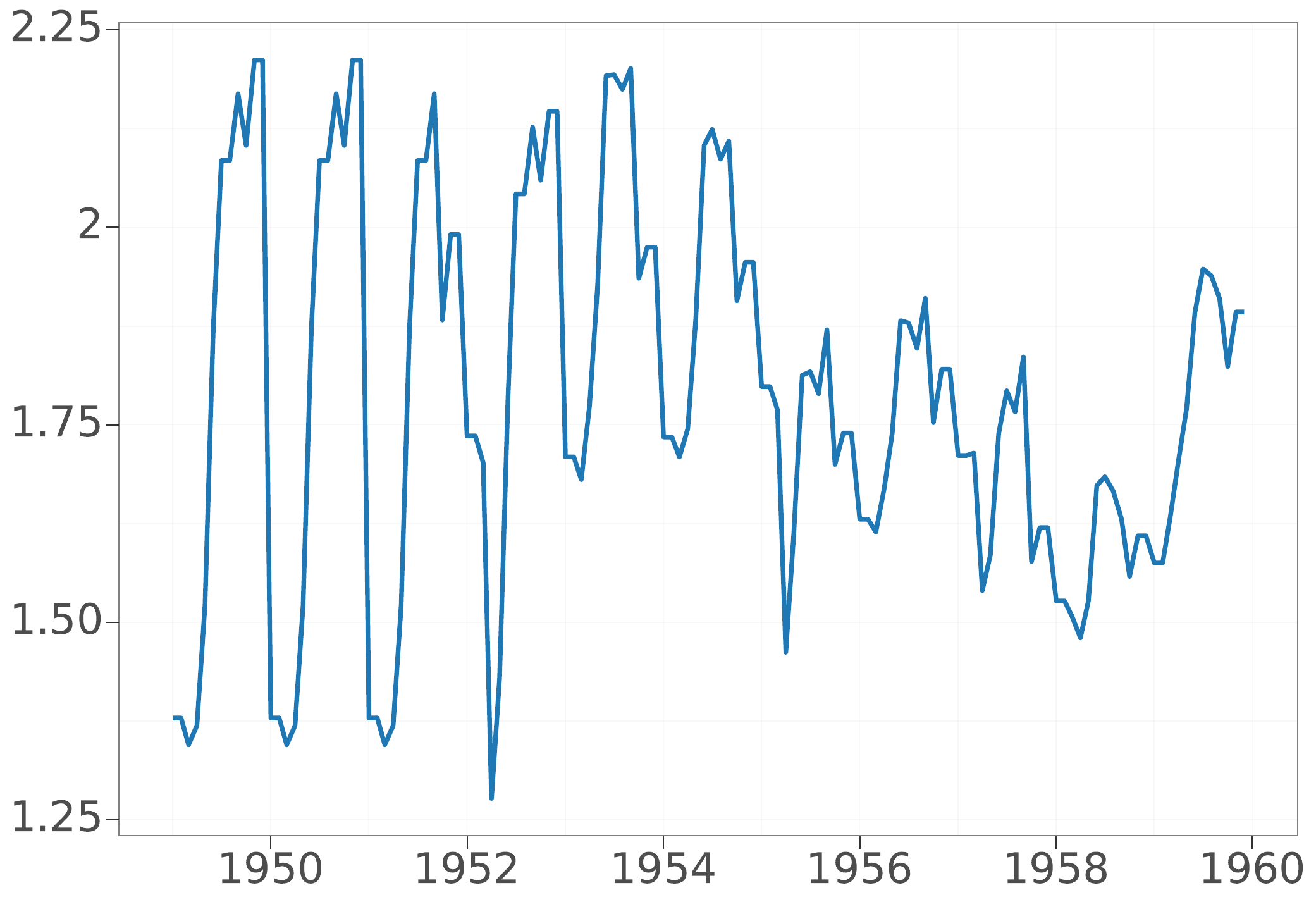}	
		\label{fig:beta1}
	\end{subfigure}
	\begin{subfigure}{0.5\textwidth}
		\centering
		\caption{$\hat{c}_{\scriptscriptstyle 1,t}$}
		\includegraphics[width=0.6\linewidth]{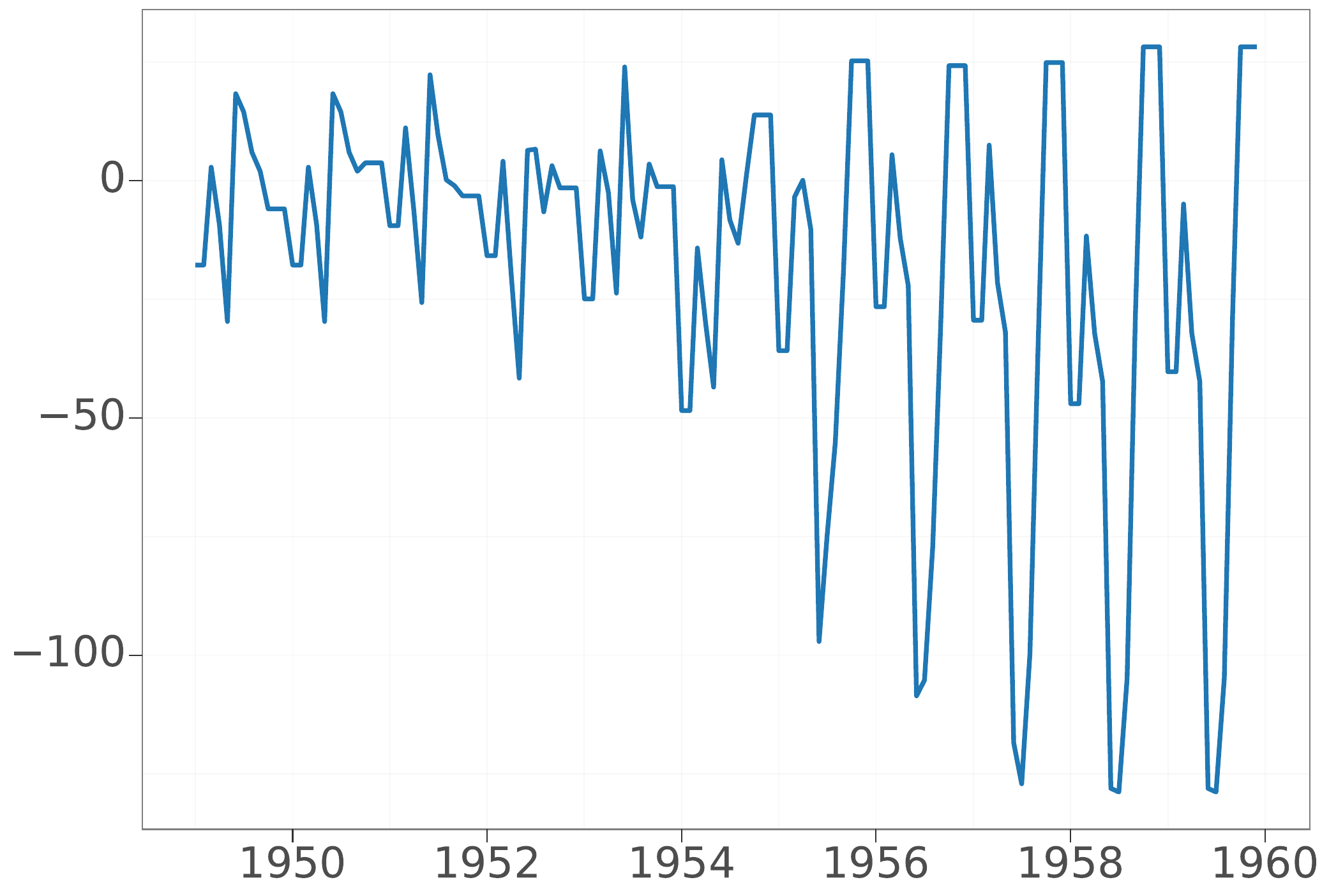}
		\label{fig:sine}
	\end{subfigure} \hspace{-8em}  %
	\begin{subfigure}{0.5\textwidth}
		\centering
		\caption{$\hat{d}_{\scriptscriptstyle 1,t}$}
		\includegraphics[width=0.6\linewidth]{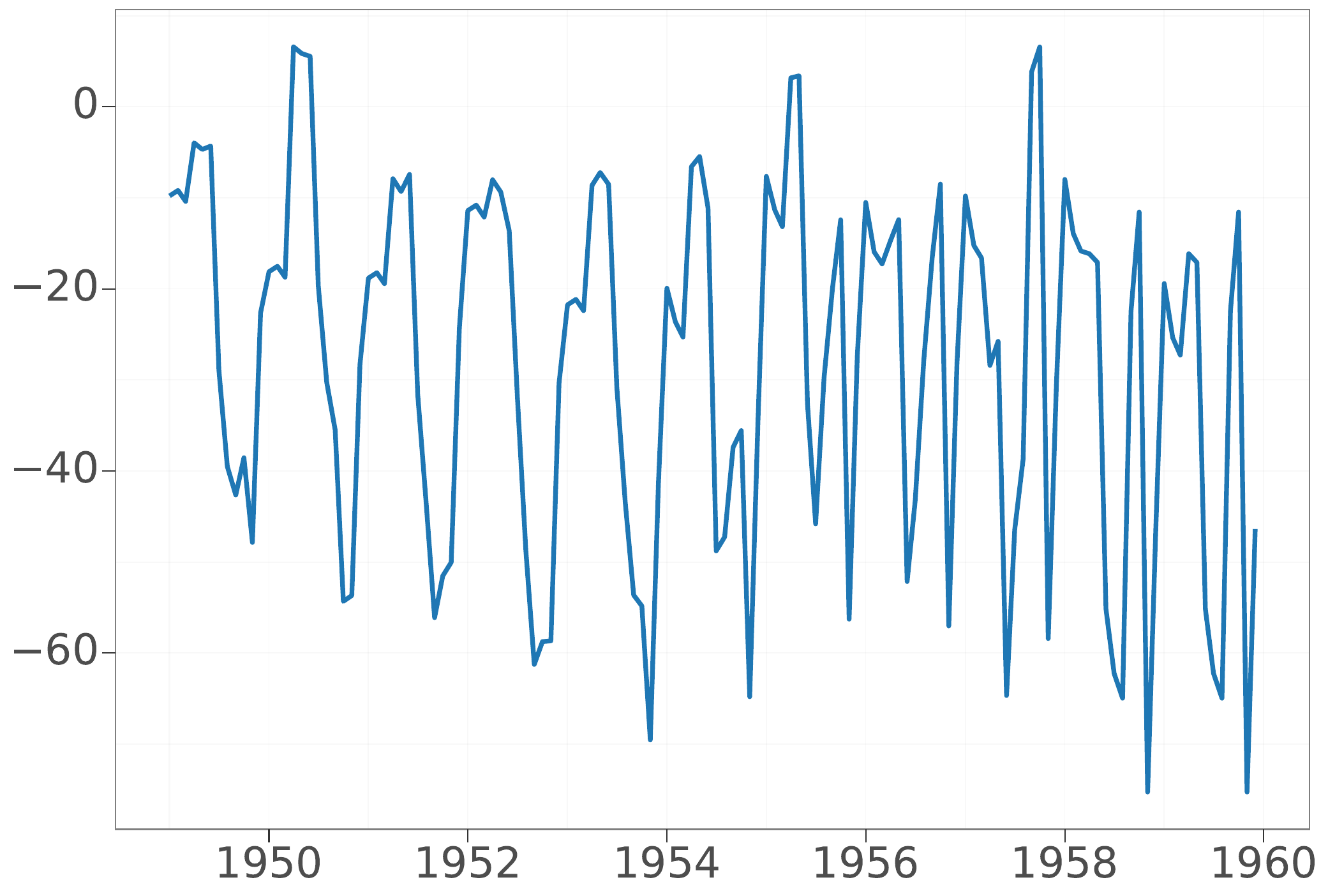}
		\label{fig:cosine}
	\end{subfigure}	
	\label{fig:stl_params}
\end{figure}

Figure \ref{fig:stl} shows that the STL components estimated via the \texttt{Hyper-Tree-STL} and the traditional LOESS-based decomposition are in close agreement, even though the trend-estimate of the \texttt{Hyper-Tree-STL} is not as smooth. Considering the current use of relatively simplistic elements for estimating trend and seasonality, their approximation quality could be enhanced by using more complex target models, e.g., via penalized splines \citep{Wood.2017}. In terms of model interpretability, we can visualize the estimated parameters over time, as illustrated in Figure \ref{fig:stl_params}.

Besides the estimated parameters, we also have access to SHAP values \citep{Lundberg.2020, Lundberg.2017} and feature importances for each estimated parameter, further increasing model interpretability.

\newpage

\begin{figure}[h!]
	\centering
	\caption{Feature Importance of \texttt{Hyper-Tree-STL} Parameters.}
	\begin{subfigure}{0.5\textwidth}
		\centering
		\caption{$\hat{a}_{\scriptscriptstyle 0,t}$}
		\includegraphics[width=0.65\linewidth]{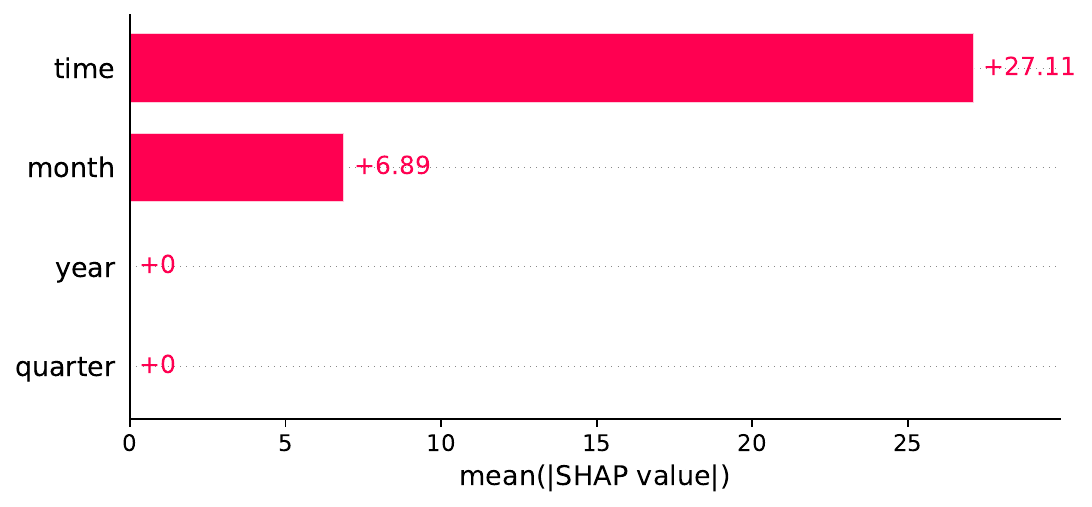}		
		\label{fig:imp_beta0}
	\end{subfigure} \hspace{-4em}  %
	\begin{subfigure}{0.5\textwidth}
		\centering
		\caption{$\hat{a}_{\scriptscriptstyle 1,t}$}
		\includegraphics[width=0.65\linewidth]{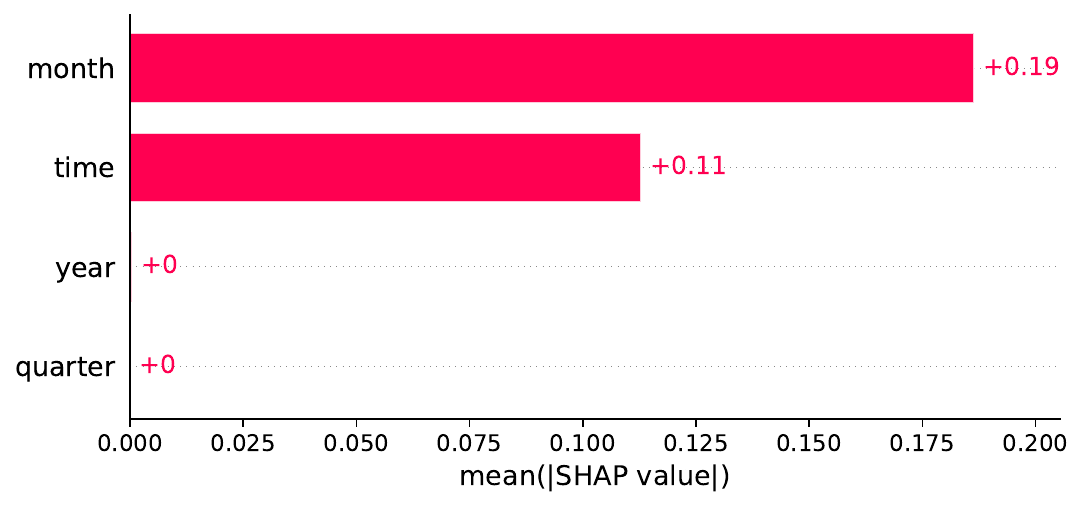}	
		\label{fig:imp_beta1}
	\end{subfigure} \\
	\begin{subfigure}{0.5\textwidth}
		\centering
		\caption{$\hat{c}_{\scriptscriptstyle 1,t}$}
		\includegraphics[width=0.65\linewidth]{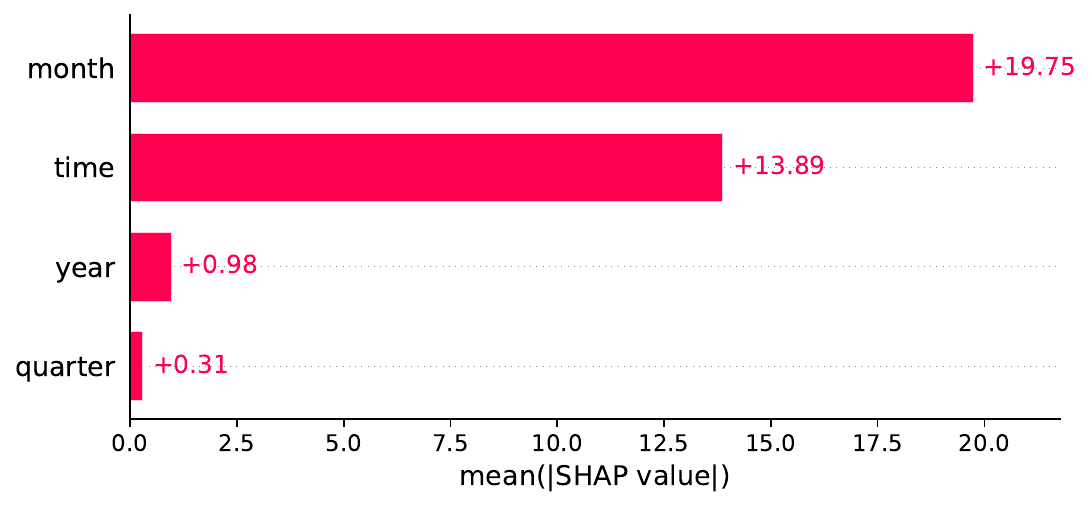}
		\label{fig:imp_sine}
	\end{subfigure} \hspace{-4em}  %
	\begin{subfigure}{0.5\textwidth}
		\centering
		\caption{$\hat{d}_{\scriptscriptstyle 1,t}$}
		\includegraphics[width=0.65\linewidth]{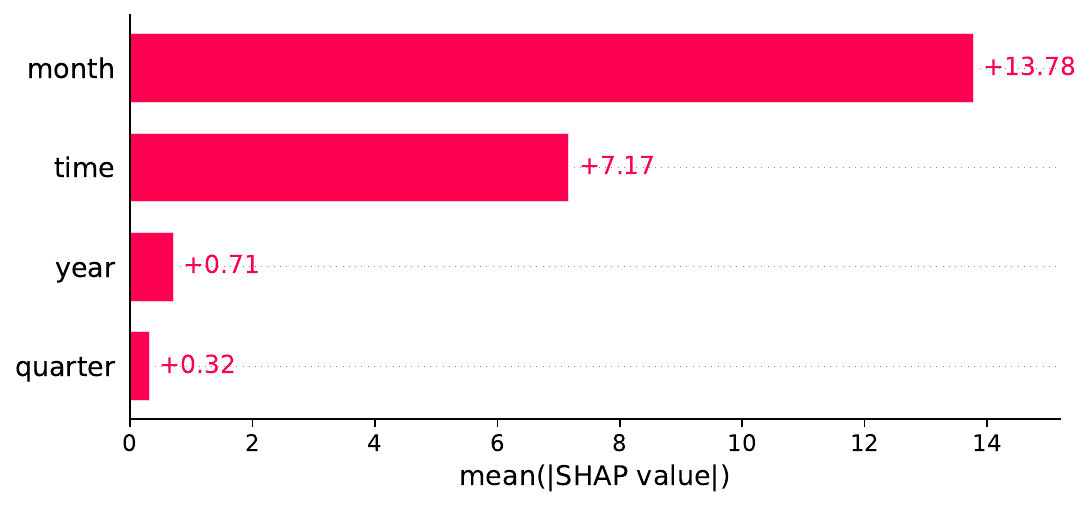}
		\label{fig:imp_cosine}
	\end{subfigure}	
	\label{fig:stl_imp_params}
\end{figure}

The feature importances for the trend parameters (Panels \ref{fig:imp_beta0} and \ref{fig:imp_beta1}) show that mostly features related to the progression of time are considered important, while the month-feature that reflects the within year patterns ranks highest for the seasonality parameters (Panels \ref{fig:imp_sine} and \ref{fig:imp_cosine}). From the above STL discussion, we can infer that our \texttt{Hyper-Tree-STL} approach gives results close to the conventional decomposition, while providing the additional capabilities of adaptive tree-based parameter estimation and enhanced interpretability. 

As the second step of our analysis, we now turn to the forecasting part for the last 12 months of the \texttt{Air Passengers} dataset. As target models, we specify the following \texttt{Hyper-Tree-AR(12)} and \texttt{Hyper-TreeNet-AR(12)} models
\begin{align}
	y_{t} = \theta_{\scriptscriptstyle 1,t}(\mathbf{x}_{\scriptscriptstyle t}) y_{t-1} + \theta_{\scriptscriptstyle 2,t}(\mathbf{x}_{\scriptscriptstyle t}) y_{t-2} + \dots + \theta_{\scriptscriptstyle 12,t}(\mathbf{x}_{\scriptscriptstyle t}) y_{t-12}
\end{align}

with time-step and feature specific parameters $\{\theta_{\scriptscriptstyle 1,t}(\mathbf{x}_{\scriptscriptstyle t}), \ldots, \theta_{\scriptscriptstyle 12,t}(\mathbf{x}_{\scriptscriptstyle t})\}$, as well as an \texttt{Hyper-Tree-ETS} model with a damped trend and multiplicative seasonality \citep{Hyndman.2021}
\begin{align}
    \ell_{t} &= \alpha_{_{t}}(\mathbf{x}_{_{t}})\frac{y_{t}}{s_{t-m_{\text{season}}}} + \bigl(1 - \alpha_{_{t}}(\mathbf{x}_{_{t}})\bigr)\bigl(\ell_{t-1} + \phi_{_{t}}(\mathbf{x}_{_{t}}) b_{t-1}\bigr) \label{eq:ets_level} \\	
	b_{t} &= \beta_{_{t}}(\mathbf{x}_{_{t}})(\ell_{t} - \ell_{t-1}) + \bigl(1 - \beta_{_{t}}(\mathbf{x}_{_{t}})\bigr)\phi_{_{t}}(\mathbf{x}_{_{t}}) b_{t-1} \label{eq:ets_trend} \\	
	s_{t} &= \gamma_{_{t}}(\mathbf{x}_{_{t}}) \frac{y_{t}}{\bigl(\ell_{t-1} + \phi_{_{t}}(\mathbf{x}_{_{t}}) b_{t-1}\bigr)} + \bigl(1 - \gamma_{_{t}}(\mathbf{x}_{_{t}})\bigr)s_{t-m_{\text{season}}} \label{eq:ets_seasonality}
\end{align}
 
where $\{\ell_{t}, b_{t}, s_{t}\}$ denote $\{\text{level}, \text{trend}, \text{seasonality}\}$ with corresponding time-step and feature specific parameters $\theta_{\scriptscriptstyle \mathcal{T}}(\mathbf{x}_{\scriptscriptstyle t}) = \bigl[\alpha_{\scriptscriptstyle t}(\mathbf{x}_{\scriptscriptstyle t}), \beta_{\scriptscriptstyle t}(\mathbf{x}_{\scriptscriptstyle t}), \gamma_{\scriptscriptstyle t}(\mathbf{x}_{\scriptscriptstyle t}), \phi_{\scriptscriptstyle t}(\mathbf{x}_{\scriptscriptstyle t})\bigr]$, where $m_{\text{season}}$ indicates the frequency of the seasonality. The h-step ahead forecast of the \texttt{Hyper-Tree-ETS} model is given by
\begin{align}
	\hat{y}_{t+h|t} = \left[\hat{\ell}_t + \sum_{u=1}^{h} \left(\prod_{v=1}^{u} \hat{\phi}_{t+v}(\mathbf{x}_{t+v})\right) \hat{b}_t \right] \, \hat{s}_{t+h - m_{\text{season}}(k_{\text{season}}+1)} 
\end{align}

The choice of the target AR(12) models is not based on automated model selection. Instead, the decision to use 12 lags stems from the monthly nature of the dataset. Similarly, the trend and seasonality of the \texttt{Air Passengers} dataset have led us to use Exponential Smoothing \citep{Holt.2004, Winters.1960} with multiplicative seasonality and damped trend as the target model for the \texttt{Hyper-Tree-ETS}, since this specification often provides accurate and robust forecasts \citep{Hyndman.2021}.

Before we turn to the results presented in Table \ref{tab:airp_metrics}, it is important to note that the \texttt{Hyper-Tree-ETS} model differs structurally from all other \texttt{Hyper-Tree} variants in a way that has direct consequences for the Hessian computation. In the AR-model variants, the fitted value $\hat{y}_{t}$ depends only on the parameters at time $t$ and the corresponding lags, making the Hessian block-diagonal across time steps and the standard GBDT second-order computation numerically stable. The ETS model, by contrast, is a state-space recurrence: equations \eqref{eq:ets_level}-\eqref{eq:ets_seasonality} show that each state $\{\ell_t, b_t, s_t\}$ depends on the smoothing parameters at all preceding time steps. The resulting Hessian is therefore dense rather than block-diagonal, and its cross-temporal second derivatives vanish or explode through long recurrences, analogous to second-order vanishing/exploding gradients in RNNs \citep{Bengio.1994, Pascanu.2013}, rendering the standard GBDT curvature estimates numerically unstable. We therefore approximate the Hessian using the Gauss-Newton method \citep{Nocedal.2006, Martens.2020}, which drops the residual-curvature term, whose second-order derivatives propagate through the full recurrence, retaining only the first-order
(Jacobian) term and guaranteeing positive semi-definite and numerically stable curvature estimates. The diagonal of the resulting approximation is estimated via Hutchinson's stochastic diagonal estimator \citep{Hutchinson.1990} with $K$ random Gaussian probes.\footnote{While gradients also propagate through the recurrence, they involve only products of Jacobians, which are stabilized by the contractive ETS state equations, whose smoothing parameters are bounded in $(0,1)$. The Hessian is more severely affected because the residual-curvature term requires second-order derivatives of the full recurrence, making the computation numerically less stable than the gradients. The Gauss-Newton approximation of the Hessian eliminates this term, retaining only the Jacobian-based component that shares the stability properties of the gradients.} 

\begin{table}[h!]
\fontsize{8pt}{9pt}\selectfont
\begin{center}
\begin{threeparttable}
\caption{\texttt{Air Passengers} Results.}	
\begin{tabular}{ll||ccccccc}
	\toprule
	& Model & MAPE & sMAPE & WAPE & RMSE & MAE & \\
	\midrule
	& \texttt{AR} & 8.630 & 9.206 & 9.302 & 54.891 & 44.292 \\
	& \texttt{AR-X} & 8.706 & 9.292 & 9.354 & 54.995 & 44.542 \\
	& \texttt{AutoARIMA} & 4.180 & 4.031 & 3.889 & 23.919 & 18.516 \\
	& \texttt{AutoARIMA-X} & 3.294 & 3.200 & 3.080 & 19.525 & 14.665 \\
	& \texttt{AutoETS} & 4.499 & 4.589 & 4.677 & 25.280 & 22.270 \\
	& \texttt{Hyper-Tree-AR} & \textbf{2.524} & \textbf{2.470} & \textbf{2.395} & \textbf{15.783} & \textbf{11.406} \\
	& \texttt{Hyper-Tree-ETS} & 3.899 & 3.965 & 3.750 & 20.469 & 17.858 \\
	& \texttt{Hyper-TreeNet-AR} & 4.119 & 4.144 & 3.972 & 21.289 & 18.915 \\
	& \texttt{LightGBM} & 3.070 & 3.068 & 2.927 & 17.765 & 13.939 \\
	& \texttt{LightGBM-AR} & 3.752 & 3.699 & 3.620 & 20.973 & 17.238 \\
	& \texttt{LightGBM-STL} & 3.796 & 3.822 & 3.923 & 23.471 & 18.678 \\
	\bottomrule
\end{tabular}
\begin{tablenotes}
	\scriptsize
	\setlength{\leftmargin}{0pt}
	\setlength{\labelsep}{0pt}
	\setlength{\labelwidth}{0pt}
	\setlength{\itemindent}{0pt}
	\item[] Reported are the forecast errors, with lower values indicating better performance and the best metrics highlighted in bold. Mean Absolute Percentage Error (MAPE); Symmetric Mean Absolute Percentage Error (sMAPE); Weighted Absolute Percentage Error (WAPE); Root Mean Squared Error (RMSE); Mean Absolute Error (MAE).
\end{tablenotes}
\label{tab:airp_metrics}
\end{threeparttable}
\end{center}
\end{table}

The results in Table \ref{tab:airp_metrics} show that Hyper-Tree models are competitive, with \texttt{Hyper-Tree-AR} being the most accurate across all evaluated metrics. Despite sharing the same underlying autoregressive structure but with static parameters, \texttt{AR-X} exhibits the lowest accuracy across all metrics. The \texttt{Hyper-Tree-AR} appears to capture the behavior of the series more effectively, likely due to its time-varying AR-parameters that are modeled as a function of time-related features and their interactions. While \texttt{AutoARIMA} benefits from automatic model selection and differencing, the selected model is an \texttt{ARIMA(1,1,0)(0,1,0)[12]}, our \texttt{Hyper-Tree-AR} achieves competitive results with a plain AR(12) specification without differencing or automated lag selection. Hence, the accuracy of the \texttt{Hyper-Tree-AR} model could likely be improved by incorporating seasonal differencing or by adapting the lag structure via automatic selection. Among the other variants, \texttt{Hyper-TreeNet-AR} and \texttt{Hyper-Tree-ETS} models compare well across metrics, though neither matches the accuracy of the \texttt{Hyper-Tree-AR}. Yet, \texttt{Hyper-Tree-ETS} yields more accurate results than its classical counterpart \texttt{AutoETS} across all metrics and remains competitive with both \texttt{AutoARIMA} variants. Conventional tree-based models, while performing well, lag behind our \texttt{Hyper-Tree-AR} model.

\subsection{Local Time Series Models: Extended Evaluation} \label{sec:local_multiple}

In the previous section, we evaluated the Hyper-Tree architectures on a single, well-behaved time series with strong seasonality and trend characteristics that can easily be modeled using time-derived features. To assess the accuracy of our Hyper-Tree architectures more comprehensively, we now extend our evaluation to additional datasets with more diverse characteristics. For the \texttt{Australian Retail Turnover} and \texttt{Tourism} datasets, we modify our \texttt{Hyper-Tree-ETS} training approach due to computational constraints. The recursive nature of ETS model training requires iterating over the entire time series at each boosting step, making per-series training of our \texttt{Hyper-Tree-ETS} model computationally expensive for datasets containing many long series. To address this challenge, we leverage a unique property of ETS models: their parameters are bounded in $[0,1]$, enabling a simplified parameterization. Rather than learning parameters for each series individually, we adopt a parameterization strategy where we set parameter values for all series globally. Specifically, we set $\{\alpha_{\scriptscriptstyle t}, \beta_{\scriptscriptstyle t}, \gamma_{\scriptscriptstyle t}, \phi_{\scriptscriptstyle t}\} =  0.3$ for all series in both datasets, selected through grid search over \{0.1, ..., 0.9\} on a hold-out set optimizing for WAPE. Table \ref{tab:local_metrics} presents the results. 

\begin{table}[h!]
\fontsize{8pt}{9pt}\selectfont
\begin{center}
\begin{threeparttable}
\caption{Local Model Results.}		
\begin{tabular}{ll||ccccccc}
\toprule
Dataset & Model & MAPE & sMAPE & WAPE & RMSE & MAE \\
\midrule
\multirow[t]{11}{*}{\texttt{Australian Electricity Demand}}
	& \texttt{AR} & 5.291 & 5.129 & 5.151 & 257,965.5 & 224,999.9 \\
	& \texttt{AR-X} & 5.166 & 5.014 & 5.035 & 253,778.7 & 222,623.4 \\
	& \texttt{AutoARIMA} & 3.421 & 3.398 & 3.430 & 199,044.3 & 158,517.8 \\
	& \texttt{AutoARIMA-X} & 4.146 & 4.048 & 4.134 & 225,535.0 & 188,831.5 \\
	& \texttt{AutoETS} & 3.480 & 3.430 & 3.479 & 177,123.1 & 133,774.5 \\
	& \texttt{Hyper-Tree-AR} & 4.140 & 4.047 & 4.076 & 218,432.8 & 193,903.3 \\
	& \texttt{Hyper-Tree-ETS} & 3.845 & 3.765 & 3.800 & 204,600.1 & 172,143.5 \\
	& \texttt{Hyper-TreeNet-AR} & \textbf{2.925} & \textbf{2.939} & \textbf{2.938} & \textbf{160,191.1} & \textbf{118,736.9} \\
	& \texttt{LightGBM} & 3.067 & 3.076 & 3.081 & 178,666.6 & 137,318.4 \\
	& \texttt{LightGBM-AR} & 5.232 & 5.086 & 5.148 & 266,541.8 & 237,194.2 \\
	& \texttt{LightGBM-STL} & 6.208 & 6.556 & 6.273 & 310,754.9 & 242,870.6 \\
	\cline{1-7}
	\multirow[t]{11}{*}{\texttt{Australian Retail Turnover}}
	& \texttt{AR} & 9.090 & 9.089 & 9.131 & 30.694 & 24.463 \\
	& \texttt{AR-X} & 8.320 & 8.309 & 8.372 & 27.563 & 21.716 \\
	& \texttt{AutoARIMA} & 7.146 & 7.096 & 7.051 & 18.533 & 16.235 \\
	& \texttt{AutoARIMA-X} & 6.846 & 6.681 & 6.755 & 18.268 & 15.731 \\
	& \texttt{AutoETS} & \textbf{6.101} & \textbf{6.113} & \textbf{6.112} & 17.128 & 14.037 \\
	& \texttt{Hyper-Tree-AR} & 6.997 & 6.781 & 7.001 & 19.728 & 16.795 \\
	& \texttt{Hyper-Tree-ETS} & 6.312 & 6.399 & 6.372 & 20.062 & 16.800 \\
	& \texttt{Hyper-TreeNet-AR} & 6.248 & 6.218 & 6.298 & \textbf{16.612} & \textbf{13.659} \\
	& \texttt{LightGBM} & 9.180 & 8.864 & 9.025 & 20.368 & 17.166 \\
	& \texttt{LightGBM-AR} & 9.482 & 8.908 & 9.455 & 23.676 & 19.736 \\
	& \texttt{LightGBM-STL} & 10.734 & 11.020 & 10.905 & 26.810 & 23.162 \\
	\cline{1-7}
	\multirow[t]{11}{*}{\texttt{Tourism}}
	& \texttt{AR} & 25.052 & 21.846 & 20.275 & 2,979.4 & 2,291.9 \\
	& \texttt{AR-X} & 24.927 & 21.644 & 19.922 & 2,924.1 & 2,238.9 \\
	& \texttt{AutoARIMA} & 21.132 & 19.647 & 18.310 & 2,624.0 & 2,092.0 \\
	& \texttt{AutoARIMA-X} & 20.467 & 19.086 & 17.957 & \textbf{2,522.9} & 1,984.3 \\
	& \texttt{AutoETS} & 20.880 & 19.320 & 18.528 & 2,735.6 & 2,164.3 \\
	& \texttt{Hyper-Tree-AR} & 22.889 & 18.813 & 18.159 & 3,172.8 & 2,605.0 \\
	& \texttt{Hyper-Tree-ETS} & 31.648 & 19.228 & 26.818 & 2,595.8 & \textbf{1,954.4} \\
	& \texttt{Hyper-TreeNet-AR} & \textbf{20.216} & \textbf{18.419} & \textbf{17.451} & 3,239.4 & 2,690.3 \\
	& \texttt{LightGBM} & 31.843 & 25.337 & 22.550 & 4,377.8 & 3,785.0 \\
	& \texttt{LightGBM-AR} & 30.448 & 23.504 & 23.422 & 4,311.3 & 3,270.9 \\
	& \texttt{LightGBM-STL} & 30.529 & 27.608 & 27.392 & 5,047.4 & 3,852.4 \\
\bottomrule
\end{tabular}
\begin{tablenotes}
\scriptsize
\setlength{\leftmargin}{0pt}
\setlength{\labelsep}{0pt}
\setlength{\labelwidth}{0pt}
\setlength{\itemindent}{0pt}
\item[] Reported are the mean forecast errors across series per dataset, with lower values indicating better performance and the best metrics highlighted in bold. Mean Absolute Percentage Error (MAPE); Symmetric Mean Absolute Percentage Error (sMAPE); Weighted Absolute Percentage Error (WAPE); Root Mean Squared Error (RMSE); Mean Absolute Error (MAE). 
\end{tablenotes}
\label{tab:local_metrics}
\end{threeparttable}
\end{center}
\end{table}

Our evaluation reveals several interesting patterns across the three datasets. Most notably, \texttt{Hyper-TreeNet-AR} consistently ranks among the top models. Our hybrid model outperforms both classical time series models and conventional tree-based approaches, demonstrating the effectiveness of combining tree-based representations with neural parameter estimation. The purely tree-based Hyper-Tree variants also show strong performance. On the \texttt{Australian Electricity Demand} dataset, \texttt{Hyper-Tree-AR} surpasses its classical counterparts (\texttt{AR}, \texttt{AR-X} and \texttt{AutoARIMA-X}) on most metrics. On the \texttt{Australian Retail Turnover} dataset, the \texttt{Hyper-Tree-ETS} model performs reasonably well, suggesting that exponential smoothing captures the dataset's seasonal patterns. For some datasets, the standard \texttt{LightGBM} model yields more accurate forecasts than the \texttt{LightGBM-AR} variant, despite the latter's use of autoregressive features. This points to the possibility that simply adding lagged target values to tree-based models may not effectively capture temporal dependencies without an appropriate inductive bias, validating our hyper-model approach that naturally integrates time series dynamics into tree-based models.

\newpage

\subsection{Global Time Series Models} \label{sec:global}

In previous sections, we evaluated Hyper-Trees in a local forecasting setting, training separate models for each time series individually without leveraging cross-series information. We now shift focus to global forecasting \citep{Januschowski.2020}, where Hyper-Trees can leverage their ability to capture shared patterns across multiple series. Operational forecasting \citep{Januschowski.2019} typically involves datasets with high-dimensional feature sets that go beyond basic time series information, including numeric and categorical features. In such environments, tree-based models typically excel, providing a suitable setting to assess the accuracy of Hyper-Trees. To create a more realistic testing ground for our framework, we add the \texttt{Rossmann} and \texttt{M5} datasets that closely resemble real-world corporate datasets, in addition to the datasets used in the local forecasting evaluation. Also, we add the commonly used \texttt{M3 Monthly} and \texttt{M3 Yearly} datasets. Local forecasting models, including \texttt{LightGBM-STL}, are excluded from our global forecasting evaluation. This decision stems from their inherent limitations in handling complex, multi-series forecasting tasks. Unlike global models, local models cannot leverage cross-series information, as they are typically trained on individual time series in isolation. This limitation in learning from multiple series simultaneously hinders their ability to capture broader patterns across the entire dataset, potentially reducing their generalization capabilities \citep{Manso.2021}. Furthermore, local models struggle to effectively incorporate categorical features, which are crucial for borrowing strength across similar series. Yet, while we exclude local models from our global model evaluation, we use \texttt{AutoETS} forecasts as the reference for calculating a scaled error metric following the Mean Absolute Scaled Error (MASE, \cite{Hyndman.2006}). This choice is based on the strong performance of the \texttt{AutoETS} among local models, as reported in Table \ref{tab:local_metrics}, making it a robust benchmark for assessing the accuracy of global forecasting models. 

Besides comparing our approach to conventional tree-based and deep-learning models, we also include a foundation time series model in our evaluation. This inclusion reflects the trend in time series forecasting towards large-scale, pre-trained models. Among others, the most prominent include Chronos \citep{Ansari.2024}, Moirai \citep{Woo.2024}, Moment \citep{Goswami.2024}, TimesFM \citep{Das.2024}, Tiny Time Mixers \citep{Ekambaram.2024} and Lag-Llama \citep{Rasul.2023}. Trained on extensive and diverse time series datasets, these models aim to improve accuracy and generalization in forecasting tasks. For our experiments, we use Chronos \citep{Ansari.2024}, which has demonstrated strong performance across multiple benchmark datasets.\footnote{According to \citet{Ansari.2024}, Chronos was not pre-trained on any of the datasets used in our evaluation and all forecasts of Chronos \citep{Ansari.2024} reported in Table \ref{tab:global_metrics} are zero-shot.} Compared to deep-learning models that learn abstract representations from time series data, it is important to note that tree-based models, including our Hyper-Tree approaches, operate on the information provided to them and cannot generate new features that are not present in the original input. Instead, they excel at modeling complex relationships within the provided feature set. To better capture the overall characteristics of each series and enhance the local adaption of global tree-based models across diverse time series datasets, we incorporate additional features using the established approach of \citet{Hyndman.2024}, which characterizes each series based on information derived from the original time series. While embeddings from foundation models like Chronos \citep{Ansari.2024} could be used as feature inputs to the tree-models as well, we opted for traditional statistical features to ensure that the evaluation of our Hyper-Tree models is not influenced by the quality of the embeddings of a foundation model.

To ensure as fair a comparison as possible across models, we have made several decisions for our experimental setup. We disable the standard lag and time-feature generation in \texttt{Deep-AR} \citep{Salinas.2020} and \texttt{TFT} \citep{Lim.2021}. Instead, we use the same lag specification and time-derived features across all models. Also, we exclude the time series derived features of \citet{Hyndman.2024} from deep-learning models and use them only in tree-based models, as these are not commonly used in neural network architectures. We estimate the 50\% quantile using \texttt{TFT} \citep{Lim.2021}, while for \texttt{Deep-AR} \citep{Salinas.2020}, we calculate the mean from 1,000 samples drawn from a forecasted Student-T distribution. In the case of \texttt{Chronos} \citep{Ansari.2024}, we use the \texttt{t5-base} version with default parameters and draw 50 samples from the forecasted distribution. We evaluate the \texttt{Hyper-Tree-ETS} model only on datasets with stable seasonality across series, in line with the multiplicative seasonality assumption of the ETS specification, which excludes the \texttt{M3 Monthly}, \texttt{M5}, and \texttt{Rossmann} datasets. Further, the vectorized \texttt{Hyper-Tree-ETS} estimation requires uniform series lengths, with the data structured in a tensor of shape \texttt{(number of series, context-length, number of channels)}. This format allows for efficient vectorized estimation across multiple time series but necessitates that all series have the same length within a given dataset. For datasets with varying series lengths, we employ the following strategy for the \texttt{Hyper-Tree-ETS} models: instead of using zeros or constants, we back-append a segment from the series itself to match the length of the longest series. This preserves the series' characteristics, preventing distortion of seasonality and trend initialization that would occur with constant or zero-padding. To ensure that padded observations do not influence the estimation of ETS parameters, we apply a mask during training that zeroes out the contributions of padded time steps and also add a binary flag as an additional feature to identify padded observations. All other models in our evaluation use the datasets in their original form without this preprocessing step. Additionally, for the \texttt{Hyper-Tree-ETS} model, we apply a scaling transformation to each series of the \texttt{Tourism} dataset by adding a constant offset value. This transformation is necessary for the multiplicative seasonality specification to avoid zeros and prevent numerical instability. For the \texttt{M3 Yearly} dataset, which lacks seasonality and is predominantly driven by trends, we adapt our \texttt{Hyper-Tree-ETS} model to an ETS formulation with a linear trend only.

Turning to the results in Table \ref{tab:global_metrics}, our Hyper-Tree models demonstrate competitive performance across a range of established baselines and datasets. On the \texttt{Australian Electricity Demand} dataset, the \texttt{Hyper-TreeNet-AR} model leads across all accuracy measures, while on the \texttt{Australian Retail Turnover} dataset, \texttt{Hyper-Tree-ETS} and \texttt{Hyper-TreeNet-AR} share the top positions across metrics. On the operationally more complex \texttt{Rossmann Store Sales} dataset, where conventional GBDTs are the typical state-of-the-art, \texttt{Hyper-Tree-AR} is the most accurate model. This shows that a classical autoregressive model, when parameterized via GBDTs, can compete with and even surpass conventional tree-based forecasting models, highlighting the competitiveness of our framework in settings where traditional GBDTs typically excel and validating its applicability in complex, feature-rich environments common to operational forecasting. For the \texttt{M3 Monthly} dataset, \texttt{Chronos} \citep{Ansari.2024} achieves the highest accuracy among all models, albeit with a higher runtime compared to our Hyper-Tree approaches. The \texttt{Hyper-Tree-ETS$^\ast$} is the most accurate model on the \texttt{M3 Yearly} data, indicating that exponential smoothing with a linear trend is well-suited for this type of annual data. The \texttt{Hyper-Tree-ETS} model with multiplicative seasonality and a damped trend shows good performance on datasets with clear seasonal patterns, though its recursive estimation leads to comparatively higher computational costs. The results of the \texttt{Hyper-Tree-ETS} for the \texttt{Tourism} dataset, however, indicate that its applicability is limited, likely due to a misalignment between the model's multiplicative seasonality assumptions and the heterogeneous seasonal patterns in the data. This illustrates a key characteristic of the Hyper-Tree framework: while inductive bias provides advantages when target model assumptions match the data characteristics, accuracy degrades when they diverge. The \texttt{M5} dataset results show that while \texttt{Deep-AR} achieves the highest accuracy across error metrics, \texttt{TFT}, \texttt{Chronos} and our \texttt{Hyper-TreeNet-AR} remain competitive alternatives, with \texttt{Hyper-TreeNet-AR} offering faster runtimes. Table \ref{tab:global_metrics} also shows that standard \texttt{LightGBM} and \texttt{LightGBM-AR} models consistently offer the fastest training times but lag on accuracy measures for some datasets. Overall, these results demonstrate that Hyper-Tree models effectively combine tree-based feature learning with time series inductive biases, bridging traditional statistical approaches and machine learning. The effectiveness of our framework depends on selecting target models whose assumptions align with the underlying data characteristics, making this choice a crucial consideration.

\newpage

\begin{table}[!htbp]
	\vspace{-5em}
	\fontsize{8pt}{9pt}\selectfont
	\begin{adjustwidth}{-1cm}{}
		\begin{center}
			\begin{threeparttable}
				\caption{Global Model Results.}			
				\begin{tabular}{ll||cccccc|c}
					\toprule
					&  &  &  &  &  &  &  &  Runtime \\
					Dataset & Model & MAPE & sMAPE & WAPE & RMSE & MAE & MASE & (Minutes) \\
					\midrule
					\multirow[t]{8}{*}{\texttt{Australian Electricity Demand}}
					& \texttt{Chronos} & 3.608 & 3.579 & 3.594 & 189,129.1 & 138,833.2 & 1.049 & 0.027 \\
					& \texttt{Deep-AR} & 3.594 & 3.628 & 3.634 & 201,459.5 & 155,123.3 & 1.142 & 1.273 \\
					& \texttt{Hyper-Tree-AR} & 4.125 & 4.034 & 4.062 & 217,635.7 & 192,981.4 & 1.300 & 0.022 \\
					& \texttt{Hyper-Tree-ETS} & 3.954 & 3.885 & 3.920 & 233,281.0 & 196,350.0 & 1.331 & 0.355 \\
					& \texttt{Hyper-TreeNet-AR} & \textbf{2.964} & \textbf{2.950} & \textbf{2.960} & \textbf{163,026.9} & \textbf{121,972.3} & \textbf{0.886} & 0.022 \\
					& \texttt{LightGBM} & 5.909 & 6.173 & 5.894 & 269,868.1 & 210,619.9 & 1.723 & \textbf{0.002} \\
					& \texttt{LightGBM-AR} & 6.086 & 5.872 & 5.976 & 281,027.3 & 254,324.2 & 1.842 & \textbf{0.002} \\
					& \texttt{TFT} & 4.607 & 4.544 & 4.594 & 231,271.9 & 191,331.3 & 1.374 & 2.015 \\
					\cline{1-9}
					\multirow[t]{8}{*}{\texttt{Australian Retail Turnover}}
					& \texttt{Chronos} & 7.117 & 6.914 & 7.026 & 18.166 & 15.832 & 1.243 & 1.252 \\
					& \texttt{Deep-AR} & 7.752 & 7.363 & 7.685 & 21.021 & 18.519 & 1.403 & 5.355 \\
					& \texttt{Hyper-Tree-AR} & 6.877 & 6.656 & 6.863 & 19.613 & 16.729 & 1.271 & 0.618 \\
					& \texttt{Hyper-Tree-ETS} & \textbf{6.202} & 6.300 & \textbf{6.252} & 17.966 & 15.062 & \textbf{1.066} & 1.166 \\
					& \texttt{Hyper-TreeNet-AR} & 6.403 & \textbf{6.292} & 6.427 & \textbf{16.179} & \textbf{13.409} & 1.090 & 0.109 \\
					& \texttt{LightGBM} & 12.950 & 12.530 & 12.785 & 22.006 & 18.866 & 2.234 & \textbf{0.012} \\
					& \texttt{LightGBM-AR} & 8.247 & 7.882 & 8.134 & 21.097 & 17.926 & 1.479 & 0.013 \\
					& \texttt{TFT} & 9.756 & 9.085 & 9.669 & 26.250 & 22.725 & 1.832 & 10.299 \\
					\cline{1-9}
					\multirow[t]{8}{*}{\texttt{M3-Monthly}}
					& \texttt{Chronos} & \textbf{16.668} & \textbf{13.874} & \textbf{13.724} & \textbf{751.987} & \textbf{618.140} & \textbf{1.138} & 5.262 \\
					& \texttt{Deep-AR} & 21.728 & 14.641 & 14.739 & 814.384 & 673.038 & 1.342 & 3.511 \\
					& \texttt{Hyper-Tree-AR} & 19.580 & 14.565 & 14.720 & 809.820 & 672.991 & 1.261 & 0.401 \\
					& \texttt{Hyper-Tree-ETS} & -- & -- & -- & -- & -- & -- & -- \\
					& \texttt{Hyper-TreeNet-AR} & 19.778 & 14.659 & 14.603 & 814.225 & 677.430 & 1.374 & 0.075 \\
					& \texttt{LightGBM} & 26.486 & 17.504 & 18.578 & 945.055 & 803.409 & 2.391 & 0.031 \\
					& \texttt{LightGBM-AR} & 21.445 & 14.479 & 14.743 & 786.295 & 654.822 & 1.406 & \textbf{0.029} \\
					& \texttt{TFT} & 19.752 & 14.657 & 14.551 & 796.385 & 653.173 & 1.231 & 4.021 \\
					\cline{1-9}
					\multirow[t]{8}{*}{\texttt{M3-Yearly}}
					& \texttt{Chronos} & \textbf{20.217} & 18.306 & 17.410 & 1,245.9 & 1,073.0 & 1.561 & 0.893 \\
					& \texttt{Deep-AR} & 28.425 & 19.340 & 22.664 & 1,447.8 & 1,236.0 & 2.741 & 1.383 \\
					& \texttt{Hyper-Tree-AR} & 24.631 & 19.634 & 20.925 & 1,647.4 & 1,424.9 & 1.839 & 0.021 \\
					& \texttt{Hyper-Tree-ETS$^\ast$} & 21.242 & \textbf{16.650} & \textbf{17.082} & \textbf{1,229.5} & \textbf{1,069.7} & \textbf{1.204} & 0.130 \\
					& \texttt{Hyper-TreeNet-AR} & 21.782 & 18.389 & 18.133 & 1,233.0 & 1,088.0 & 1.558 & 0.031 \\
					& \texttt{LightGBM} & 30.376 & 23.676 & 26.071 & 1,605.0 & 1,402.6 & 4.164 & \textbf{0.003} \\
					& \texttt{LightGBM-AR} & 24.798 & 17.554 & 20.522 & 1,781.6 & 1,510.1 & 1.870 & 0.004 \\
					& \texttt{TFT} & 27.197 & 18.603 & 20.682 & 1,372.6 & 1,195.8 & 1.488 & 1.912 \\
					\cline{1-9}
					\multirow[t]{8}{*}{\texttt{M5}}
					& \texttt{Chronos} & 13.997 & 13.326 & 13.274 & 81.722 & 63.557 & 0.971 & 0.875 \\
					& \texttt{Deep-AR} & \textbf{13.231} & \textbf{12.712} & \textbf{12.605} & \textbf{74.547} & \textbf{59.611} & \textbf{0.934} & 1.463 \\
					& \texttt{Hyper-Tree-AR} & 17.583 & 14.918 & 15.608 & 91.369 & 74.621 & 1.120 & 0.237 \\
					& \texttt{Hyper-Tree-ETS} & -- & -- & -- & -- & -- & -- & -- \\
					& \texttt{Hyper-TreeNet-AR} & 15.290 & 13.671 & 13.926 & 79.713 & 63.052 & 0.990 & 0.049 \\
					& \texttt{LightGBM} & 20.469 & 16.166 & 18.080 & 78.348 & 63.039 & 1.150 & \textbf{0.012} \\
					& \texttt{LightGBM-AR} & 15.571 & 13.910 & 14.174 & 87.723 & 70.362 & 1.054 & 0.015 \\
					& \texttt{TFT} & 14.106 & 13.045 & 13.100 & 79.366 & 63.579 & 0.961 & 2.301 \\
					\cline{1-9}
					\multirow[t]{8}{*}{\texttt{Rossmann Store Sales}}
					& \texttt{Chronos} & 12.799 & 12.260 & 12.284 & 1,140.4 & 857.111 & 0.487 & 20.137 \\
					& \texttt{Deep-AR} & 9.897 & 9.342 & 9.609 & 860.314 & 665.007 & 0.383 & 8.482 \\
					& \texttt{Hyper-Tree-AR} & \textbf{8.830} & 9.023 & \textbf{8.881} & \textbf{798.023} & \textbf{622.134} & 0.358 & 8.880 \\
					& \texttt{Hyper-Tree-ETS} & -- & -- & -- & -- & -- & -- & -- \\
					& \texttt{Hyper-TreeNet-AR} & 9.437 & 9.597 & 9.449 & 857.004 & 664.496 & 0.385 & 0.693 \\
					& \texttt{LightGBM} & 10.456 & 10.559 & 10.352 & 895.679 & 712.080 & 0.423 & \textbf{0.155} \\
					& \texttt{LightGBM-AR} & 9.187 & \textbf{8.898} & 8.997 & 806.228 & 622.404 & 0.359 & 0.189 \\
					& \texttt{TFT} & 9.258 & 8.913 & 8.982 & 822.939 & 624.228 & \textbf{0.357} & 11.328 \\
					\cline{1-9}
					\multirow[t]{8}{*}{\texttt{Tourism}}
					& \texttt{Chronos} & \textbf{22.113} & 23.925 & 21.739 & 2,870.8 & 2,318.0 & 1.211 & 2.420 \\
					& \texttt{Deep-AR} & 23.926 & 20.408 & 20.076 & \textbf{2,760.9} & \textbf{2,234.0} & 1.178 & 5.808 \\
					& \texttt{Hyper-Tree-AR} & 22.525 & \textbf{19.375} & \textbf{18.618} & 3,059.3 & 2,509.5 & \textbf{1.047} & 0.855 \\
					& \texttt{Hyper-Tree-ETS} & 28.638 & 25.991 & 22.431 & 4,501.9 & 3,504.2 & 1.278 & 0.986 \\
					& \texttt{Hyper-TreeNet-AR} & 23.731 & 20.603 & 19.587 & 3,031.1 & 2,515.8 & 1.113 & 0.135 \\
					& \texttt{LightGBM} & 47.469 & 63.132 & 47.401 & 9,715.2 & 9,055.1 & 3.440 & 0.030 \\
					& \texttt{LightGBM-AR} & 27.392 & 21.774 & 21.157 & 4,360.4 & 3,450.6 & 1.258 & \textbf{0.029} \\
					& \texttt{TFT} & 25.617 & 21.950 & 21.409 & 3,476.8 & 2,848.5 & 1.270 & 10.424 \\
					\bottomrule
					\end{tabular}
				\label{tab:global_metrics}
				\begin{tablenotes}
					\scriptsize
					\setlength{\leftmargin}{0pt}
					\setlength{\labelsep}{0pt}
					\setlength{\labelwidth}{0pt}
					\setlength{\itemindent}{0pt}
					\item[]Reported are the mean forecast errors across series per dataset and total runtimes in minutes, with lower values indicating better performance and the best metrics highlighted in bold. Mean Absolute Percentage Error (MAPE); Symmetric Mean Absolute Percentage Error (sMAPE); Weighted Absolute Percentage Error (WAPE); Root Mean Squared Error (RMSE); Mean Absolute Error (MAE); Mean Absolute Scaled Error (MASE), which is calculated relative to the \texttt{AutoETS} forecasts (MAE of model / MAE of \texttt{AutoETS}). Entries with '--' indicate that the model was not trained. For the \texttt{M3 Yearly} dataset, the \texttt{Hyper-Tree-ETS} parameterizes an ETS model with a linear trend only, as indicated by '$^\ast$'. For all other applicable datasets, the \texttt{Hyper-Tree-ETS} generates parameters for ETS models with multiplicative seasonality and a damped trend.
				\end{tablenotes}
			\end{threeparttable}
		\end{center}
	\end{adjustwidth}
\end{table}

\newpage

\subsubsection{Ablation Studies} \label{sec:ablation}

In this section, we conduct a series of ablation studies. By systematically removing or modifying components, we aim to gain a better understanding of the contribution of each model element to overall accuracy for the \texttt{Hyper-TreeNet-AR} and \texttt{Hyper-Tree-AR} models. Our ablations are conducted on the \texttt{Rossmann Store Sales} dataset, which closely reflects practical forecasting scenarios in both the scale of available time series and the complexity of the underlying data and features. This provides a suitable environment for evaluating the sensitivity of our models to various components. For our ablation studies, we have chosen the following modifications:

\begin{itemize}[itemsep=-2pt]
	\item \texttt{A1}: Increase the dimension of tree-embeddings $d$ from 1 to 5.
	\item \texttt{A2}: Disable linear-tree option of \cite{Shi.2019}.
	\item \texttt{A3}: Increase hidden-dimension of the MLP from 128 to 256.
	\item \texttt{A4}: Remove random projection layer.
	\item \texttt{A5}: Reduce the number of lags from 21 to 14.
	\item \texttt{A6}: Remove \textit{tsfeatures} \citep{Hyndman.2024} from the set of features.
	\item \texttt{A7}: Exclude GBDTs as encoding layer and use MLP only.
	\item \texttt{A8}: Use \texttt{TreeNet} to forecast directly, without learning the parameters of a target model.
	\item \texttt{A9}: Following the two-step approach described by \cite{He.2014}, we first train a LightGBM model \citep{Ke.2017} and extract the leaf indices from the decision trees. These leaf indices are then used as input features to a separately trained MLP.
	\item \texttt{A10}: Train model with a shared gradient flow according to \textit{Option 1} as described in Table \ref{tab:grads}.
	\item \texttt{A11}: Instead of using time-varying parameters, we average the parameters across the forecast horizon and use constant AR-parameters for generating forecasts. 
\end{itemize}

Each variant (\texttt{A1} - \texttt{A11}) is derived from the \texttt{Base} model as reported in Table \ref{tab:global_metrics}, with a specific modification applied to create each variation. This approach allows us to isolate and evaluate the impact of individual components. Table \ref{tab:ablation} shows the results. We first analyze the impact of each ablation on \texttt{Hyper-TreeNet-AR}, followed by \texttt{Hyper-Tree-AR}.

\begin{table}[h!]
	\fontsize{8pt}{9pt}\selectfont
	\begin{center}
		\begin{threeparttable}
			\caption{\texttt{Rossmann Store Sales} Ablation Study Results.}
			\begin{tabular}{ll||cccccc}
				\toprule
				Model & Ablation & MAPE & sMAPE & WAPE & RMSE & MAE & MASE \\						
                \midrule
				\multirow[t]{4}{*}{\texttt{Hyper-Tree-AR}} 
				& \texttt{Base} & 8.830 & 9.023 & 8.881 & 798.023 & 622.134 & 0.358 \\
				& \texttt{A2} & \textbf{8.638} & \textbf{8.811} & \textbf{8.692} & \textbf{780.799} & \textbf{608.668} & \textbf{0.351} \\
				& \texttt{A5} & 9.069 & 8.936 & 8.996 & 822.958 & 623.887 & 0.358 \\
				& \texttt{A6} & 10.343 & 10.428 & 10.251 & 921.092 & 708.853 & 0.415 \\
				& \texttt{A11} & 24.753 & 21.315 & 21.423 & 1,791.4 & 1,478.2 & 0.844 \\
				 	\midrule
				 \multirow[t]{7}{*}{\texttt{Hyper-TreeNet-AR}} 
				 & \texttt{Base} & 9.437 & 9.597 & 9.449 & 857.004 & 664.496 & 0.385 \\
				 & \texttt{A1} & \textbf{9.071} & \textbf{9.163} & \textbf{9.099} & \textbf{827.530} & \textbf{637.683} & \textbf{0.367} \\
				 & \texttt{A2} & 9.689 & 9.870 & 9.691 & 876.076 & 681.289 & 0.397 \\
				 & \texttt{A3} & 9.370 & 9.530 & 9.359 & 845.485 & 657.503 & 0.381 \\
				 & \texttt{A4} & 9.954 & 9.930 & 9.726 & 876.308 & 682.243 & 0.394 \\
				 & \texttt{A5} & 10.067 & 10.502 & 10.190 & 914.730 & 718.379 & 0.417 \\
				 & \texttt{A6} & 11.395 & 11.407 & 10.981 & 981.758 & 767.239 & 0.451 \\
				 & \texttt{A7} & 12.053 & 11.790 & 11.598 & 1,014.9 & 803.500 & 0.471 \\
				 & \texttt{A8} & 19.024 & 19.445 & 18.916 & 1,588.9 & 1,371.7 & 0.776 \\
				 & \texttt{A9} & 11.473 & 11.709 & 11.420 & 1,005.5 & 787.651 & 0.466 \\
				 & \texttt{A10} & 9.507 & 9.597 & 9.502 & 864.105 & 667.223 & 0.386 \\
				 & \texttt{A11} & 21.945 & 20.337 & 20.206 & 1,726.6 & 1,398.7 & 0.800 \\
				\bottomrule
			\end{tabular}
			\begin{tablenotes}
				\scriptsize
				\setlength{\leftmargin}{0pt}
				\setlength{\labelsep}{0pt}
				\setlength{\labelwidth}{0pt}
				\setlength{\itemindent}{0pt}
				\item[] Reported are the mean forecast errors across series, with lower values indicating better performance and the best metrics highlighted in bold. Mean Absolute Percentage Error (MAPE); Symmetric Mean Absolute Percentage Error (sMAPE); Weighted Absolute Percentage Error (WAPE); Root Mean Squared Error (RMSE); Mean Absolute Error (MAE); Mean Absolute Scaled Error (MASE), which is calculated relative to the \texttt{AutoETS} forecasts (MAE of model / MAE of \texttt{AutoETS}).
				\end{tablenotes}
			\label{tab:ablation}
		\end{threeparttable}
	\end{center}
\end{table}

Increasing the dimension of tree embeddings from 1 to 5 (\texttt{A1}) leads to an improvement across all error metrics, demonstrating that higher-dimensional representations enable more nuanced parameter estimation, albeit with increased computational cost. Disabling the linear tree option (\texttt{A2}) leads to a modest degradation in accuracy, while doubling the MLP hidden dimension (\texttt{A3}) slightly improves over the base model. The removal of the random projection layer (\texttt{A4}) degrades accuracy, confirming that this layer effectively diversifies low-dimensional tree embeddings into more expressive representations for the MLP decoder. Reducing the lag structure (\texttt{A5}) results in a notable drop in accuracy, emphasizing that an extended context is crucial for capturing the underlying temporal dependencies. Removing time series derived features (\texttt{A6}) hurts performance, highlighting an important property of the Hyper-Tree framework: since parameter learning is based on GBDTs, Hyper-Trees inherit the feature-based partitioning behavior characteristic of tree-models. This becomes critical in operational forecasting, where global models often struggle with adapting to the dynamics of individual series \citep{Godahewa.2021b}. This limitation, confirmed by observing similar degradation for the \texttt{LightGBM-AR} model under the same ablation, is not unique to Hyper-Trees but reflects a general characteristic of tree-based models when contextual features are missing.\footnote{The ablation metrics for \texttt{LightGBM-AR} are MAPE: 10.01; sMAPE: 9.68; WAPE: 9.62; RMSE: 855.75; MAE: 665.07.} The drop in accuracy when GBDTs are excluded as the encoding layer and an MLP is used only (\texttt{A7}) demonstrates the critical role of tree-based feature encoding and validates the effectiveness of our GBDT-neural hybrid architecture. The substantial drop in accuracy when the \texttt{Hyper}-part is removed and the \texttt{TreeNet} is used to forecast directly (\texttt{A8}) validates that learning the parameters of a target model is crucial for the \texttt{Hyper-TreeNet}'s effectiveness. The two-stage approach (\texttt{A9}) shows markedly lower accuracy than our jointly trained model. This gap in accuracy likely stems from differences in optimization objective and representation quality. In the two-stage approach, the GBDT is optimized for direct forecasting rather than representation learning, creating outputs in the form of leaf indices that are not suited as inputs to the MLP for estimating parameters of the target model. In contrast, our end-to-end framework explicitly optimizes the GBDT for generating informative embeddings specifically relevant for estimating the target model parameters. This joint learning enables co-adaptation between components, with the network guiding the GBDT toward more useful representations than would be possible via independent training. While we opted for \textit{Option 2} as the primary gradient method, results from \texttt{A10} indicate negligible differences in model accuracy between the two gradient flow approaches for this particular forecasting task and dataset. Nevertheless, different problem domains or datasets might reveal more pronounced differences, and the impact of the gradient flow choice presents an interesting direction for future research. The most severe deterioration in accuracy is observed for \texttt{A11}, where time-varying parameters are replaced with constant AR-parameters averaged across the forecast horizon. This decline in accuracy provides evidence that the time-varying nature of parameters is an important component of our framework's effectiveness. By constraining the model to constant parameters, we eliminate its ability to adapt to evolving patterns across the forecast horizon.

Turning to the \texttt{Hyper-Tree-AR} model, we observe that in contrast to the \texttt{Hyper-TreeNet-AR} model, disabling linear trees (\texttt{A2}) actually improves performance. Apart from this, the \texttt{Hyper-Tree-AR} shows similar sensitivity to a reduced lag structure (\texttt{A5}) and to removing the additional set of series-derived features (\texttt{A6}). Replacing time-varying parameters (\texttt{A11}) with constant parameters leads to the most substantial decline in accuracy of the \texttt{Hyper-Tree-AR} model, with MAPE nearly tripling and other metrics more than doubling compared to the base model. This result supports our conclusion that modeling parameters as functions of features is a critical aspect of our framework's forecasting capability.

\section{Framework Analysis and Considerations} \label{sec:framework_considerations}

The experiments presented in previous sections demonstrate the effectiveness of our Hyper-Tree framework across a range of forecasting tasks. In this section, we reflect on key architectural and conceptual aspects of Hyper-Trees, touching upon both their strengths and limitations. We focus on the following aspects: the conceptual differences between operating in parameter space versus function space, the inherent feature dependency characteristic of tree-based models and their implications for Hyper-Trees, the role of target model selection, the extrapolation capabilities that differentiate our approach from conventional tree-based forecasting models, and the constraints on parameter variation. These considerations help contextualize our empirical results and highlight practical implications for applying Hyper-Trees in real-world forecasting scenarios.

\subsection{Parameter vs. Function Space}

An important conceptual aspect of the Hyper-Tree framework is its operation in parameter space rather than function space, which introduces structural and interpretive advantages. In function space, each forecasted value represents a future time point, with potentially unbounded values ranging from negative to positive infinity. In contrast, the parameter space provides a more constrained and semantically meaningful setting. Parameters of classical time series models, such as ARIMA \citep{Box.2015} or Exponential Smoothing \citep{Holt.2004, Winters.1960}, correspond to specific temporal behaviors and carry inherent structural constraints. For instance, AR coefficients typically decay with increasing lag order, while smoothing parameters are bounded between 0 and 1. This structure aligns with known time series dynamics and introduces an inductive bias that likely supports more stable learning. A further advantage of operating in parameter space is the ability of Hyper-Trees to generate dynamic, time-varying parameters. Unlike traditional approaches where parameters remain fixed throughout the forecasting process, modeling parameters as functions of input features allows the framework to better adapt to evolving temporal patterns \citep{Lee.2023}. The parameter-centric modeling design also enhances interpretability and facilitates the incorporation of domain knowledge. Parameters with clear theoretical or practical meaning allow for easier validation, and domain-specific constraints can be readily encoded into the parameters, improving the model's ability to create plausible forecasts. Operating in parameter space also extends to the embeddings in the Hyper-TreeNet architecture. Rather than learning representations optimized for direct forecasting, the tree component generates embeddings specifically tailored for parameter estimation. These embeddings capture feature relationships most relevant for estimating time-varying coefficients, which the neural network then maps to the actual target model parameters. This ensures that both the tree embeddings and the network outputs operate within the same parameter space, maintaining consistency throughout the modeling process and facilitating effective learning of time-varying parameters.

\subsection{Feature Dependency}

Since parameter learning is based on GBDTs, Hyper-Trees inherit the characteristic feature-based partitioning of tree-based models. Unlike neural networks that can learn hierarchical representations from time series data through their multi-layer architectures, tree-based models require explicit features to capture relevant patterns. As with conventional GBDTs, the quality of features directly influences the Hyper-Tree's ability to learn meaningful parameter mappings. Previous research has shown that global models can benefit from operating on more homogeneous subsets of data \citep{Godahewa.2021b}, suggesting that effective feature-based partitioning can improve model performance. This is particularly important since global forecasting models often struggle with not being localized enough to particular series, especially in datasets with heterogeneous series. As such, any global tree-based model, including our Hyper-Tree framework, requires features that effectively identify similar patterns and create homogeneous groups. In operational contexts with rich feature sets, our framework demonstrates competitive accuracy. However, in scenarios with limited feature availability or where relevant information are not adequately captured in the features, any tree-based model faces limitations. This characteristic differentiates Hyper-Trees from deep learning approaches like Deep-AR \citep{Salinas.2020} or TFT \citep{Lim.2021}, which learn complex temporal representations through their recurrent or attention-based architectures. While our experiments show that Hyper-Trees perform well across various benchmarks with relatively simple feature sets, practitioners should be mindful of this requirement when applying our framework.

\subsection{Target Model Dependency}

A defining characteristic of the Hyper-Tree framework is its reliance on a pre-specified target model, whose parameters are learned as functions of input features. This design choice introduces a strong inductive bias that can be both an advantage and a limitation. The accuracy of Hyper-Trees is intrinsically bound to how well the selected target model aligns with the underlying dynamics of the data. Our experiments demonstrate that autoregressive (AR) \citep{Box.2015} and exponential smoothing (ETS) \citep{Holt.2004, Winters.1960} specifications perform well across a range of datasets. However, both impose structural assumptions that may not hold. For instance, AR models assume specific relationships between lagged observations, while ETS models impose specific assumptions on trend and seasonality components. When time series characteristics deviate from these assumptions, the limitation stems not from the Hyper-Tree framework itself, but from the choice of the target model. An inappropriate specification can impair forecast accuracy, underscoring the importance of careful model selection. As with any modeling approach, practitioners should select target models that reflect the underlying data-generating process. One might argue that models like Deep-AR \citep{Salinas.2020}, TFT \citep{Lim.2021} or LightGBM \citep{Ke.2017} are more flexible when applied to highly irregular operational time series, as they can learn arbitrary functions without imposing strict structural assumptions on the data-generating process. This may hold true in certain settings and datasets, as these models can capture a broad range of dependencies without being constrained to the inductive bias of a predefined target model. In principle, a sufficiently complex neural network or tree ensemble can approximate any data-generating process given enough data and model capacity. However, this generality may come at a cost. Without the guidance of a well-justified inductive bias that can guide forecasts toward temporal structures, these models may require substantially more data to learn patterns or may not explicitly account for structural temporal relationships that are well-understood in time series analysis. Hyper-Trees, by contrast, balance flexibility with structure, leveraging the interpretability and structural assumptions embedded in classical time series models with the flexible feature learning capabilities of tree-based approaches. 

The dependency on a suitably chosen target model becomes evident when forecasting specific types of time series. For example, intermittent time series, characterized by sporadic patterns with frequent zero values, pose significant forecasting challenges for conventional AR or ETS models. This is particularly relevant for datasets like the M5 data \citep{Makridakis.2022, Makridakis.2022b}, where the most disaggregated series exhibit highly intermittent patterns. Such series require specialized approaches that can accommodate both the frequency of zero values and the magnitude of non-zero occurrences. The modularity of our architecture, however, can be extended to incorporate more specialized target models. Future research could explore integrating models specifically designed for intermittent demand, e.g., Croston's method \citep{Croston.1972}. When such specialized models replace standard AR or ETS target models, the Hyper-Tree framework learns the relevant parameters, e.g., zero probability and demand size, as functions of features.

\subsection{Extrapolation Properties}

The inability of tree-based models to forecast values beyond the range of the training data is an important limitation to consider when evaluating their practical utility. This raises the natural question about the extrapolation properties of the Hyper-Tree framework and touches upon a fundamental characteristic of tree-based models, i.e., their piece-wise constant nature. This piece-wise constant constraint intersects with two main components of the Hyper-Tree architecture: the generation of model parameters and the creation of tree-embeddings. We first explore how our approach addresses the piece-wise constant limitations in parameter generation, followed by a discussion of tree-embeddings.

\paragraph{Parameter Generation} The Hyper-TreeNet variant mitigates this limitation through its network-based parameter generation. Unlike conventional tree-based models that create piece-wise constant outputs, the Hyper-TreeNet uses an MLP to generate the target model parameters. The tree-based component creates embeddings first which are then transformed by the MLP into target model parameters. Beyond the Hyper-TreeNet variant, Hyper-Tree models can benefit from the linear-tree extension introduced by \citet{Shi.2019}, which replaces piece-wise constant with piece-wise linear trees. Importantly, our ablation studies indicate that replacing linear with piece-wise constant trees has only marginal effects on forecast accuracy for both \texttt{Hyper-Tree-AR} and \texttt{Hyper-TreeNet-AR} models, with accuracy even slightly improving when using piece-wise constant parameter outputs for the former. This highlights a crucial distinction between our Hyper-Tree architecture and conventional GBDTs. Even when Hyper-Tree parameters are subject to piece-wise constant restrictions, the resulting forecasts can effectively extend beyond the training data range. This capability results from our approach's design principle of operating in parameter space, where parameters are first generated and then applied to the target model, preventing the piece-wise constant constraint from transferring to the forecasts themselves. Additionally, features at each time step translate into context-specific parameters, allowing the model to adapt to changing conditions even if the parameter values are constrained to values within the range learned during training.

\paragraph{Embedding Generation} While our Hyper-Tree framework addresses the extrapolation limitations of traditional GBDTs, it is important to acknowledge that the tree-based encoding component without the linear-tree option of \citet{Shi.2019} is subject to piece-wise constant constraints. In the Hyper-TreeNet architecture, the GBDT component partitions the feature space through hierarchical binary splits, creating regions where all observations within the same leaf node receive identical embedding values. This discretization limits tree embeddings to a finite number of distinct values determined by the number of leaf nodes across all trees. However, these discrete embeddings only serve as intermediate representations that are subsequently transformed by the MLP component, which effectively generates continuous parameter values. Additionally, the random projection matrix helps diversify the representations before MLP processing, further mitigating limitations of the piece-wise constant tree embeddings. Our ablation studies indicate that removing the linear-tree option \citep{Shi.2019} has only minor effects on embedding quality and overall forecast accuracy. Nevertheless, a more comprehensive analysis of how tree-generated embeddings impact forecast accuracy presents an important direction for future research.

\subsection{On Parameter Variation} \label{sec:param_constraints}

A potential concern with our time-varying parameter framework might be that, if coefficients were allowed to vary unconstrained over time, the model class could become arbitrarily flexible. Such a model could fit any training series perfectly, but at the cost of generalization and interpretability. Our Hyper-Tree framework mitigates this concern because parameters are functions of features, not unconstrained values that vary freely over time. More formally, Hyper-Trees model parameters as $\theta_{_{j, it}} = f_j(\mathbf{x}_{_{it}})$, where $\mathbf{x}_{_{it}}$ represents features, i.e., temporal indicators, as well as categorical and numeric covariates. This imposes a consistency constraint in the sense that similar feature contexts yield similar parameters. For instance, in the \texttt{Air Passengers} dataset with $\mathbf{x}_{_{it}} = \{\text{month}_t, \text{quarter}_t\}$ being the only features, all January observations across different years share identical parameters, as illustrated in Figure \ref{fig:params_airp}. This parameter consistency holds because all temporal features used across the experiments are date-derived (quarter, month, week, day-of-week, etc.) rather than raw time indices ($t=1, \ldots, T$), except for the STL-decomposition example in Section \ref{sec:local_single}. This ensures that parameter variation is tied to temporal patterns rather than to the passage of time itself. Further, the learned mapping $f(\cdot)$ is capacity-constrained through gradient boosting regularization, including tree depth and constraints on leaf size and leaf values. This keeps parameter variation tied to meaningful feature patterns rather than allowing for unconstrained coefficient paths. The same mechanism also preserves parameter interpretability. In contrast to an unconstrained specification with coefficients free to vary at each time step, where many parameter paths can yield indistinguishable forecasts, our feature-driven approach anchors parameters to observable covariates through the learned mapping $f(\cdot)$. Practitioners can examine how coefficients respond to specific conditions such as promotions, holidays, or store characteristics.\footnote{See Figures \ref{fig:params_airp}, \ref{fig:params_rossmann} and \ref{fig:params_m5} for examples.} 

Beyond these architectural considerations, our experimental results provide evidence that the time-varying parameters do not collapse into unconstrained flexibility. If Hyper-Trees were effectively learning arbitrary time-varying coefficient paths, one would expect poor out-of-sample accuracy. Instead, Hyper-Tree models consistently show strong accuracy across datasets and often outperform their static-parameter counterparts (Tables \ref{tab:airp_metrics}, \ref{tab:local_metrics} and \ref{tab:global_metrics}). This pattern indicates that the learned parameters capture genuine temporal structure rather than overfitting to training patterns.

\section{Conclusion and Future Research} \label{sec:conclusion}

With the Hyper-Tree framework, we have ventured into the comparatively unexplored use of gradient boosted trees for modeling the parameters of classical time series models. By repurposing GBDTs to learn the parameters of target models, we have combined the effectiveness of boosted trees on tabular data with the strengths of classical time series models. This approach naturally induces a time series inductive bias to tree models and addresses the limitations of conventional GBDT-based forecasts. We also suggest a novel approach to resolve the scaling issue of GBDTs, contributing to the growing literature on combining decision trees and neural networks. In this hybrid approach, the tree serves as an encoding layer, transforming features into informative representations, while the network functions as a decoder, mapping these embeddings to the target model parameters. Through comprehensive experiments, we demonstrate the effectiveness of our framework across several forecasting scenarios and datasets. Our findings demonstrate the potential of tree-based models for learning the parameters of target models, bridging the gap between traditional time series methods and modern gradient boosting approaches. This perspective opens new ways of understanding and using boosted tree models beyond their conventional applications in time series forecasting.

Looking forward, the horizon of future research in this domain is broad and inviting. While we have introduced an initial conception of tree-based hyper-models, there remain several compelling avenues for future investigation. Our current focus on relatively simple target models opens up opportunities to extend our approach to more complex architectures. Exploring more general state space formulations would offer a unified and flexible framework for modeling complex temporal dynamics, while more specialized target models tailored to specific time series characteristics, such as intermittent demand, pose another valuable extension. Extending the autoregressive target model to vector-autoregressive (VAR) formulations \citep{Sims.1980, Luetkepohl.2005} would enable the framework to capture cross-series dependencies, moving from univariate to multivariate forecasting. Also, extending the framework to include automated model and lag selection, along with sparsity-inducing penalties on target model parameters, would further enhance the adaptability of our Hyper-Tree models. Additionally, improving the runtime efficiency of the \texttt{Hyper-Tree-ETS} would broaden its applicability to larger-scale problems, making it a more versatile tool for diverse time series forecasting applications. A probabilistic extension of our approach also merits consideration. Incorporating uncertainty quantification into the Hyper-Tree framework would offer a more comprehensive understanding of the underlying time series dynamics. Adapting statistical inference frameworks for time-varying parameters to our Hyper-Tree context would be another valuable extension, enabling the construction of confidence intervals, hypothesis testing, and formal assessment of tree-based time-varying parameters. Besides extending the approach of \citet{Spuck.2025} to a time series context, Bayesian tree ensembles such as BART \citep{Chipman.2010} offer another potential direction for such inference, as replacing the GBDT component that learns the parameter mapping with a Bayesian formulation would provide posterior distributions over the target model parameters directly. The impact of gradient flow options on model behavior across different domains and datasets presents another interesting area for future exploration. Our ablation studies show minimal differences between shared and separated gradient approaches, but this may vary in other contexts. A more comprehensive analysis of tree-based embeddings and their representational capacity poses another interesting direction for future research. Beyond these extensions, combining gradient boosted trees and neural networks leverages the complementary strengths of both model classes, bridging tree-based and deep learning approaches. This integration points toward a class of hybrid models that open new avenues for flexible forecasting architectures.

\section*{Implementation} 
The code will be made available on the official repository \url{https://github.com/StatMixedML/Hyper-Trees} upon the final publication of this paper.

\newpage
\bibliography{literature, literature2}

\vfill
\section*{Declaration of Interest}
\noindent The authors declare no competing interests.

\newpage
\appendix
\begin{center}
	{\Large\textbf{Appendices}}
\end{center}

\vspace{1em}

\section{Notation} \label{app:notation}

\setcounter{table}{0}
\renewcommand{\thetable}{A\arabic{table}}

The following table summarizes the notation used throughout the paper.

{\fontsize{8pt}{7pt}\selectfont                                                                                                                                                                                                                                                 
	\begin{longtable}{ll}                                                                                                                                                                                                                                                       
		\caption{Paper Notation.} \label{tab:notation} \\                                                                                                                                                                                                                       
		\toprule                                                                                                                                                                                                                                                                
		\textbf{Symbol} & \textbf{Description} \\                                                                                                                                                                                                                               
		\midrule
		\endfirsthead
		
		\multicolumn{2}{c}{\tablename\ \thetable{} -- continued from previous page} \\
		\toprule
		\textbf{Symbol} & \textbf{Description} \\
		\midrule
		\endhead
		
		\midrule
		\multicolumn{2}{r}{\textit{continued on next page}} \\
		\endfoot
		
		\bottomrule
		\endlastfoot
		
		\multicolumn{2}{l}{\textit{Core components}} \\
		\midrule
		$\mathcal{T}$ & Hyper-Tree (GBDT component) \\
		$\mathcal{M}$ & Target time series model (e.g., AR($p$), ETS) \\
		$\mathcal{N}$ & Neural network (MLP) in the Hyper-TreeNet architecture \\
		$\mathcal{L}$ & Loss function \\
		$\mathcal{L}_{\scriptscriptstyle 2}$ & Mean squared error (MSE) loss \\
		$\mathcal{L}_{\text{split}}$ & Loss reduction from a candidate split \\
		$\tau_{\scriptscriptstyle 0.5}$ & Quantile loss evaluated at the median \\
		$\text{NLL}_{\scriptscriptstyle \text{T}}$ & Negative log-likelihood of a Student-T distribution \\
		\midrule
		\multicolumn{2}{l}{\textit{Data and indexing}} \\
		\multicolumn{2}{p{.95\linewidth}}{\vspace{0.5em} We use joint panel indexing $\{it\}$ (no comma) when the subscript identifies one observation (series $i$ at time $t$), including shifted time as in $y_{it-j}$. Commas separate a parameter label from the observation
			index, as in $\theta_{j,it}$. Local / single-series contexts drop the $i$ subscript.} \\
		\midrule
		$y_i$, $x_i$ & Target and feature vector for the $i$-th observation (generic GBDT context), $i = 1, \ldots, N$ \\
		$y_{it}$, $\mathbf{x}_{it}$ & Target and feature vector for series $i$ at time $t$ (panel / global time series context) \\
		$y_t$, $\mathbf{x}_t$ & Target and feature vector at time $t$ (local / single-series context) \\
		$X_{\scriptscriptstyle \text{Tree}}$ & Feature space of the Hyper-Tree \\
		$N$ & Number of observations, with index $i = 1, \ldots, N$ (distinct from the calligraphic $\mathcal{N}$)  \\
		$P$ & Number of target model parameters, with index $j = 1, \ldots, P$ \\
		$C$ & Dimension of the feature space \\
		$p$ & Number of AR lags \\
		$t$ & Time index, $t = 1, \ldots, T$ \\
		$T$ & Length of the time series (distinct from the calligraphic $\mathcal{T}$) \\
		$h$ & Forecast horizon \\
		$m$ & Boosting iteration, with $m = 1, \ldots, M$ \\
		$M$ & Total number of boosting iterations (distinct from the calligraphic $\mathcal{M}$) \\
		$f: \mathcal{X}^C \rightarrow \Theta^P$ & Learned global mapping from features to target model parameters \\
		\midrule
		\multicolumn{2}{l}{\textit{Tree outputs and updates}} \\
		\midrule
		$\hat{\psi}_{\scriptscriptstyle \mathcal{T}}$ & Output of the GBDT (tree output) \\
		$\hat{\psi}_{\scriptscriptstyle \mathcal{M}}$ & Forecast of the target time series model \\
		$\hat{\psi}^{(m)}_{i_{\scriptscriptstyle \mathcal{T}}}$ & Tree output for observation $i$ at iteration $m$ \\
		$\hat{\delta}^{(m)}_{i_{\scriptscriptstyle \mathcal{T}}}$ & Incremental update at iteration $m$ \\
		$\nu$ & Learning rate \\
		$q(\cdot)$ & Tree structure mapping input $x$ to leaf $j$ \\
		$w^{*}_{j}$ & Optimal weight assigned to leaf $j$ \\
		$\lambda$ & Regularization parameter \\
		$I_j$ & Index set of observations assigned to leaf $j$ \\
		$I_L$, $I_R$ & Index sets of left and right child nodes after a split \\
		$I = I_L \cup I_R$ & Union of left and right index sets \\
		\midrule
		\multicolumn{2}{l}{\textit{Gradients and Hessians}} \\
		\midrule
		$g_i$, $h_i$ & First- and second-order derivatives of $\mathcal{L}$ w.r.t. the tree output (conventional GBDT) \\
		$g_i^{(j)}$, $h_i^{(j)}$ & First- and second-order derivatives of $\mathcal{L}$ w.r.t. the $j$-th Hyper-Tree parameter \\
		$G_j$, $H_j$ & Aggregated gradients and Hessians across observations in leaf $j$ \\
		$K$ & Number of Gaussian probes in the Hutchinson diagonal Hessian estimator \\
		\newpage		
		\multicolumn{2}{l}{\textit{Target model parameters}} \\
		\multicolumn{2}{p{.95\linewidth}}{\vspace{0.5em} Unhatted parameter symbols denote the true time-varying parameters of the target model, modeled as functions of features. Hats denote Hyper-Tree estimates, or estimates used in forecasts.\vspace{0.5em}} \\
		\midrule
		$\theta_{\scriptscriptstyle \mathcal{T}}$,  $\hat{\theta}_{\scriptscriptstyle \mathcal{T}}$ & Target model parameters of the Hyper-Tree and their estimates  \\
		$\theta_{\scriptscriptstyle \mathcal{N}}$,  $\hat{\theta}_{\scriptscriptstyle \mathcal{N}}$ & Target model parameters of the Hyper-TreeNet (via the MLP) and their estimates \\
		$\hat{\theta}^{(j)}_{i_{\scriptscriptstyle \mathcal{T}}}$ & Estimated $j$-th target model parameter for observation $i$ \\
		$\theta_{j,it}(\mathbf{x}_{it})$ & AR($p$) coefficient for lag $j$, series $i$, time $t$ (panel / global), $j = 1, \ldots, p$ \\
		$\theta_{j,t}(\mathbf{x}_{t})$ & AR($p$) coefficient for lag $j$ at time $t$ (local / single-series), $j = 1, \ldots, p$ \\
		$\theta^{\mu}_{j,t}(\mathbf{x}_t)$ & Parameter of the mean component (distributional extension) \\
		$\theta^{\sigma}_{t}(\mathbf{x}_t)$ & Parameter of the standard deviation component (distributional extension) \\
		$\mu_t$, $\hat{\mu}_t$ & Mean of the target distribution and its estimate \\
		$\sigma_t$, $\hat{\sigma}_t$ & Standard deviation of the target distribution and its estimate \\
		$\hat{y}_{t+h|t}$ & $h$-step-ahead point forecast conditional on information at time $t$ \\
		\midrule
		\multicolumn{2}{l}{\textit{STL-specific parameters}} \\
		\midrule
		$a_{0,t}(\mathbf{x}_{t})$ & Intercept of the trend component (STL example) \\
		$a_{1,t}(\mathbf{x}_{t})$ & Slope of the trend component (STL example) \\
		$c_{r,t}(\mathbf{x}_{t})$ & Sine weight of the $r$-th seasonal component (STL example) \\
		$d_{r,t}(\mathbf{x}_{t})$ & Cosine weight of the $r$-th seasonal component (STL example) \\
		$r$ & Seasonal component index, $r = 1, \ldots, N_{\text{season}}$ \\
		$N_{\text{season}}$ & Number of seasonal components in the Fourier decomposition \\
		\midrule
		\multicolumn{2}{l}{\textit{ETS-specific parameters}} \\
		\midrule
		$\ell_t$, $\hat{\ell}_t$ & Level component and its estimate \\
		$b_t$, $\hat{b}_t$ & Trend component and its estimate \\
		$s_t$, $\hat{s}_t$ & Seasonality component and its estimate \\
		$\alpha_t(\mathbf{x}_t)$ & Level smoothing parameter \\
		$\beta_t(\mathbf{x}_t)$ & Trend smoothing parameter \\
		$\gamma_t(\mathbf{x}_t)$ & Seasonality smoothing parameter \\
		$\phi_t(\mathbf{x}_t)$, $\hat{\phi}_t(\mathbf{x}_t)$ & Trend dampening parameter and its estimate \\
		$m_{\text{season}}$ & Frequency of the seasonal cycle in the STL \& ETS specifications \\
		$k_{\text{season}}$ & Seasonality repeat index in the $h$-step-ahead ETS forecast \\
		$u$ & Summation index in the $h$-step-ahead ETS forecast, $u = 1, \ldots, h$ \\                                                                                                                                                                                                 
		$v$ & Product index for cumulative dampening, $v = 1, \ldots, u$ \\
		\midrule
		\multicolumn{2}{l}{\textit{Hyper-TreeNet components}} \\
		\midrule
		$\mathcal{E}_{\scriptscriptstyle \mathcal{T}}$, $\mathcal{\hat{E}}_{\scriptscriptstyle \mathcal{T}}$ & Tree embeddings generated by the GBDT component and its estimate\\
		$\mathbf{W}_{\text{proj}}$ & Random projection matrix, $\mathbf{W}_{\text{proj}} \in \mathbb{R}^{k \times d}$ \\
		$\mathbf{z}$ & Expanded representation passed to the MLP, $\mathbf{z} \in \mathbb{R}^{k \times N}$ \\
		$d$ & Dimension of tree embeddings \\
		$k$ & Output dimension of the projection (equal to number of target model parameters) \\
	\end{longtable}
}

\vfill

\section{Datasets}

\setcounter{table}{0}
\renewcommand{\thetable}{B\arabic{table}}

\begin{table}[th!]
	\fontsize{6pt}{7pt}\selectfont
	\begin{center}
		\caption{Dataset Descriptions.}
		\label{tab:data}
		\begin{tabular}{l||ccccccc}
		\toprule
		Dataset & \# Series & Frequency & \multicolumn{3}{c}{Series Length} & Lags & Forecast Horizon \\
		\cmidrule(lr){4-6}
		& & & min & avg & max & &  \\
		\midrule
			\texttt{Air Passengers} & $1$ & \textit{monthly} & $144$ & $144$ & $144$ & $12$ & $12$ \\
			\texttt{Australian Electricity Demand} & $5$ & \textit{monthly} & $158$ & $159$ & $160$ & $12$ & $24$ \\
			\texttt{Australian Retail Turnover} & $133$ & \textit{monthly} & $441$ & $441$ & $441$ & $12$ & $24$ \\	
			\texttt{M3 Monthly} & $1,428$ & \textit{monthly} & $66$ & $117$ & $144$ & $12$ & $18$ \\	
			\texttt{M3 Yearly} & $645$ & \textit{yearly} & $20$ & $28$ & $47$ & $3$ & $6$ \\
			\texttt{M5} & $70$ & \textit{daily}  & $1,968$ & $1,969$ & $1,969$ & $14$ & $28$ \\	
			\texttt{Rossmann Store Sales} & $1,115$ & \textit{daily} & $592$ & $757$ & $942$ & $21$ & $40$ \\
			\texttt{Tourism} & $366$ & \textit{monthly} & $91$ & $299$ & $333$ & $12$ & $24$ \\
			\bottomrule
		\end{tabular}
	\end{center}
\end{table}

\section{Hyper-Parameters}

\setcounter{table}{0}
\renewcommand{\thetable}{C\arabic{table}}

\texttt{Hyper-Tree-AR}, \texttt{Hyper-TreeNet-AR}, and \texttt{Hyper-Tree-ETS} additionally use \texttt{deterministic=True} and \texttt{force\_row\_wise=True} to ensure reproducibility across runs, though these settings may increase training time.\footnote{See                                  \url{https://lightgbm.readthedocs.io/en/latest/Parameters.html} for details.} For the \texttt{Hyper-Tree-ETS} model, the Hutchinson diagonal estimator uses $K=5$ Gaussian probes across all experiments.

\vfill

\begin{TableNotes}%
	\tiny
	\setlength{\leftmargin}{0pt}
	\setlength{\labelsep}{0pt}
	\setlength{\labelwidth}{0pt}
	\setlength{\itemindent}{0pt}
	\item[] Unlisted hyper-parameters use default values. $\mathcal{L}_{_{2}}$ denotes MSE-Loss. Entries with '--' indicate that the model was not trained.
\end{TableNotes}

{\fontsize{6pt}{7pt}\selectfont
	\begin{longtable}{@{}l||ccccccc@{}}
		\caption{Local Model Hyper-Parameters.} \label{tab:local_hyperparams}  \\
		\toprule
		\multicolumn{8}{c}{\textbf{Air Passengers}} \\
		\midrule
		Model & Learning Rate & Linear Tree & Epochs & Loss \\
		\midrule
		\texttt{Hyper-Tree-AR} & \textit{1e-01} & \textit{True} & \textit{100} & $\mathcal{L}_{_{2}}$ \\
		\texttt{Hyper-Tree-ETS} & \textit{1e-03} & \textit{False} & \textit{100} & $\mathcal{L}_{_{2}}$ \\
		\texttt{Hyper-TreeNet-AR} & \textit{1e-01} & \textit{True} & \textit{100} & $\mathcal{L}_{_{2}}$ \\
		\texttt{LightGBM} & \textit{1e-01} & \textit{True} & \textit{100} & $\mathcal{L}_{_{2}}$ \\
		\texttt{LightGBM-AR} & \textit{1e-01} & \textit{True} & \textit{100} & $\mathcal{L}_{_{2}}$ \\
		\texttt{LightGBM-STL} & \textit{1e-01} & \textit{True} & \textit{100} & $\mathcal{L}_{_{2}}$ \\
		\midrule		
		\multicolumn{8}{c}{\textbf{Australian Electricity Demand}} \\
		\midrule
		Model & Learning Rate & Linear Tree & Epochs & Loss \\
		\midrule
		\texttt{Hyper-Tree-AR} & \textit{1e-01} & \textit{True} & \textit{100} & $\mathcal{L}_{_{2}}$ \\
		\texttt{Hyper-Tree-ETS} & \textit{1e-01} & \textit{False} & \textit{100} & $\mathcal{L}_{_{2}}$ \\
		\texttt{Hyper-TreeNet-AR} & \textit{1e-01} & \textit{True} & \textit{100} & $\mathcal{L}_{_{2}}$ \\
		\texttt{LightGBM} & \textit{1e-01} & \textit{True} & \textit{100} & $\mathcal{L}_{_{2}}$ \\
		\texttt{LightGBM-AR} & \textit{1e-01} & \textit{True} & \textit{100} & $\mathcal{L}_{_{2}}$ \\
		\texttt{LightGBM-STL} & \textit{1e-01} & \textit{True} & \textit{100} & $\mathcal{L}_{_{2}}$ \\
		\midrule
		\multicolumn{8}{c}{\textbf{Australian Retail Turnover}} \\
		\midrule
		Model & Learning Rate & Linear Tree & Epochs & Loss \\
		\midrule
		\texttt{Hyper-Tree-AR} & \textit{1e-01} & \textit{True} & \textit{100} & $\mathcal{L}_{_{2}}$ \\
		\texttt{Hyper-Tree-ETS} & -- & -- & -- & -- \\
		\texttt{Hyper-TreeNet-AR} & \textit{1e-01} & \textit{True} & \textit{100} & $\mathcal{L}_{_{2}}$ \\
		\texttt{LightGBM} & \textit{1e-01} & \textit{True} & \textit{100} & $\mathcal{L}_{_{2}}$ \\
		\texttt{LightGBM-AR} & \textit{1e-01} & \textit{True} & \textit{100} & $\mathcal{L}_{_{2}}$ \\
		\texttt{LightGBM-STL} & \textit{1e-01} & \textit{True} & \textit{100} & $\mathcal{L}_{_{2}}$ \\
		\midrule	
		\multicolumn{8}{c}{\textbf{Tourism}} \\
		\midrule
		Model & Learning Rate & Linear Tree & Epochs & Loss \\
		\midrule
		\texttt{Hyper-Tree-AR} & \textit{1e-01} & \textit{True} & \textit{100} & $\mathcal{L}_{_{2}}$ \\
		\texttt{Hyper-Tree-ETS} & -- & -- & -- & -- \\
		\texttt{Hyper-TreeNet-AR} & \textit{1e-01} & \textit{True} & \textit{100} & $\mathcal{L}_{_{2}}$ \\
		\texttt{LightGBM} & \textit{1e-01} & \textit{True} & \textit{100} & $\mathcal{L}_{_{2}}$ \\
		\texttt{LightGBM-AR} & \textit{1e-01} & \textit{True} & \textit{100} & $\mathcal{L}_{_{2}}$ \\
		\texttt{LightGBM-STL} & \textit{1e-01} & \textit{True} & \textit{100} & $\mathcal{L}_{_{2}}$ \\
		\bottomrule
		\insertTableNotes
	\end{longtable}
}

\begin{TableNotes}%
	\tiny
	\setlength{\leftmargin}{0pt}
	\setlength{\labelsep}{0pt}
	\setlength{\labelwidth}{0pt}
	\setlength{\itemindent}{0pt}
	\item[] \textit{H}: Forecast Horizon, $\mathcal{L}_{_{2}}$: MSE-Loss, \textit{NLL$_{\fontsize{3pt}{3pt}\selectfont \text{T}}$}: Negative log-likelihood of Student-T distribution, $\tau_{_{0.5}}$: quantile loss evaluated at median. For \texttt{Hyper-TreeNet-AR}, the learning rate shows tree-part / MLP-part. Unlisted hyper-parameters use default values. Entries with '--' indicate that the model was not trained. For the \texttt{M3 Yearly} dataset, the \texttt{Hyper-Tree-ETS} parameterizes an ETS model with a linear trend only, as indicated by '$^\ast$'. For all other applicable datasets, the \texttt{Hyper-Tree-ETS} generates parameters for ETS models with multiplicative seasonality and a damped trend.
\end{TableNotes}

\newpage

{\fontsize{5pt}{6pt}\selectfont
\begin{longtable}{@{}l||ccccccccc@{}}
	\caption{Global Model Hyper-Parameters.} \label{tab:global_hyperparams}  \\
	\toprule
	
	\multicolumn{10}{c}{\textbf{Australian Electricity Demand}} \\
	\midrule
	Model & Learning Rate & Linear Tree & Epochs & Hidden Layers & Batch Size & \# Layers & \# Heads & Context Length & Loss \\ 
	\midrule
	\texttt{Deep-AR} & \textit{1e-03} & -- & \textit{100} & \textit{128} & \textit{128} & \textit{3} & -- & \textit{2h} & \textit{NLL$_{\fontsize{3pt}{3pt}\selectfont \text{T}}$} \\
	\texttt{Hyper-Tree-AR} & \textit{1e-01} & \textit{True} & \textit{100} & -- & -- & --  & -- & -- & $\mathcal{L}_{_{2}}$ \\
	\texttt{Hyper-Tree-ETS} & \textit{1e-02} & \textit{False} & \textit{100} & -- & -- & --  & -- & -- & $\mathcal{L}_{_{2}}$ \\
	\texttt{Hyper-TreeNet-AR} & \textit{1e-01 / 1e-03} & \textit{True} & \textit{100} & \textit{128} & -- & --  & -- & -- & $\mathcal{L}_{_{2}}$ \\
	\texttt{LightGBM} & \textit{1e-01} & \textit{True} & \textit{100} & -- & -- & -- & --  & -- & $\mathcal{L}_{_{2}}$ \\
	\texttt{LightGBM-AR} & \textit{1e-01} & \textit{True} & \textit{100} & -- & -- & -- & -- & -- & $\mathcal{L}_{_{2}}$ \\
	\texttt{TFT} & \textit{1e-03} & -- & \textit{100} & \textit{128} & \textit{128} & --  & \textit{4} & \textit{2h} &  $\tau_{_{0.5}}$ \\
	\midrule
	
	\multicolumn{10}{c}{\textbf{Australian Retail Turnover}} \\
	\midrule
	Model & Learning Rate & Linear Tree & Epochs & Hidden Layers & Batch Size & \# Layers & \# Heads & Context Length & Loss \\ 
	\midrule
	\texttt{Deep-AR} & \textit{1e-03} & -- & \textit{500} & \textit{128} & \textit{128} & \textit{3} & -- & \textit{2h} & \textit{NLL$_{\fontsize{3pt}{3pt}\selectfont \text{T}}$} \\
	\texttt{Hyper-Tree-AR} & \textit{1e-01} & \textit{True} & \textit{500} & -- & -- & --  & -- & -- & $\mathcal{L}_{_{2}}$ \\
	\texttt{Hyper-Tree-ETS} & \textit{1e-01} & \textit{True} & \textit{100} & -- & -- & --  & -- & -- & $\mathcal{L}_{_{2}}$ \\
	\texttt{Hyper-TreeNet-AR} & \textit{1e-01 / 1e-03} & \textit{True} & \textit{500} & \textit{128} & -- & --  & -- & -- &  $\mathcal{L}_{_{2}}$ \\
	\texttt{LightGBM} & \textit{1e-01} & \textit{True} & \textit{200} & -- & -- & -- & --  & -- & $\mathcal{L}_{_{2}}$ \\
	\texttt{LightGBM-AR} & \textit{1e-01} & \textit{True} & \textit{200} & -- & -- & -- & -- & -- &  $\mathcal{L}_{_{2}}$ \\
	\texttt{TFT} & \textit{1e-03} & -- & \textit{500} & \textit{128} & \textit{128} & --  & \textit{4} & \textit{2h} & $\tau_{_{0.5}}$ \\
	\midrule		
	\multicolumn{10}{c}{\textbf{M3 Monthly}} \\
	\midrule
	Model & Learning Rate & Linear Tree & Epochs & Hidden Layers & Batch Size & \# Layers & \# Heads & Context Length & Loss \\ 	
	\midrule
	
	\texttt{Deep-AR} & \textit{1e-03} & -- & \textit{200} & \textit{128} & \textit{128} & \textit{3} & -- & \textit{2h} & \textit{NLL$_{\fontsize{3pt}{3pt}\selectfont \text{T}}$} \\
	\texttt{Hyper-Tree-AR} & \textit{1e-01} & \textit{True} & \textit{200} & -- & -- & --  & -- & -- & $\mathcal{L}_{_{2}}$ \\
	\texttt{Hyper-Tree-ETS} & -- & -- & -- & -- & -- & --  & -- & -- & -- \\
	\texttt{Hyper-TreeNet-AR} & \textit{1e-01 / 1e-03} & \textit{True} & \textit{200} & \textit{128} & -- & --  & -- & -- &  $\mathcal{L}_{_{2}}$ \\
	\texttt{LightGBM} & \textit{1e-01} & \textit{False} & \textit{500} & -- & -- & -- & --  & -- & $\mathcal{L}_{_{2}}$ \\
	\texttt{LightGBM-AR} & \textit{1e-01} & \textit{False} & \textit{500} & -- & -- & -- & -- & -- &  $\mathcal{L}_{_{2}}$ \\
	\texttt{TFT} & \textit{1e-03} & -- & \textit{200} & \textit{128} & \textit{128} & --  & \textit{4} & \textit{2h} & $\tau_{_{0.5}}$ \\
	\midrule	
			
	\multicolumn{10}{c}{\textbf{M3 Yearly}} \\
	\midrule
	Model & Learning Rate & Linear Tree & Epochs & Hidden Layers & Batch Size & \# Layers & \# Heads & Context Length & Loss \\ 
	\midrule
	\texttt{Deep-AR} & \textit{1e-03} & -- & \textit{100} & \textit{128} & \textit{128} & \textit{3} & -- & \textit{2h} & \textit{NLL$_{\fontsize{3pt}{3pt}\selectfont \text{T}}$} \\
	\texttt{Hyper-Tree-AR} & \textit{1e-02} & \textit{True} & \textit{100} & -- & -- & --  & -- & -- & $\mathcal{L}_{_{2}}$ \\
	\texttt{Hyper-Tree-ETS$^\ast$} & \textit{1e-01} & \textit{False} & \textit{100} & -- & -- & --  & -- & -- & $\mathcal{L}_{_{2}}$ \\
	\texttt{Hyper-TreeNet-AR} & \textit{1e-03 / 1e-03} & \textit{True} & \textit{100} & \textit{128} & -- & --  & -- & -- &  $\mathcal{L}_{_{2}}$ \\
	\texttt{LightGBM} & \textit{1e-01} & \textit{True} & \textit{100} & -- & -- & -- & --  & -- & $\mathcal{L}_{_{2}}$ \\
	\texttt{LightGBM-AR} & \textit{1e-01} & \textit{True} & \textit{100} & -- & -- & -- & -- & -- &  $\mathcal{L}_{_{2}}$ \\
	\texttt{TFT} & \textit{1e-03} & -- & \textit{100} & \textit{128} & \textit{128} & --  & \textit{4} & \textit{2h} & $\tau_{_{0.5}}$ \\
	\midrule	
	
	\multicolumn{10}{c}{\textbf{M5}} \\
	\midrule
	Model & Learning Rate & Linear Tree & Epochs & Hidden Layers & Batch Size & \# Layers & \# Heads & Context Length & Loss \\ 
	\midrule
	\texttt{Deep-AR} & \textit{1e-03} & -- & \textit{100} & \textit{128} & \textit{128} & \textit{3} & -- & \textit{2h} & \textit{NLL$_{\fontsize{3pt}{3pt}\selectfont \text{T}}$} \\
	\texttt{Hyper-Tree-AR} & \textit{1e-02} & \textit{True} & \textit{100} & -- & -- & --  & -- & -- & $\mathcal{L}_{_{2}}$ \\
	\texttt{Hyper-Tree-ETS} & -- & -- & -- & -- & -- & --  & -- & -- & -- \\
	\texttt{Hyper-TreeNet-AR} & \textit{1e-02 / 1e-03} & \textit{True} & \textit{100} & \textit{128} & -- & --  & -- & -- & $\mathcal{L}_{_{2}}$ \\
	\texttt{LightGBM} & \textit{1e-01} & \textit{True} & \textit{100} & -- & -- & -- & --  & -- & $\mathcal{L}_{_{2}}$ \\
	\texttt{LightGBM-AR} & \textit{1e-01} & \textit{True} & \textit{100} & -- & -- & -- & -- & -- & $\mathcal{L}_{_{2}}$ \\
	\texttt{TFT} & \textit{1e-03} & -- & \textit{100} & \textit{128} & \textit{128} & --  & \textit{4} &  \textit{2h} & $\tau_{_{0.5}}$ \\
	\midrule	
	
	\multicolumn{10}{c}{\textbf{Rossmann Store Sales}} \\
	\midrule
	Model & Learning Rate & Linear Tree & Epochs & Hidden Layers & Batch Size & \# Layers & \# Heads & Context Length & Loss \\ 
	\midrule
	\texttt{Deep-AR} & \textit{1e-03} & -- & \textit{500} & \textit{128} & \textit{128} & \textit{3} & -- & \textit{2h} & \textit{NLL$_{\fontsize{3pt}{3pt}\selectfont \text{T}}$} \\
	\texttt{Hyper-Tree-AR} & \textit{1e-02} & \textit{True} & \textit{500} & -- & -- & --  & -- & -- & $\mathcal{L}_{_{2}}$ \\
	\texttt{Hyper-Tree-ETS} & -- & -- & -- & -- & -- & --  & -- & -- & -- \\
	\texttt{Hyper-TreeNet-AR} & \textit{1e-02 / 1e-03} & \textit{True} & \textit{500} & \textit{128} & -- & --  & -- & -- & $\mathcal{L}_{_{2}}$ \\
	\texttt{LightGBM} & \textit{1e-01} & \textit{True} & \textit{500} & -- & -- & -- & --  & -- & $\mathcal{L}_{_{2}}$ \\
	\texttt{LightGBM-AR} & \textit{1e-01} & \textit{True} & \textit{500} & -- & -- & -- & -- & -- & $\mathcal{L}_{_{2}}$ \\
	\texttt{TFT} & \textit{1e-03} & -- & \textit{500} & \textit{128} & \textit{128} & --  & \textit{4} & \textit{2h} &  $\tau_{_{0.5}}$ \\
	\midrule
	\multicolumn{10}{c}{\textbf{Tourism}} \\
	\midrule
	Model & Learning Rate & Linear Tree & Epochs & Hidden Layers & Batch Size & \# Layers & \# Heads & Context Length & Loss \\ 
	\midrule
	\texttt{Deep-AR} & \textit{1e-03} & -- & \textit{500} & \textit{128} & \textit{128} & \textit{3} & -- & \textit{2h} & \textit{NLL$_{\fontsize{3pt}{3pt}\selectfont \text{T}}$} \\
	\texttt{Hyper-Tree-AR} & \textit{1e-01} & \textit{True} & \textit{500} & -- & -- & --  & -- & -- & $\mathcal{L}_{_{2}}$ \\
	\texttt{Hyper-Tree-ETS} & \textit{2e-02} & \textit{True} & \textit{100} & -- & -- & --  & -- & -- & $\mathcal{L}_{_{2}}$ \\
	\texttt{Hyper-TreeNet-AR} & \textit{1e-01 / 1e-03} & \textit{True} & \textit{500} & \textit{128} & -- & --  & -- & -- & $\mathcal{L}_{_{2}}$ \\
	\texttt{LightGBM} & \textit{1e-01} & \textit{True} & \textit{500} & -- & -- & -- & --  & -- & $\mathcal{L}_{_{2}}$ \\
	\texttt{LightGBM-AR} & \textit{1e-01} & \textit{True} & \textit{500} & -- & -- & -- & -- & -- & $\mathcal{L}_{_{2}}$ \\
	\texttt{TFT} & \textit{1e-03} & -- & \textit{500} & \textit{128} & \textit{128} & --  & \textit{4} & \textit{2h} &  $\tau_{_{0.5}}$ \\
	\bottomrule
	\insertTableNotes
\end{longtable}
}

\newpage
\section{Example Forecasts}

\setcounter{figure}{0}
\renewcommand{\thefigure}{D\arabic{figure}}

\begin{figure}[h!]
	\fontsize{7pt}{8pt}\selectfont
	\centering
	\caption{Global Model Forecasts.}
	\includegraphics[width=1.0\linewidth]{example_forecasts1.pdf}
	\label{fig:sample_fcsts}
\end{figure}

\clearpage  

\begin{figure}[h!]
	\fontsize{7pt}{8pt}\selectfont
	\ContinuedFloat
	\centering
	\caption*{Figure \thefigure\ (continued).}
	\includegraphics[width=1.0\linewidth]{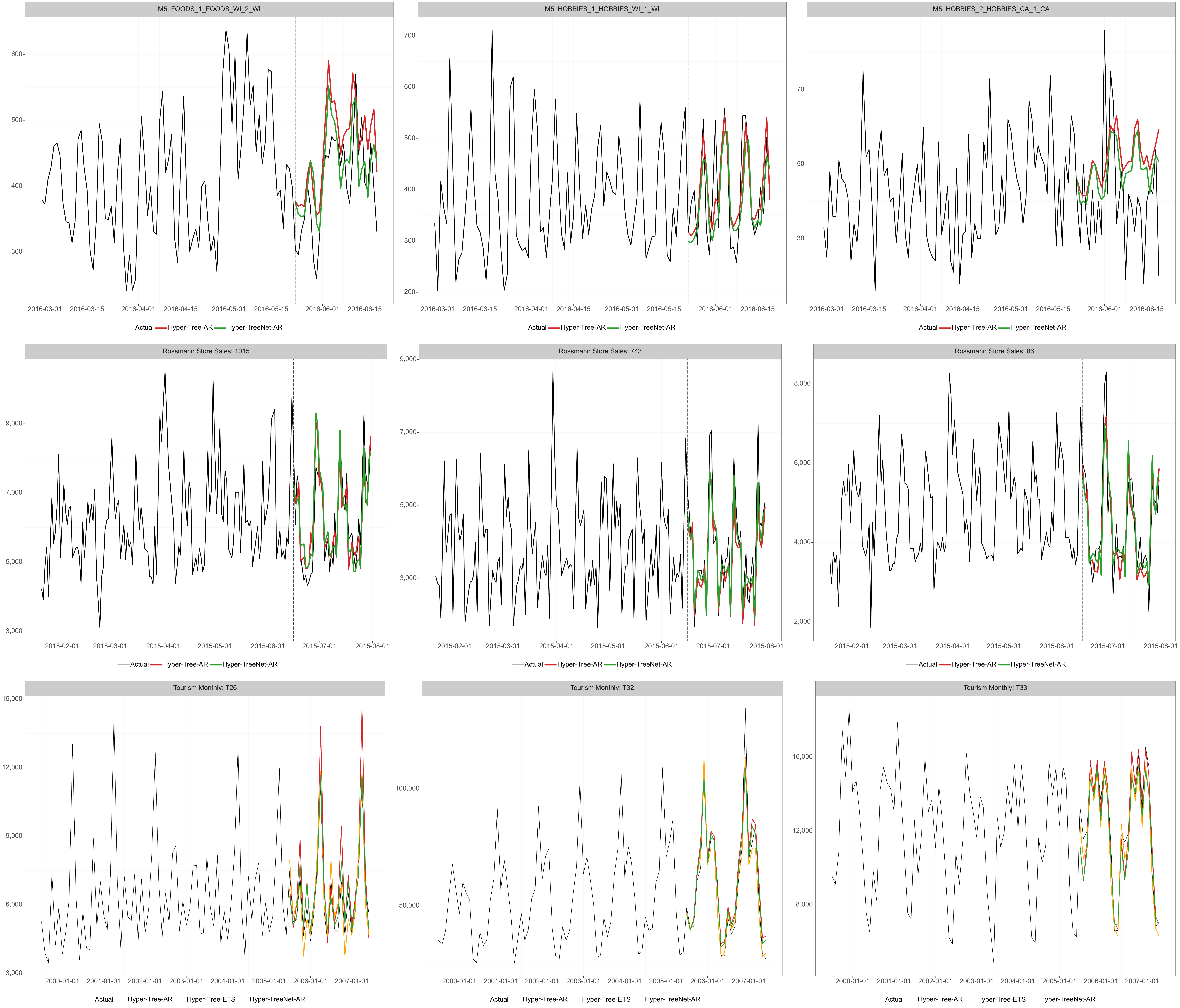}  
\end{figure}

\newpage
\section{Gradient Flow Strategies in Hybrid Tree-Network Models} \label{sec:gradients}
\setcounter{figure}{0}
\setcounter{table}{0}
\renewcommand{\thefigure}{E\arabic{figure}}
\renewcommand{\thetable}{E\arabic{table}}

When integrating GBDTs with neural networks in our Hyper-TreeNet architecture, it is important to consider how to reconcile their different learning paradigms. GBDTs operate in a discrete optimization space with threshold-based splits creating piece-wise constant regions, while neural networks optimize in a continuous parameter space with smooth gradient fields. Combining them into a joint framework requires careful consideration of how gradient information flows between the model components during training. In the following, we outline and discuss two possible approaches, each with different implications for the learning characteristics of the individual components: a fully shared gradient flow between the network and GBDT (\textit{Option 1}) and parameter updates within the same iteration but with separated gradient flows (\textit{Option 2}). Both options are sketched in Table \ref{tab:grads} with pseudo-codes and exemplified for a \texttt{Hyper-TreeNet-AR($p$)} model.

\begin{table}[h!]
	\centering
	\caption{Pseudocode of two Estimation Strategies for \texttt{Hyper-TreeNet-AR($p$)}: Shared vs. Separate Gradient Flows. \vspace{-1em}}	
	\label{tab:grads}	
	
	\begin{multicols}{2}		
		\begin{algorithm}[H]
			\scriptsize			
			\captionsetup{name=Option}
			\caption{Shared Gradient}
			\begin{algorithmic}
				\scriptsize
				\Require features, target, lags, gbdt, network
				
				\For{\texttt{iter = 1 \textbf{to} num\_iter}}:
				\State \hspace{0.5em} \textbf{GBDT Embeddings}
				\State \hspace{1em} gbdt\_embeds = gbdt(features)
				
				\State \hspace{0.5em} \textbf{Update Network}
				\State \hspace{1em} optimizer.zero\_grad()
				
				\State \hspace{1em} ar\_params = network(gbdt\_embeds)
				\State \hspace{1em} forecasts = forecast(ar\_params, lags) 
				\State \hspace{1em} loss = loss\_fn(forecasts, target)
				
				\State \hspace{1em} loss.backward()
				
				\State \hspace{0.5em} \textbf{Update GBDT}
				\State \hspace{1em} grad, hess = derivatives(loss, gbdt\_embeds)
				\State \hspace{1em} optimizer.step()	
				
				\EndFor

				\State
				\State
				\State
				\State
				\State
				
			\end{algorithmic}
		\end{algorithm}

		\columnbreak
		
		\begin{algorithm}[H]
			\scriptsize
			\captionsetup{name=Option}
			\caption{Separate Gradient}
			\begin{algorithmic}
				\scriptsize
				\Require features, target, lags, gbdt, network
				\For{\texttt{iter = 1 \textbf{to} num\_iter}}:
				\State \hspace{0.5em} \textbf{GBDT Embeddings}
				\State \hspace{1em} gbdt\_embeds = gbdt(features)
				
				\State \hspace{0.5em} \textbf{Update Network}
				\State \hspace{1em} network.train()
				\State \hspace{1em} optimizer.zero\_grad()
				\State \hspace{1em} ar\_params = network(gbdt\_embeds)
				\State \hspace{1em} forecasts\_net = forecast(ar\_params, lags) 
				\State \hspace{1em} loss\_net = loss\_fn(forecasts\_net, target)
				\State \hspace{1em} loss\_net.backward()
				\State \hspace{1em} optimizer.step()
				
				\State \hspace{0.5em} \textbf{Update GBDT}
				\State \hspace{1em} network.eval() \Comment{Inference mode}
				\State \hspace{1em} ar\_params = network(gbdt\_embeds)
				\State \hspace{1em} forecasts\_gbdt = forecast(ar\_params, lags) 
				\State \hspace{1em} loss\_gbdt = loss\_fn(forecasts\_gbdt, target)
				\State \hspace{1em} grad, hess = derivatives(loss\_gbdt, gbdt\_embeds)
				
				\EndFor
			\end{algorithmic}
		\end{algorithm}
		
	\end{multicols}
\end{table}

For \textit{Option 1}, the gradients flow through the GBDT and network in a single backpropagation pass. This provides a fully integrated optimization where both the tree and the network are updated jointly. While this approach offers conceptual elegance through its complete end-to-end training, it may not preserve the distinctive characteristics of each model component. Forcing these different learning paradigms into a single gradient flow could compromise the tree's discrete decision boundaries and the network's continuous optimization landscape, potentially diminishing their respective strengths. For \textit{Option 2}, each iteration alternates between two optimization phases: the network parameters are updated first using the current GBDT embeddings, followed by calculating gradients and Hessians for the GBDT using the updated network parameters. This separation of gradient flows reduces the risk of neural network gradients interfering with GBDT updates, allowing each model to maintain its distinctive strengths.  While \textit{Option 2} deviates from a full joint optimization, it preserves the core concept of integrated training. The separation potentially enables each component to better leverage its strengths while still benefiting from iterative feedback between them. The models still inform each other during each iteration, with the GBDT generating informative parameter-space representations from the input features that the network maps to target model parameters. The mutual learning process continues throughout training as GBDT embeddings influence the network's training and the network's state guides GBDT updates in each iteration. While both gradient flow options are valid approaches to joint optimization, we default to \textit{Option 2} for all experiments, allowing each model component, GBDT and neural network, to maintain its distinct characteristics while still benefiting from the iterative feedback between them, though users can also select the fully integrated gradient approach in the model implementation.

\newpage
\section{Time-Varying Parameters}

\setcounter{figure}{0}
\renewcommand{\thefigure}{F\arabic{figure}}

A defining aspect of our Hyper-Tree framework is its ability to estimate time-varying parameters. Unlike conventional time series models that estimate a set of constant, non-varying coefficients, Hyper-Trees dynamically adapt parameters as a function of features, enabling them to respond to evolving environments. This is illustrated in the following figures with visualizations of estimated and forecasted AR-coefficients for the \texttt{Air Passengers}, \texttt{Rossmann Store Sales} and \texttt{M5} datasets. For clarity, we display only a subset of each series' history, rather than the full series length. 

\begin{figure}[h!]
	\fontsize{7pt}{8pt}\selectfont
	\centering
	\caption{Time-Varying AR(12) Parameters for \texttt{Air Passengers} Dataset.}
	\includegraphics[width=1.0\linewidth]{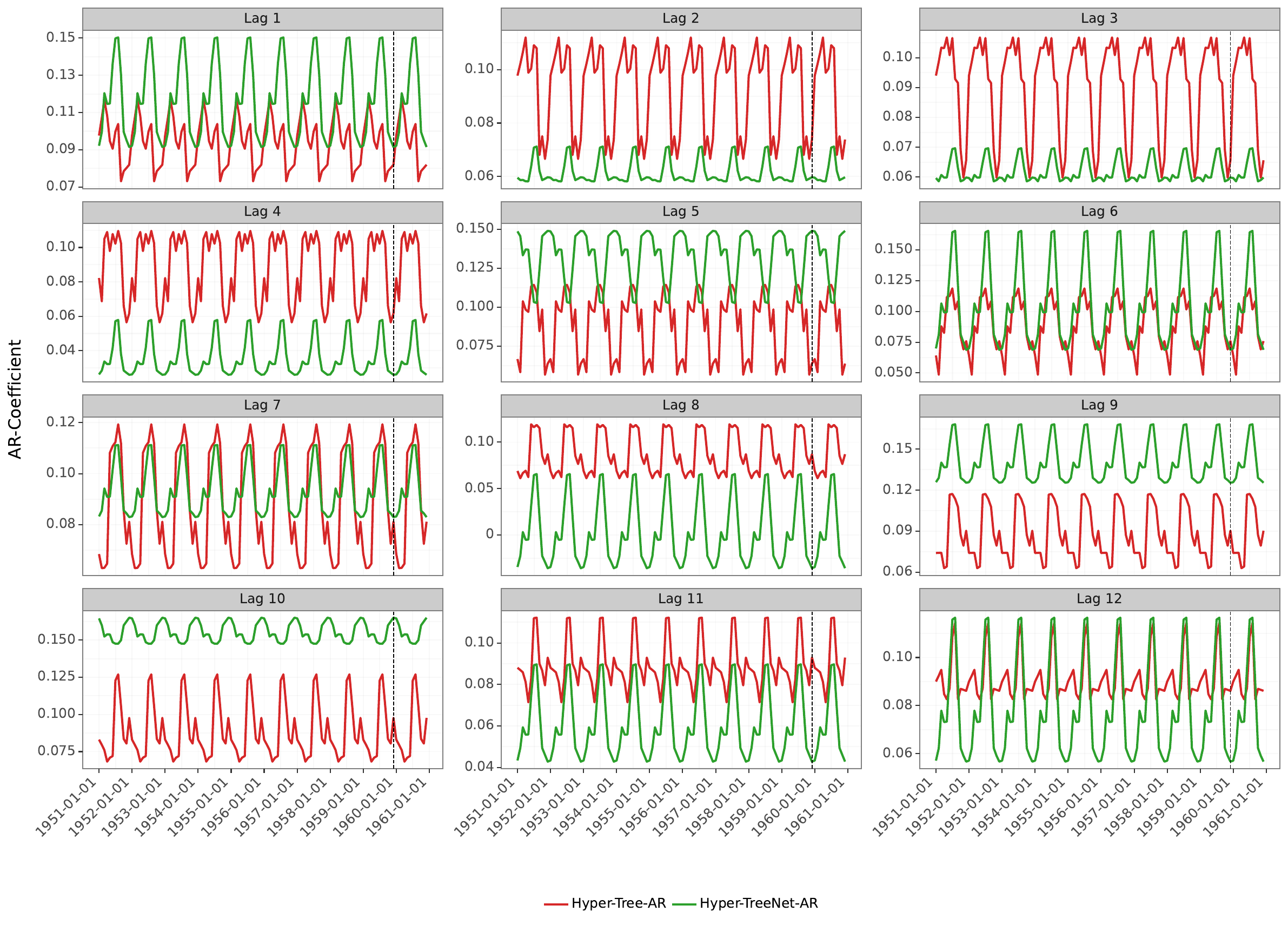}
	\label{fig:params_airp}
\end{figure}

The pronounced seasonal patterns of AR-parameters in Figure \ref{fig:params_airp} show how the models adapt to cyclical behaviors in the underlying series. For the \texttt{Rossmann Store Sales} and \texttt{M5} datasets, the influence of features is visible in the parameter plots, where distinct shifts and variations correspond to changes in feature values. These parameters continue to vary for the forecast horizon (indicated by dotted vertical lines), demonstrating how our approach maintains adaptivity beyond the training data. This dynamic parameterization allows Hyper-Trees to effectively respond to changing environments without requiring model re-training, providing a distinctive capability over conventional approaches with fixed parameter estimates \citep{Lee.2023}.

\newpage

\begin{figure}[h!]
	\fontsize{7pt}{8pt}\selectfont
	\centering
	\caption{Time-Varying AR(21) Parameters for \texttt{Rossmann Store Sales} Dataset.}
	\includegraphics[width=1.0\linewidth]{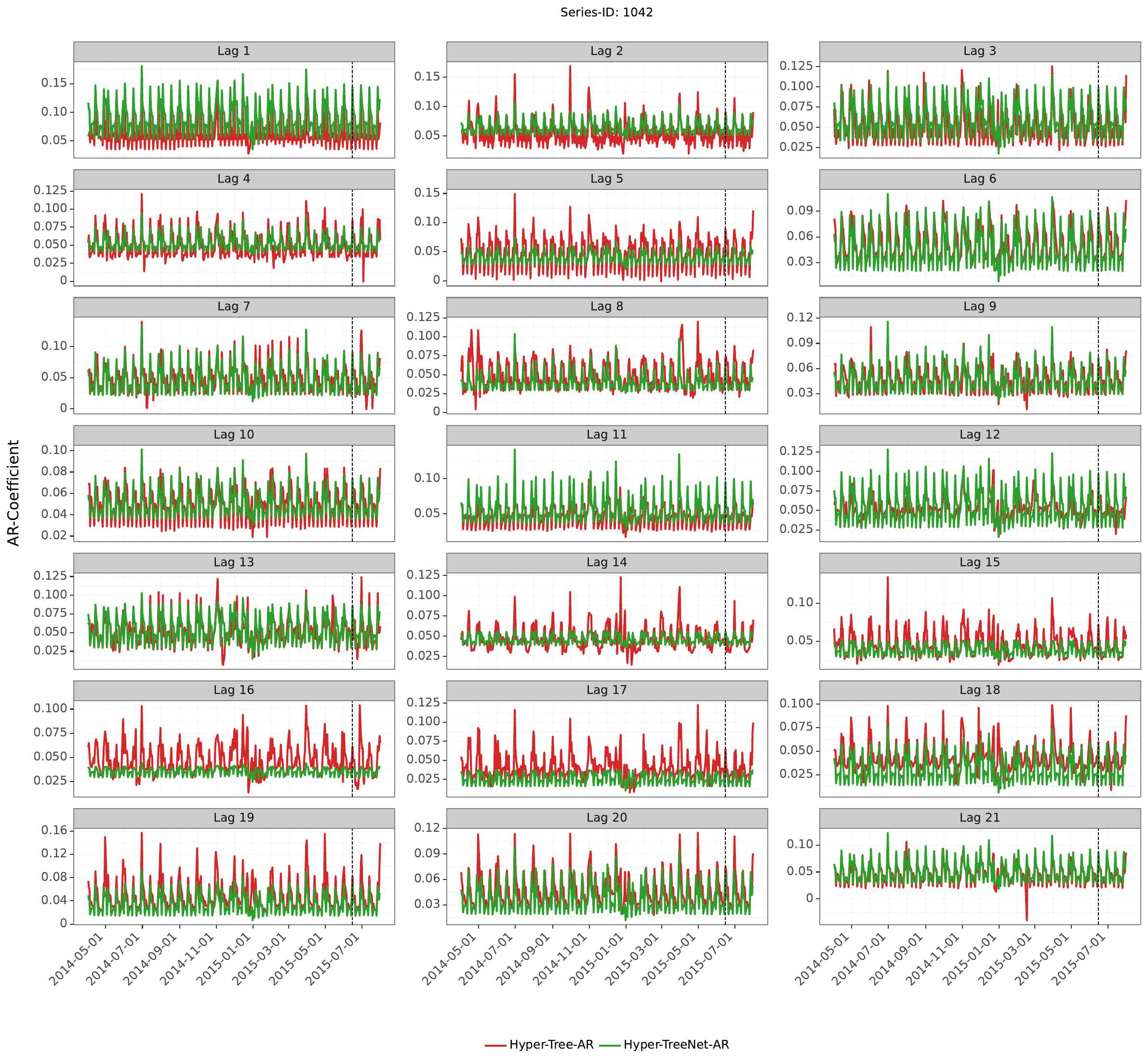}
	\label{fig:params_rossmann}
\end{figure}

\newpage

\begin{figure}[h!]
	\fontsize{7pt}{8pt}\selectfont
	\centering
	\caption{Time-Varying AR(14) Parameters for \texttt{M5} Dataset.}
	\includegraphics[width=1.0\linewidth]{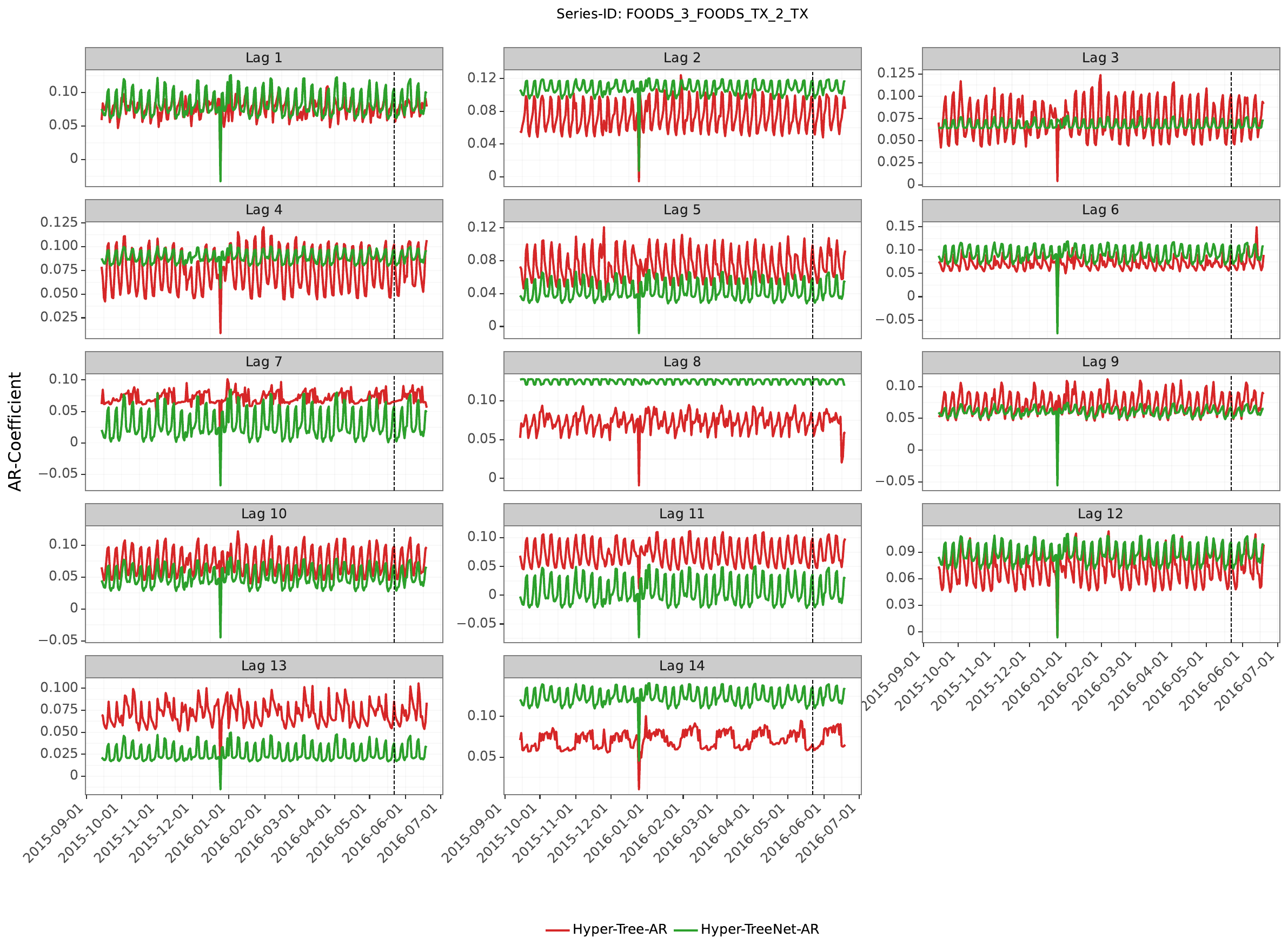}
	\label{fig:params_m5}
\end{figure}

It is important to note that the parameters estimated by \texttt{Hyper-Tree-AR(p)} and \newline \texttt{Hyper-TreeNet-AR(p)} models will generally differ from those estimated by conventional AR($p$) models. While conventional AR models estimate constant coefficients, Hyper-Trees estimate time-varying parameters that adapt to changing conditions based on features. This difference implies that autoregressive parameters in our framework represent conditional, context-specific relationships rather than fixed global values. Additionally, these dynamic parameters compensate for components not explicitly modeled in the target AR-structure. When the underlying data contains covariate effects, such as sales campaigns in the \texttt{M5} and \texttt{Rossmann} datasets, Hyper-Tree parameters naturally adapt to capture these influences rather than reflecting only the pure autoregressive structure. Another important consideration relates to stationarity conditions. Conventional AR($p$) models require parameter values within specific bounds to ensure stationarity, with these conditions assessed globally across the entire series. In our Hyper-Tree framework, however, AR-parameters vary with time and features, creating locally time-varying stationarity properties that may differ across time periods and feature contexts \citep{Dahlhaus.2000, Dahlhaus.2012}. This introduces the possibility of non-stationary episodes within an otherwise well-behaved process, requiring careful monitoring of parameter trajectories over time.

\newpage
\section{Embedding Analysis}\label{app:embedding_dim}

\setcounter{figure}{0}
\setcounter{table}{0}
\renewcommand{\thefigure}{G\arabic{figure}}
\renewcommand{\thetable}{G\arabic{table}}

The dimensionality of tree embeddings is an important architectural choice in the \texttt{Hyper-TreeNet} framework. While we use $d=1$ as the default for all experiments, this section investigates the role of embedding dimensions in more detail. We first evaluate how the choice of the embedding dimension affects forecast accuracy across multiple datasets, and then also visualize the learned embeddings to understand what patterns they capture. 

\subsection{Effect on Forecast Accuracy}

To quantify the accuracy-efficiency trade-off, we evaluate embedding dimensions $d \in \{1, 3, 5, 10\}$ in a global time series model setting, varying only $d$ while keeping all remaining hyper-parameters fixed at their global model configurations as reported in Table \ref{tab:global_hyperparams}. Table \ref{tab:embedding-evaluation} presents results across datasets.

\begin{table}[!htbp]
	\fontsize{8pt}{9pt}\selectfont
	\begin{adjustwidth}{-1cm}{}
		\begin{center}
			\begin{threeparttable}
		\caption{Embedding Dimension Analysis for \texttt{Hyper-TreeNet-AR}.}
		\begin{tabular}{lc||cccccc|c}
			\toprule
			&  &  &  &  &  &  &  &  Runtime \\
			Dataset & Embedding-Dim & MAPE & sMAPE & WAPE & RMSE & MAE & MASE & (Minutes) \\
			\midrule	
			\multirow[t]{8}{*}{\texttt{Air Passengers}} 
			 & \texttt{1} & 4.119 & 4.144 & 3.972 & 21.289 & 18.915 & 0.849 & \textbf{0.019} \\
			 & \texttt{3} & 2.474 & 2.454 & 2.364 & \textbf{14.737} & 11.256 & 0.505 & 0.023 \\
			 & \texttt{5} & \textbf{2.449} & \textbf{2.405} & \textbf{2.342} & 15.595 & \textbf{11.154} & \textbf{0.501} & 0.026 \\
			 & \texttt{10} & 2.597 & 2.532 & 2.390 & 16.663 & 11.382 & 0.511 & 0.033 \\
			\cline{1-9}	
			\multirow[t]{8}{*}{\texttt{Australian Electricity Demand}} 
			 & \texttt{1} & \textbf{2.964} & \textbf{2.950} & \textbf{2.960} & \textbf{163,026.9} & \textbf{121,972.3} & \textbf{0.886} & \textbf{0.022} \\
			 & \texttt{3} & 3.255 & 3.248 & 3.232 & 169,567.5 & 129,566.5 & 0.959 & 0.031 \\
			 & \texttt{5} & 3.507 & 3.456 & 3.474 & 189,624.7 & 158,295.1 & 1.090 & 0.039 \\
			 & \texttt{10} & 4.792 & 4.699 & 4.693 & 240,631.8 & 216,584.5 & 1.497 & 0.055 \\
			\cline{1-9}
			\multirow[t]{8}{*}{\texttt{Australian Retail Turnover}} 
			& \texttt{1} & \textbf{6.403} & \textbf{6.292} & \textbf{6.427} & \textbf{16.179} & \textbf{13.409} & \textbf{1.090} & \textbf{0.109} \\
			& \texttt{3} & 9.429 & 10.135 & 9.819 & 34.982 & 29.894 & 2.006 & 0.194 \\
			& \texttt{5} & 11.208 & 10.627 & 11.204 & 47.466 & 41.005 & 2.468 & 0.261 \\
			& \texttt{10} & 10.008 & 11.035 & 10.192 & 20.635 & 16.789 & 1.730 & 0.468 \\
			\cline{1-9}
			\multirow[t]{8}{*}{\texttt{M3 Monthly}} 
			& \texttt{1} & 19.778 & 14.659 & 14.603 & 814.225 & 677.430 & 1.374 & \textbf{0.075} \\
			& \texttt{3} & 20.119 & 14.590 & 14.568 & 809.894 & 673.574 & 1.357 & 0.132 \\
			& \texttt{5} & 19.698 & 14.235 & 14.342 & \textbf{793.316} & \textbf{658.416} & \textbf{1.287} & 0.179 \\
			& \texttt{10} & \textbf{19.672} & \textbf{14.233} & \textbf{14.321} & 796.733 & 661.793 & 1.437 & 0.364 \\
			\cline{1-9}
			\multirow[t]{8}{*}{\texttt{M3 Yearly}} 
			& \texttt{1} & \textbf{21.782} & 18.389 & \textbf{18.133} & \textbf{1,233.0} & \textbf{1,088.0} & 1.558 & \textbf{0.031} \\
			& \texttt{3} & 22.919 & \textbf{18.184} & 19.000 & 1,425.5 & 1,242.2 & \textbf{1.539} & 0.039 \\
			& \texttt{5} & 24.506 & 18.594 & 20.560 & 1,697.2 & 1,454.1 & 1.589 & 0.048 \\
			& \texttt{10} & 24.733 & 18.672 & 20.645 & 1,687.8 & 1,451.9 & 1.662 & 0.075 \\
			\cline{1-9}
			\multirow[t]{8}{*}{\texttt{M5}} 
			& \texttt{1} & \textbf{15.290} & \textbf{13.671} & \textbf{13.926} & \textbf{79.713} & \textbf{63.052} & \textbf{0.990} & \textbf{0.049} \\
			& \texttt{3} & 16.055 & 13.990 & 14.378 & 81.141 & 65.103 & 1.017 & 0.078 \\
			& \texttt{5} & 16.699 & 14.457 & 14.946 & 86.269 & 70.111 & 1.079 & 0.109 \\
			& \texttt{10} & 16.737 & 14.434 & 14.893 & 89.909 & 72.489 & 1.087 & 0.199 \\
			\cline{1-9}
			\multirow[t]{8}{*}{\texttt{Rossmann Store Sales}} 
			& \texttt{1} & 9.437 & 9.597 & 9.449 & 857.004 & 664.496 & 0.385 & \textbf{0.693} \\
			& \texttt{3} & 9.616 & 9.804 & 9.657 & 874.438 & 677.550 & 0.389 & 1.435 \\
			& \texttt{5} & \textbf{9.071} & \textbf{9.163} & \textbf{9.099} & \textbf{827.530} & \textbf{637.683} & \textbf{0.367} & 2.078 \\
			& \texttt{10} & 9.903 & 10.219 & 10.116 & 913.647 & 708.252 & 0.410 & 3.858 \\
			\cline{1-9}
					\multirow[t]{8}{*}{\texttt{Tourism}} 
			& \texttt{1} & \textbf{23.731} & \textbf{20.603} & \textbf{19.587} & \textbf{3,031.1} & \textbf{2,515.8} & \textbf{1.113} & \textbf{0.135} \\
			& \texttt{3} & 31.894 & 24.327 & 24.489 & 5,120.5 & 4,346.6 & 1.554 & 0.280 \\
			& \texttt{5} & 40.221 & 29.981 & 31.602 & 6,829.6 & 5,932.0 & 2.179 & 0.376 \\
			& \texttt{10} & 34.926 & 29.595 & 28.542 & 3,813.3 & 3,108.7 & 1.743 & 0.694 \\
			\bottomrule
		\end{tabular}
		\label{tab:embedding-evaluation}
		\begin{tablenotes}
		\scriptsize
		\setlength{\leftmargin}{0pt}
		\setlength{\labelsep}{0pt}
		\setlength{\labelwidth}{0pt}
		\setlength{\itemindent}{0pt}
		\item[] Reported are the mean forecast errors across series per dataset and total runtimes in minutes, with lower values indicating better performance and the best metrics highlighted in bold. Mean Absolute Percentage Error (MAPE); Symmetric Mean Absolute Percentage Error (sMAPE); Weighted Absolute Percentage Error (WAPE); Root Mean Squared Error (RMSE); Mean Absolute Error (MAE); Mean Absolute Scaled Error (MASE), which is calculated relative to the \texttt{AutoETS} forecasts (MAE of model / MAE of \texttt{AutoETS}). All experiments use identical hyper-parameters as reported in Table \ref{tab:global_hyperparams}. The \texttt{Air Passengers} dataset contains a single series, for which the local and global model settings are equivalent.
		\end{tablenotes}
	\end{threeparttable}
\end{center}
\end{adjustwidth}
\end{table}

Despite its simplicity, $d=1$ achieves the best or near-best accuracy on most datasets. The \texttt{Air Passengers} dataset shows the most pronounced benefit from higher embedding dimensions, while \texttt{Rossmann} and \texttt{M3 Monthly} show smaller gains. Several factors might explain this pattern.

First, the GBDT's recursive partitioning captures feature interactions implicitly through root-to-leaf decision paths, with successive splits across categorical, numeric, and temporal features. Hence, each leaf corresponds to a region of the feature space (e.g., `weekday, promotion, store type A'), and the embeddings capture this compressed context, representing complex non-linear relationships and interaction effects rather than encoding raw features directly. This may explain why a single embedding dimension can often be sufficient, since the trees already provide an interaction-rich feature representation. The random projection layer in our \texttt{Hyper-TreeNet} architecture provides another key mechanism that compensates for low-dimensional tree embeddings. As described in Section \ref{sec:hta}, this layer expands the $d$-dimensional tree embeddings to $k$ dimensions (where $k$ equals the number of target model parameters) before the MLP processes it, effectively creating $k$ views of the original embeddings. This diversity proves essential: without the expansion, forecast accuracy deteriorates, as ablation study \texttt{A4} in Table \ref{tab:ablation} demonstrates. The random projection hence serves as an effective bridge between the compact tree embeddings and the MLP, providing sufficient representational diversity even when $d=1$. Beyond the role of random projections, higher embedding dimensions increase the overall model capacity and introduce additional learning targets in the GBDT component. For datasets with limited complexity or sample size, this additional capacity may not be warranted and can lead to degraded generalization. In such cases, the constraint imposed by $d=1$ acts as a beneficial information bottleneck, forcing the tree ensemble to compress information into a single dimension that captures the most salient patterns while discarding noise.

However, for datasets with rich feature spaces, this same compression can become a limiting factor. The \texttt{Rossmann} dataset illustrates this case: it contains a rich set of categorical features (store types, assortment levels, promotional indicators, $\ldots$) and temporal features (day-of-week, holiday indicators, school holiday patterns, $\ldots$).  When the feature space is sufficiently complex and the dataset large enough to support learning additional parameters, $d=1$ may impose an overly restrictive bottleneck, forcing the GBDT to compress complex feature information into a single embedding. The accuracy gains with higher-dimensional embeddings can be understood through the lens of representational capacity: with $d>1$, the tree ensemble can encode multiple complementary aspects of the feature space, enabling richer downstream parameter adaptation. Interestingly, the \texttt{Air Passengers} dataset also benefits from higher embedding dimensions (WAPE improving from 3.972 at $d=1$ to 2.342 at $d=5$), despite being a single series with only temporal features. Here, the benefit appears to stem not from feature complexity but rather from the dataset's pronounced seasonality: with $d>1$, the model can encode complementary aspects of the seasonal pattern, rather than compressing all temporal information into a single dimension. For higher dimensions ($d > 5$), a different pattern emerges. Across the datasets evaluated, except for modest gains on the \texttt{M3 Monthly} dataset, $d=10$ does not yield accuracy improvements over lower dimensions. This suggests diminishing returns: beyond a certain point, additional embedding capacity no longer captures meaningful structure but instead seems to increase the risk of overfitting or training instability. The same pattern is also apparent on \texttt{Australian Retail Turnover} and \texttt{Tourism}, where $d=5$ reduces accuracy substantially.

Based on these results, we offer the following general guidance for practitioners. For most forecasting applications with modest feature sets and moderate dataset sizes, $d=1$ provides an effective default choice, offering computational efficiency while achieving competitive accuracy. The random projection mechanism ensures adequate representational capacity despite the low embedding dimensionality. Intermediate embedding dimensions ($d=3$) typically perform within a few percentage points of the best variant on most datasets, providing a reasonable middle ground between the efficiency of $d=1$ and the additional capacity of $d=5$. For complex operational datasets with rich categorical features, $d=5$ may yield meaningful accuracy improvements, though as the \texttt{Australian Retail Turnover} and \texttt{Tourism} results show, this is not guaranteed and validation against lower embedding dimensions is advisable. Beyond $d=5$, dimensions generally show diminishing returns and sometimes degrade performance further, with $d=10$ rarely improving over lower dimensions. Given the dataset-dependent nature of optimal embedding dimensionality and the absence of reliable a priori guidance, practitioners without strong assumptions on their dataset characteristics may prefer to treat $d$ as a hyper-parameter, with $d \in \{1, 3, 5\}$ providing a reasonable search range.

\subsection{Visualizing Tree Embeddings}

The quantitative results in the preceding section evaluate how the choice of the embedding dimension affects forecast accuracy. To provide insight into the internal representations learned by the \texttt{Hyper-TreeNet-AR} model, we now visualize the tree embeddings for the \texttt{Air Passengers} dataset using both $d=1$ and $d=10$ configurations. We selected this dataset for its pronounced and regular seasonality, which facilitates interpretation of how embeddings encode temporal patterns.

\begin{figure}[h!]
	\centering
	\caption{Single tree embedding ($d=1$) for the \texttt{Air Passengers} dataset. Blue indicates training period, orange indicates forecast period.}
	\includegraphics[width=0.48\linewidth]{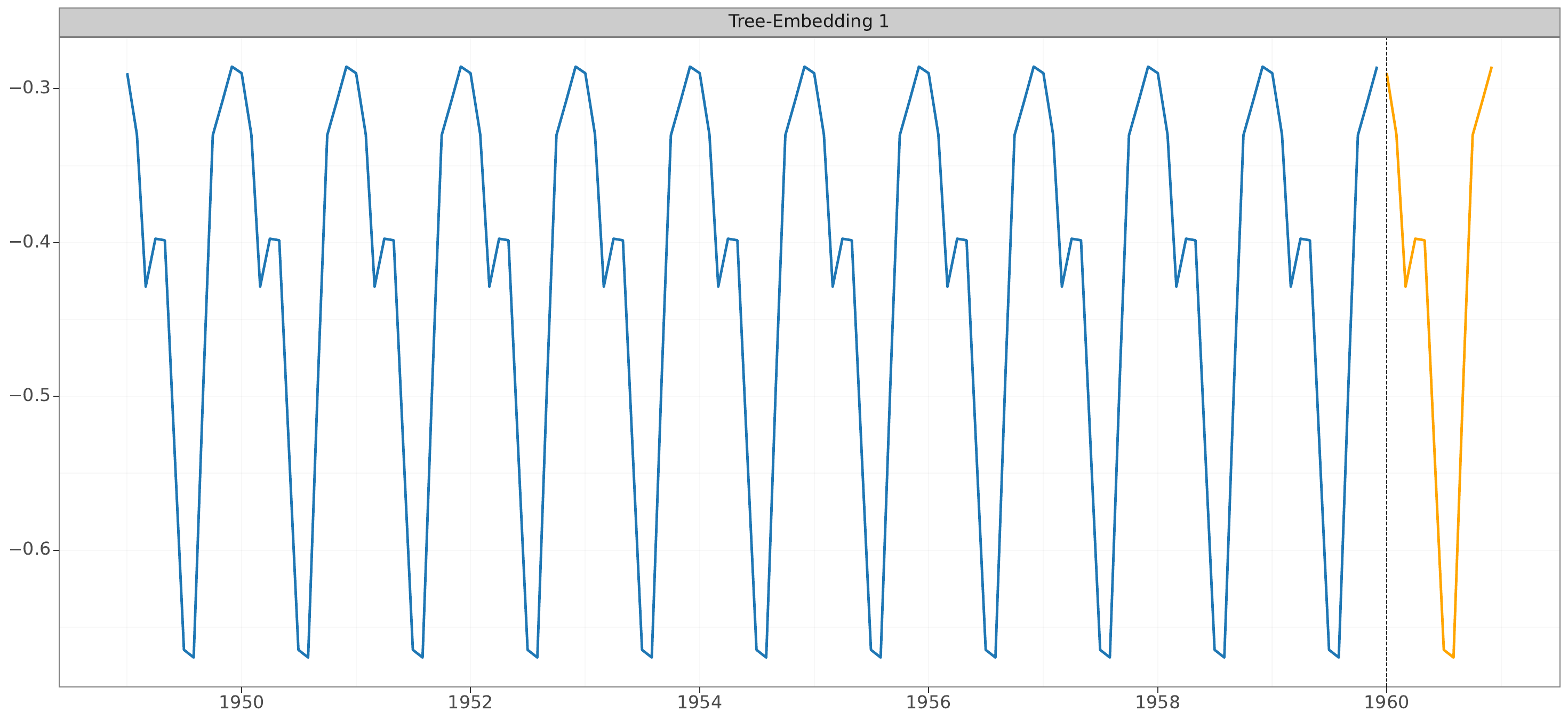}
	\label{fig:embed_d1}
\end{figure}

The single tree embedding in Figure \ref{fig:embed_d1} shows a regular 12-month oscillation. Since the only features are time-derived temporal indicators (month, quarter), identical months across different years map to identical embedding values, producing the observed year-over-year consistency. This visually confirms the discussion in Section \ref{sec:param_constraints} that similar feature contexts yield similar embeddings, preventing arbitrary parameter variation. The transition from training (blue) to forecast (orange) period shows no discontinuity, with the embedding continuing its seasonal oscillation naturally because the temporal features follow the same 12-month cycle. The single embedding compresses all seasonal information into one dimension, forcing the model to represent the annual cycle along a single axis. Despite this compression, the model achieves reasonable accuracy, demonstrating that the random projection layer can partially compensate for low-dimensional embeddings by expanding this single signal into multiple views for the MLP.

\begin{figure}[h!]
	\centering
	\caption{Multiple tree embeddings ($d=10$) for the \texttt{Air Passengers} dataset. Blue indicates training period, orange indicates forecast period.}
	\includegraphics[width=0.79\linewidth]{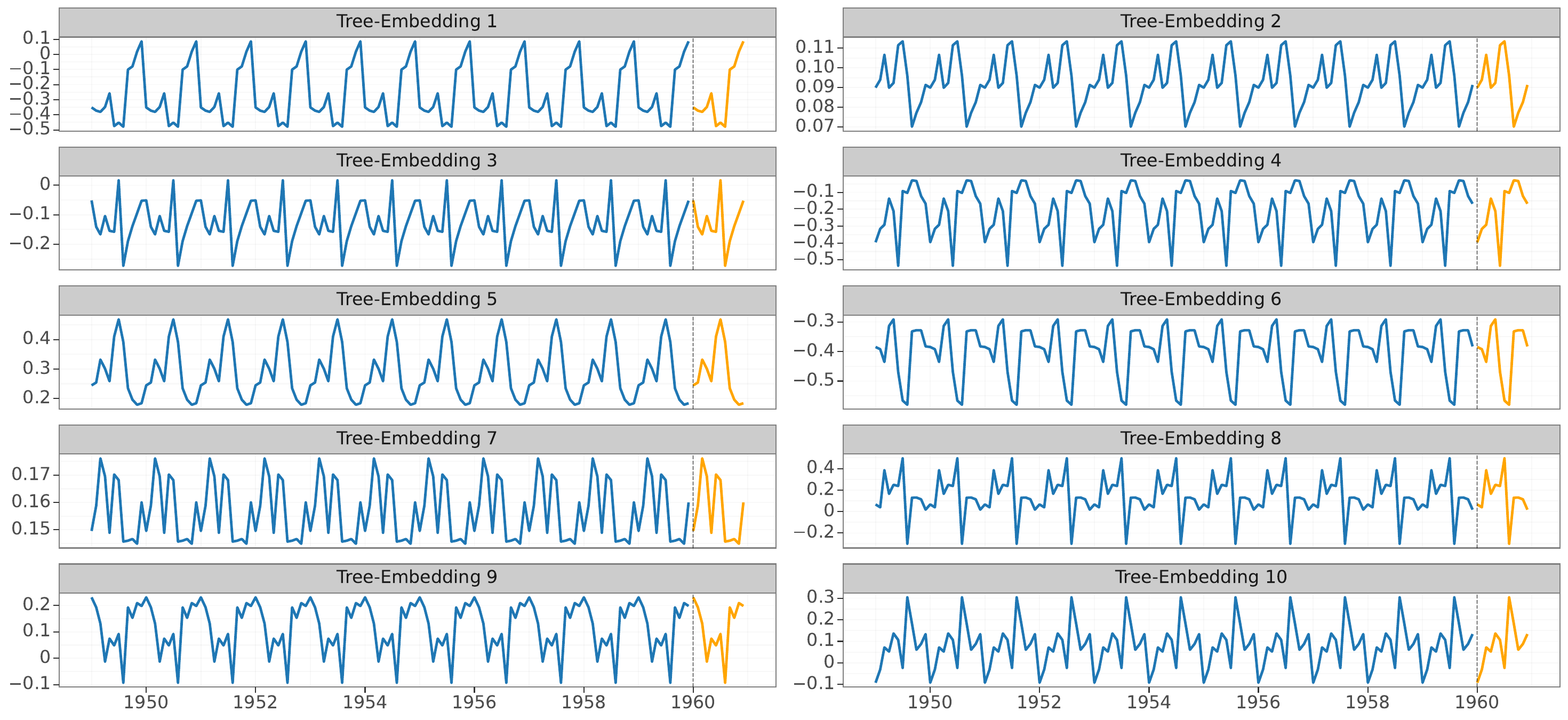}
	\label{fig:embed_d10}
\end{figure}

\newpage

With $d=10$, a more diverse set of patterns emerges. Most embeddings exhibit clear seasonal behavior, but with distinct characteristics: some show opposite phases, with one peaking where another troughs, while others differ in amplitude or shape. This diversity allows the model to represent complementary aspects of the seasonal structure that a single embedding would compress into one dimension. However, not every embedding is equally amenable to interpretation. For example, embeddings 7 and 9 show oscillatory behavior, but the signal is more subtle and does not align as directly with a single seasonal component. Comparing the two configurations hence reveals a trade-off between interpretability and accuracy. The single embedding ($d=1$) offers a clear, immediately interpretable representation of the seasonal cycle, allowing practitioners to directly read off seasonal effects from the embedding. The $d=10$ configuration sacrifices this clarity for improved accuracy, distributing seasonal information across multiple dimensions that are individually harder to interpret but collectively more expressive.\footnote{Users of the \texttt{Hyper-TreeNet-AR} model have access to SHAP values \citep{Lundberg.2017, Lundberg.2020}, which quantify how individual features influence each embedding dimension.} Compared to the single embedding, the higher-dimensional representation achieves higher accuracy (WAPE: 2.390 vs. 3.972 for $d=1$). However, consistent with Table \ref{tab:embedding-evaluation}, $d=5$ achieves the highest accuracy (WAPE: 2.342), with $d=10$ showing slight degradation. This aligns with our general finding that moderate to low embedding dimensions often provide the best balance between representational capacity and generalization. The relatively small gap in accuracy between $d=5$ and $d=10$ on this dataset likely reflects the pronounced seasonality of the \texttt{Air Passengers} dataset, where additional embedding capacity captures genuine seasonal variation rather than noise. As with $d=1$, all embeddings show smooth continuation from training to forecast periods, confirming that the feature-driven parameterization generalizes well to future time points with similar feature contexts. 

\newpage
\section{Hardware Specifications}

All experiments were conducted on a local machine with the following specifications:

\begin{itemize}
	\item CPU: 13th Gen Intel(R) Core(TM) i9-13900H
	\item CPU: Cores: 14
	\item RAM: 64 GB
	\item GPU: NVIDIA RTX 3500 Ada Generation Laptop GPU
	\item GPU Memory: 12 GB
\end{itemize}

\end{document}